\documentclass{article}

\usepackage[nonatbib, final]{neurips_2024}

\usepackage[utf8]{inputenc} 
\usepackage[T1]{fontenc}    
\usepackage{hyperref}       
\usepackage{url}            
\usepackage{booktabs}       
\usepackage{amsfonts}       
\usepackage{nicefrac}       
\usepackage{microtype}      
\usepackage{xcolor}         

\usepackage{graphicx}
\usepackage{comment}

\usepackage{times}
\usepackage{amsmath}
\usepackage{amssymb}
\usepackage{amsthm}

\usepackage{caption}

\usepackage{enumitem}
\usepackage{makecell}
\usepackage{bm}
\usepackage{mathtools}
\usepackage{algorithm,algpseudocode}
\usepackage{wrapfig}

\usepackage{float}
\usepackage{subfig}

\usepackage{multirow}

\usepackage{amsmath,amsfonts,bm}

\def\eqref#1{equation~\ref{#1}}

\def\ceil#1{\lceil #1 \rceil}
\def\floor#1{\lfloor #1 \rfloor}
\def\1{\bm{1}}

\DeclareMathAlphabet{\mathsfit}{\encodingdefault}{\sfdefault}{m}{sl}
\SetMathAlphabet{\mathsfit}{bold}{\encodingdefault}{\sfdefault}{bx}{n}

\DeclareMathOperator*{\argmin}{arg\,min}

\def\eqref#1{(\ref{#1})}

\theoremstyle{definition}

\theoremstyle{remark}

\allowdisplaybreaks

\newsavebox{\leftbox}
\newsavebox{\rightbox}
\definecolor{Gray}{gray}{0.9}

\makeatletter
\def\adl@drawiv#1#2#3{%
        \hskip.5\tabcolsep
        \xleaders#3{#2.5\@tempdimb #1{1}#2.5\@tempdimb}%
                #2\z@ plus1fil minus1fil\relax
        \hskip.5\tabcolsep}
\newcommand{\cdashlinelr}[1]{%
   \noalign{\vskip\aboverulesep
            \global\let\@dashdrawstore\adl@draw
            \global\let\adl@draw\adl@drawiv}
   \cdashline{#1}
   \noalign{\global\let\adl@draw\@dashdrawstore
            \vskip\belowrulesep}}
\makeatother

\title{All-in-One Image Coding for Joint Human-Machine Vision with Multi-Path Aggregation}

\author{
   Xu Zhang \quad Peiyao Guo \quad Ming Lu\thanks{Corresponding author} \quad Zhan Ma \\
   School of Electronic Science and Engineering \\
   Nanjing University \\
   \texttt{\{xu.zhang,peiyao\}@smail.nju.edu.cn} \quad \texttt{\{minglu,mazhan\}@nju.edu.cn}
}

\begin{document}

\maketitle

\begin{abstract}
  Image coding for multi-task applications, catering to both human perception and machine vision, has been extensively investigated. Existing methods often rely on multiple task-specific encoder-decoder pairs, leading to high overhead of parameter and bitrate usage, or face challenges in multi-objective optimization under a unified representation, failing to achieve both performance and efficiency. To this end, we propose Multi-Path Aggregation (MPA) integrated into existing coding models for joint human-machine vision, unifying the feature representation with an all-in-one architecture. MPA employs a predictor to allocate latent features among task-specific paths based on feature importance varied across tasks, maximizing the utility of shared features while preserving task-specific features for subsequent refinement. Leveraging feature correlations, we develop a two-stage optimization strategy to alleviate multi-task performance degradation. Upon the reuse of shared features, as low as 1.89\% parameters are further augmented and fine-tuned for a specific task, which completely avoids extensive optimization of the entire model. Experimental results show that MPA achieves performance comparable to state-of-the-art methods in both task-specific and multi-objective optimization across human viewing and machine analysis tasks. Moreover, our all-in-one design supports seamless transitions between human- and machine-oriented reconstruction, enabling task-controllable interpretation without altering the unified model. Code is available at \url{https://github.com/NJUVISION/MPA}.
\end{abstract}

\section{Introduction}
\label{sec:intro}

\textit{Coding for multi-task} that satisfies both human perception and machine vision has been extensively explored over the past few years~\cite{song2021variablerate, chen2023transtic, yan2021sssic, choi2022scalable, li2024humanmachine}. The intuitive and simple solution to achieve optimal performance for various tasks involves defining distinct encoder-decoder pairs tailored to specific instances with multiple bitstreams (separate~\cite{liu2022improving, song2021variablerate, chen2023transtic, le2021image, chamain2021endtoend, wang2023deep} or scalable~\cite{liu2021semantics, yan2021sssic, hu2020towards, wang2022towards, choi2022scalable, harell2022ratedistortion}), which however incurs significant parameter overhead and inefficient bitrate consumption. As a result, alternative efforts~\cite{chamain2021endtoendarxiv, duan2023unified, feng2022omnipotent, li2024humanmachine} attempt to enhance the compression performance across multiple tasks by deriving \textit{a unified and compact representation} of input images. Typically, they rely on a generalized encoder for feature extraction and deploy diverse decoding models to support corresponding vision tasks. However, they still suffer from the parametric inefficiency of multiple dedicated decoders. Recent work has made some progress in designing \textit{a unified compression model} for multi-task, where the encoder generates a unified representation, and the decoder focuses on task-oriented reconstruction.  Popular techniques include conditional generation~\cite{agustsson2023multirealism, iwai2024controlling} and residual prediction~\cite{iwai2021fidelitycontrollable, ghouse2023residual, korber2023egic}, enabling user-controllable modulation of decoded results through task-driven guidance. While achieving performance comparable to decoder-specific approaches, they solely focus on human-centric requirements, such as perception-oriented and fidelity-oriented aspects. More critically, the challenge of multi-objective optimization~\cite{miettinen1999nonlinear, ehrgott2005multicriteria}, particularly in the context of multi-task learning~\cite{sener2018multitask, momma2022multiobjective}, persists in the unified paradigm for joint human-machine vision, offering the performance largely inferior to that optimized for each task independently~\cite{le2021image, chamain2021endtoend, chamain2021endtoendarxiv}. The crux of the problem lies in the indiscriminate treatment of the unified representation across diverse tasks, which necessitates searching for an accurate Pareto front for transition between tasks. This requires the use of massive and complicated techniques to fit it~\cite{lin2019pareto, mahapatra2020multitask, ma2020efficient, wang2023deep} and multi-objective optimization with variable weights~\cite{agustsson2023multirealism, iwai2024controlling, ghouse2023residual, korber2023egic}. Its failure can lead to significant performance degradation. And as the number of tasks increases, the difficulty of optimal search grows dramatically~\cite{dimitriadis2023pareto, hu2023revisiting, gao2021towards}. So far, few approaches consider the correlation of features across tasks to ensure efficient multi-task collaboration while also optimizing for the specificity of distinct tasks.

To overcome the challenges above, we propose a unified image coding method with Multi-Path Aggregation (MPA) for joint human-machine vision tasks. Specifically, a set of Multi-Layer Perceptron (MLP) branches is inserted to form multiple aggregation paths, replacing the single-path MLP in the feature transform blocks~\cite{liu2021swin,liu2022convnet} which currently dominate compression models~\cite{lu2022transformer, duan2023lossy, lu2022highefficiency, zhu2022transformerbased, liu2023learned}. Each path is customized and tailored to different tasks with varying complexities. Considering diverse importance of features across different tasks, we devise a predictor for importance scores to allocate latent features among task-specific paths based on their importance. Leveraging feature correlations across tasks, we develop an efficient two-stage optimization strategy with fine-tuning partial parameters on generalized features to alleviate multi-task performance degradation, avoiding extensive optimization of the entire model. This strategy significantly eases the optimization of multi-task coding while maintaining performance comparable to other fully optimized models. Considering any viewing task for human vision (e.g., low distortion or high realism) or analysis task for machine vision (e.g., high-level or low-level vision task), MPA can switch flexibly between them with seamless transitions, enabling a single representation to be interpreted in different ways within a unified model.

Our contributions are summarized as follows:

\begin{itemize}[leftmargin=15pt,itemsep=2pt,topsep=2pt]
\item[$\bullet$] Considering different importance of features for different tasks, we propose MPA to non-uniformly treat features. By developing an importance score predictor, MPA can allocate generalized latent features among task-specific paths based on their varying importance. This enables a \emph{unified model} to support multi-task coding with seamless transitions in an all-in-one manner.

\item[$\bullet$] Leveraging feature correlations across tasks, we propose a two-stage optimization strategy with fine-tuning partial parameters on generalized features to overcome the challenge of multi-objective optimization in multi-task coding. This strategy allows MPA to be easily extended to support new tasks without independent optimization of separate task-specific models.

\item[$\bullet$] Extensive experiments demonstrate that using the unified model, MPA achieves rate-distortion (R-D) and rate-perception (R-P) performance for human vision comparable to other state-of-the-art (SOTA) models and significantly improves accuracy for analysis tasks close to the fully fine-tuned ones, outperforming other separately optimized methods.
\end{itemize}

\begin{figure}[t]
\begin{center}
\subfloat[Separate pairs]{\includegraphics[scale=0.425]{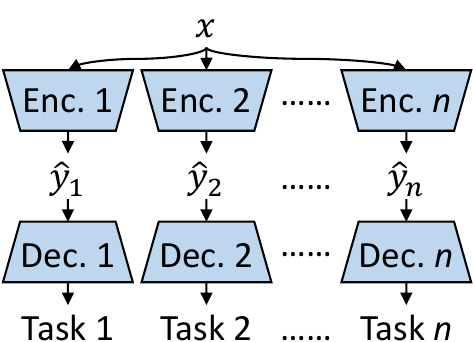}\label{subfig:separate}}\hspace{30pt}
\subfloat[Unified representation]{\includegraphics[scale=0.425]{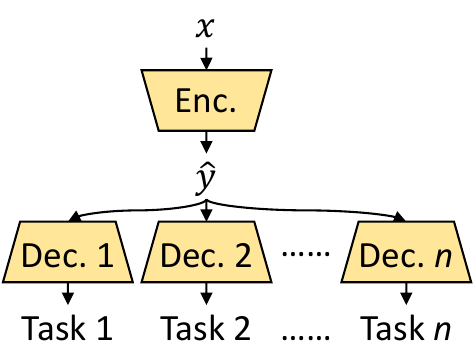}\label{subfig:unified_representation}}\hspace{30pt}
\subfloat[Unified model]{\includegraphics[scale=0.425]{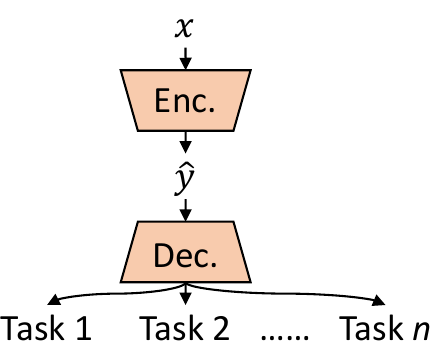}\label{subfig:unified_model}}\hfill
\caption{Paradigm comparison for multi-task coding.}
\label{fig:paradigm}
\end{center}
\end{figure}

\section{Related work}
\label{sec:related_work}

\textbf{Separate pairs.}
Developing multiple task-specific encoder-decoder pairs, as shown in Fig.~\ref{subfig:separate}, is easy to optimize and beneficial to high performance but has significant parameter and bitrate overhead~\cite{liu2022improving, song2021variablerate, chen2023transtic, le2021image, chamain2021endtoend}. Song et al.~\cite{song2021variablerate} introduced Spatially-adaptive Feature Transform (SFT) for tuning bit allocation, incurring significant latency for optimizing quality map for each image to achieve best task performance. Chen et al.~\cite{chen2023transtic} leveraged visual prompt tuning (VPT)~\cite{jia2022visual} for task-specific optimization, resulting in quadratic complexity due to the self-attention mechanism with respect to the number of VPT tokens. Scalable coding~\cite{liu2021semantics, yan2021sssic, hu2020towards, wang2022towards, choi2022scalable, harell2022ratedistortion} improved bitstream efficiency by embedding multiple bitstreams in a scalable manner. However, organizing representations for various tasks in a layered manner without introducing redundancy is challenging.

\textbf{Unified representation.}
Some efforts~\cite{chamain2021endtoendarxiv, duan2023unified, feng2022omnipotent, li2024humanmachine} aimed to derive a unified and compact representation of input images but developed separate decoders (Fig.~\ref{subfig:unified_representation}) to ease optimization challenges. Feng et al.~\cite{feng2022omnipotent} compressed intermediate features from a vision backbone as a generic representation, while Duan et al.~\cite{duan2023unified} proposed a plug-and-play adapter to bridge compressed representations with existing vision backbones, both requiring optimization of the backbone for optimal performance. Li and Zhang~\cite{li2024humanmachine} used a semantic enhancement network to improve analysis accuracy without jointly optimizing the vision backbone but still faced parameter overhead due to multiple decoders.

\textbf{Unified model.}
The paradigm in Fig.~\ref{subfig:unified_model} uses a single encoder to generate a unified representation and a single decoder for task-specific reconstructions~\cite{gao2021towards, agustsson2023multirealism, ghouse2023residual, korber2023egic, iwai2024controlling}. Gao et al.~\cite{gao2021towards} optimized the unified model for multiple analysis tasks jointly without transitions, resulting in trade-offs and suboptimal performance. Ghouse et al.~\cite{ghouse2023residual} and Korber et al.~\cite{korber2023egic} reconstructed images with a basic model and used additional modules to predict and add residuals for task-specific goals, while Agustsson et al.~\cite{agustsson2023multirealism} and Iwai et al.~\cite{iwai2024controlling} developed hyper-parameter conditioning modules to modulate the transition process. These methods require variable weighted objectives during optimization processes, making it challenging to extend to more tasks.

\section{A unified framework: multi-path aggregation}
\label{sec:framework}

\begin{figure}[t]
\begin{center}
\includegraphics[width=0.97\textwidth]{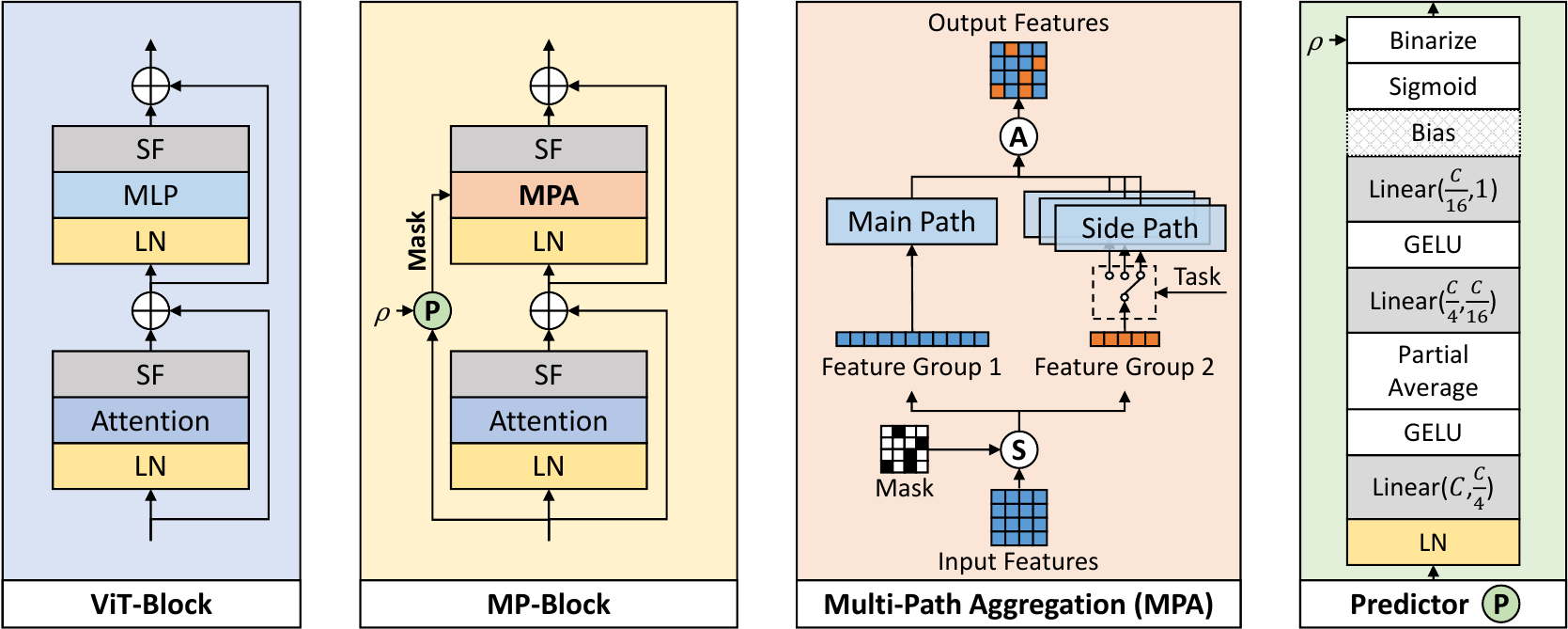}\hfill
\caption{The proposed Multi-Path Aggregation (MPA). Compared to typical Vision Transformer (ViT) block~\cite{vaswani2017attention, dosovitskiy2021vit}, original MLP is replaced with MPA. LN and SF are Layer Normalization and Scaling Factors~\cite{Chen2020}. \textcircled{\small{P}}, \textcircled{\small{S}} and \textcircled{\small{A}} denote predictor, split and aggregation respectively. $C$ represents the number of input channels. $\rho$ is the ratio $\rho_{\text{enc}}$ in the encoder or the ratio $\rho_{\text{dec}}$ in the decoder.}
\label{fig:MPA}
\end{center}
\end{figure}

\subsection{Preliminaries}

Typically, the LIC model involves an encoder network $E(\bm{x})$ that maps the input image $\bm{x}$ to a compact latent representation $\bm{\hat{y}}$, and a decoder network $G(\bm{\hat{y}})$ that reconstructs $\bm{\hat{x}}$ as an approximation to $\bm{x}$. For the best compression performance, the optimization objective minimizes the distortion between the original and reconstructed images while reducing the expected bitrate $r(\bm{\hat{y}})$, using $\mathbb{E}_{\bm{x} \sim p_{\bm x}}[-\log_{2}p_{\bm{\hat{y}}}(\bm{\hat{y}})]$ characterized by entropy model $P$. To achieve perceptually pleasing reconstruction, a Generative Adversarial Network (GAN) strategy is often incorporated, where a conditional discriminator $D(\text{cond.}, \text{input})$ is used to assist in model optimization. In our work, the MPA is implemented in TinyLIC~\cite{lu2022highefficiency} to validate its functionality for joint human-machine vision coding, while the GAN strategy in HiFiC~\cite{mentzer2020highfidelity} is used for guiding perceptual optimization. Furthermore, to support continuously variable-rate coding, we introduce the scaling factor (SF) modulation $\bm{s}_q$~\cite{Chen2020} and non-linear interpolation~\cite{cui2021asymmetric, lee2022selective} in each Multi-Path Block (MP-Block in Fig.~\ref{fig:MPA}). The compression quality $q \in [1, Q_{\text{max}}]$ matching in the encoder and decoder determines the bitrate consumption, while the task orientation $\alpha \in [0, 1]$ in the decoder controls the continuously transitions between tasks.

\subsection{Multi-path aggregation}
\label{sec:mpa}

To support various task optimizations flexibly with the unified model, we develop an efficient architecture named MPA. As illustrated in Fig.~\ref{fig:MPA}, the proposed method contains two key modules, multiple paths with MLP layers of varying complexity, and a predictor for binarized mask $\bm{M}$. Different from recent advances in Mixture-of-Experts (MoE)~\cite{du2022glam, fedus2022switch}, the paths are divided into two groups: the main and the side paths. The main path captures generalized features for various tasks, while the side paths are fine-tuned based on the generalized features. The input features are allocated into different paths on the basis of $\bm{M}$. The features corresponding to ``1'' in $\bm{M}$ will enter the main path, while the remaining features will enter the side paths. Unlike existing methods, which apply consistent transformations~\cite{ghouse2023residual, korber2023egic, agustsson2023multirealism, iwai2024controlling} to all features, MPA only exploits a few parameters in the side path to refine the generalized features for task transitions.

\begin{wrapfigure}[14]{r}{0.45\textwidth}
\vspace{-2.2em}
\begin{minipage}{0.45\textwidth}
\begin{algorithm}[H]
    \begin{center}
    \caption{Multi-Path Aggregation}\label{alg:MPA}
    \begin{algorithmic}[1]
        \Require{feature $\bm{x}$, mask $\bm{M}$, task index $i_{\text{task}}$}
        \If{encoding}
            \State $\phi_{\text{main}} \leftarrow \phi_{\text{hq}}, \phi_{\text{side}} \leftarrow \phi_{\text{lq}}$
        \ElsIf{decoding}
            \State $\phi_{\text{main}} \leftarrow \phi_{\text{perc}}$
            \State $\phi_{\text{side}} \leftarrow \{\phi_{\text{MSE}}, \phi_{\text{cls}}, \phi_{\text{seg}}\}[i_{\text{task}}]$
        \EndIf
        \State $\{\bm{x}_1, \bm{x}_2\} \leftarrow \text{Split}(\bm{x}, \bm{M})$
        \State $\bm{x}_1 \leftarrow \phi_{\text{main}}(\bm{x}_1), \bm{x}_2 \leftarrow \phi_{\text{side}}(\bm{x}_2)$
        \State $\bm{x} \leftarrow \text{Aggregate}(\bm{x}_1, \bm{x}_2)$
        \item[]
        \Return{$\bm{x}$}
    \end{algorithmic}
    \end{center}
\end{algorithm}
\end{minipage}
\end{wrapfigure}

On the encoder side, we set the original MLP in ViT block as the main (high-quality) path $\phi_{\text{hq}}$ adapted to high bitrate features and add a bottleneck MLP as the side (low-quality) path $\phi_{\text{lq}}$ specific to low bitrate coding. For task-controllable LIC, the features optimized for perceptual loss are the most generalized ones~\cite{korber2023egic}. Therefore, on the decoder side, the main path is the Perc. (Perceptual) Path $\phi_{\text{perc}}(\cdot)$, while the side paths contain the MSE path $\phi_{\text{MSE}}(\cdot)$, the Cls. (Classification) Path $\phi_{\text{cls}}(\cdot)$, and the Seg. (Segmentation) Path $\phi_{\text{seg}}(\cdot)$, standing for four representative tasks to showcase the versatility of MPA, i.e., high-realism, low-distortion, high-level visual, and low-level visual image reconstructions. Setting the task index $i_{\text{task}}$ by the user, MPA can realize the aggregation of the main path and one of the side path. To be efficient, we configure each path for a different complexity depending on the characteristics of tasks. Specifically, the paths for human vision (i.e., $\phi_{\text{perc}}(\cdot)$ and $\phi_{\text{MSE}}(\cdot)$) are designed as an inverted (inv.) bottleneck MLP~\cite{liu2022convnet} to achieve higher realism and lower distortion, while the paths for machine vision (i.e., $\phi_{\text{cls}}(\cdot)$ and $\phi_{\text{seg}}(\cdot)$) was designed as a bottleneck MLP to reduce inference latency. Having an input feature $\bm{x} \in \mathbb{R}^{H \times W \times C}$, $i_{\text{task}}$ and a mask $\bm{M} \in \{0,1\}^{H \times W \times 1}$, MPA processes as Alg.~\ref{alg:MPA}. In our implementation, we integrate MPA into the 1st, 2nd and 3rd stages in both the encoder and decoder, without the 4th stage since its spatial size is too small to produce a smooth transition.

\subsection{Importance score predictor}
\label{sec:predictor}

For the portability of MPA, we introduce a lightweight importance score predictor as illustrated in Fig.~\ref{fig:MPA} (denoted as \textcircled{\small{P}}) to generate mask and allocate features to each path. In addition to only three parametric linear layers for fast point-wise computation, we use a non-parametric partial average layer $\mathcal{A}(\cdot)$ to capture multi-scale information. Given intermediate feature $\bm{u} \in \mathbb{R}^{H \times W \times C'}$, $\mathcal{A}(\cdot)$ performs global average pooling on the latter $C' / 2$ channels of intermediate feature followed by expansion to the spatial size of feature, which aggregates information from all features as global information. The other $C' / 2$ channels are left unchanged as local information. Formally, the partial average operation can be formulated as
\begin{equation}
\label{eq:average}
\mathcal{A}\left(\bm{u}(h, w, c)\right) =
\begin{cases}
  \frac{\sum_{i=1}^H \sum_{j=1}^W \bm{u}(i,j,c)}{H \times W}, & \text{if}\ c > C' / 2, \\
  \bm{u}(h, w, c), & \text{otherwise}.
\end{cases}
\end{equation}
A hyper parameter ($q$ or $\alpha$) is leveraged to control the feature allocation between the main and side paths, which indicates the performance transition between tasks. In the encoder, the aggregation of $\phi_{\text{hq}}$ and $\phi_{\text{lq}}$ is associated with bitrate, thus we approximate a linear relationship between bitrate and aggregation ratio $\rho_{\text{enc}}$ following~\cite{lee2022selective} by an inverse log transformation
\begin{equation}
\label{eq:rho}
    \rho_{\text{enc}} = \mathcal{F}(q) = \frac{\beta^{(q-1)/(Q_{\text{max}}-1)} - 1}{\beta-1} \in [0, 1], ~~q \in [1, Q_{\text{max}}],
\end{equation}
where $\beta$ is the base of the log transformation and $Q_{\text{max}}$ represents the maximum value of the range of $q$. Thus, as $q$ increases, more features will enter $\phi_{\text{hq}}(\cdot)$ for supporting high-quality reconstruction. The aggregation in the decoder is irrelevant to bitrate, and $\rho_{\text{dec}} = 1 - \alpha$ so that as $\alpha$ increases, more features will enter the side paths. For efficiency, the mask is only generated once at the beginning of a stage and shared by all MP-blocks throughout the stage.

During training, since the direct sampling of $\bm{M}$ based on $\rho$ from the importance score is non-differentiable, we first bias the output $\bm{u}^{\prime}$ of the last $\text{Linear}(\cdot)$ in the predictor using a set of learnable parameters $b$ to get shifted logits corresponding to each discrete $q$ or $\alpha$, and then use Gumbel-Sigmoid~\cite{jang2017categorical, meng2022adavit} with threshold $\tau = 0.5$ to soften the sampling process as Eq.~\eqref{eq:gumbel_sigmoid}. After training, the predictor can binarize the score map according to $\rho$, and the bias layer in Fig.~\ref{fig:MPA} will be discarded.
\begin{equation}
\label{eq:gumbel_sigmoid}
    \bm{M} = \text{Gumbel-Sigmoid}(\bm{u}^{\prime} + b, \tau).
\end{equation}

\section{Optimization strategy}
\label{sec:training}

\subsection{Stage 1: training a generalized basic model}
\label{sec:training_stage1}

Since the perceptually optimized LIC model is well aligned with both human and machine perception, we first train a variable-rate model based on TinyLIC~\cite{lu2022highefficiency} and GAN method from HiFiC~\cite{mentzer2020highfidelity} for subsequent extension. We add SF in all ViT blocks and implement MPA in the encoder. Following the common practice in literatures~\cite{mentzer2020highfidelity, agustsson2023multirealism, iwai2024controlling}, the optimized losses are Eqs.~\eqref{eq:loss_d} and~\eqref{eq:loss_egp} for the joint optimization of the encoder $E$, decoder $G$, entropy model $P$ and discriminator $D$:
\begin{align}
    \mathcal{L}_\text{ratio} &= \frac{1}{S}\sum_{s=1}^S \left( \rho_{\text{enc}} - \frac{1}{{H^{(s)}}{W^{(s)}}}\sum_{h=1}^{H^{(s)}}\sum_{w=1}^{W^{(s)}}{\bm{M}}^{(s)}(h,w) \right)^2,\label{eq:loss_ratio} \\
    \mathcal{L}_G &= \mathbb{E}_{\bm{\hat y} \sim p_{\bm{y}}}[-\log ( D(\bm{\hat y}, G(\bm{\hat y})) )],\label{eq:loss_g} \\
    \mathcal{L}_D &= \mathbb{E}_{\bm{\hat y} \sim p_{\bm{y}}}[-\log (1-D(\bm{\hat y}, G(\bm{\hat y}))] + \mathbb{E}_{\bm{x}\sim p_{\bm{x}} }[-\log D(E(\bm{x}), \bm{x})],\label{eq:loss_d} \\
    \mathcal{L}_{EGP} &= \mathbb{E}_{\bm{x}\sim p_{\bm{x}}} [\lambda_{r}^{(q)} r(\bm{\hat y}) + d(\bm{x}, \bm{\hat x})] + \lambda_{G} \mathcal{L}_{G} + \lambda_{\text{perc}} \mathcal{L}_{\text{perc}} + \lambda_{\text{ratio}} \mathcal{L}_{\text{ratio}},
    \label{eq:loss_egp}
\end{align}
where $\bm{x}$, $\bm{\hat{x}}$, $\bm{y}$ and $\bm{\hat{y}}$ are the input image, reconstructed image, compressed latents before and after quantization, respectively. $\mathcal{L}_{\text{ratio}}$ is introduced to optimize the predictors in MP-Blocks which constrains the ratio of ``1''s in $\bm{M}$ at all encoder stages to the encoder ratio target $\rho_{\text{enc}}$. $S$ is the number of stages applying MPA, while $H^{(s)}$ and $W^{(s)}$ represent the spatial size at Stage $s$. $r(\cdot)$ and $d(\cdot, \cdot)$ represent the bitrate estimated by the entropy model $P$ and $0.01\text{MSE}(\cdot, \cdot)$ (pixel values are scaled to $[0,255]$). $\lambda_{r}^{(q)}$ in Eq.~\eqref{eq:loss_egp} varies according to the sampling of quality level $q$ during training. Learned Perceptual Image Patch Similarity (LPIPS)~\cite{zhang2018unreasonable} is chosen for $\mathcal{L}_{\text{perc}}$.

\subsection{Stage 2: extending and optimizing side paths for multi-task}
\label{sec:training_stage2}

Obtaining a generalized model, we can add side paths to the decoder to achieve task-controllable image coding. In this training stage, there is no need to train the model entirely, but only to optimize the added parameters, i.e., the added side path $\phi_{\text{side}}$ and the corresponding predictor $\phi_{\text{pred}}$. By sampling $\alpha$ during training, the main path and the added side path are randomly aggregated to fit the transition between different tasks. The training objective is simplified to
\begin{align}
    (\phi_{\text{side}}^*, \phi_{\text{pred}}^*) &= \argmin_{\phi_{\text{side}}, \phi_{\text{pred}}} \mathbb{E}_{\bm{x}\sim p_{\bm{x}}} [\lambda_{r}^{(q)} r(\bm{\hat y})] + \lambda_{\text{task}} \mathcal{L}_{\text{task}} + \lambda_{\text{ratio}} \mathcal{L}_{\text{ratio}},\label{eq:loss_rt}
\end{align}
where $\phi_{\text{side}}^*$ and $\phi_{\text{pred}}^*$ are the optimal parameters of the added side path and predictor, $\mathcal{L}_{\text{task}}$ represents the task loss, and the target ratio in $\mathcal{L}_{\text{ratio}}$ is changed to decoder ratio $\rho_{\text{dec}}$. The options for task loss are varied. When optimizing for MSE, $\mathcal{L}_{\text{task}}$ is just a simple MSE loss measured between $\bm{x}$ and $\bm{\hat{x}}$. When optimizing for a visual analysis task, $\mathcal{L}_{\text{task}}$ is a compound loss containing $d(\bm{x}, \bm{\hat x})$, $\mathcal{L}_{\text{perc}}$ and the full loss function of the task. Normalization should be applied if needed for augmentation. The task model is frozen during optimization. For instance, using the classification model $\text{ClsModel}(\cdot)$ with $\text{Norm}(\cdot)$, cross-entropy is used to compute loss between the classification result of $\bm{\hat{x}}$ and the ground truth $GT$:
\begin{align}
    \mathcal{L}_{\text{task}} &= \text{CrossEntropy}(\text{ClsModel}(\text{Norm}(\bm{\hat{x}})), GT) + \mathbb{E}_{\bm{x}\sim p_{\bm{x}}}[d(\bm{x}, \bm{\hat x})] + \lambda_{\text{perc}} \mathcal{L}_{\text{perc}}.
    \label{eq:loss_task}
\end{align}

\section{Experiments}

\begin{figure}[t]
\begin{center}
\subfloat{\includegraphics[scale=0.403]{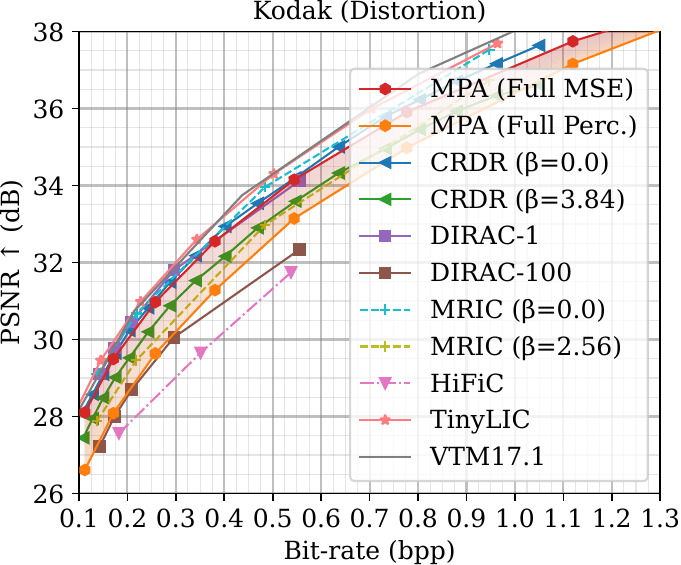}\label{subfig:kodak_rd}}
\subfloat{\includegraphics[scale=0.403]{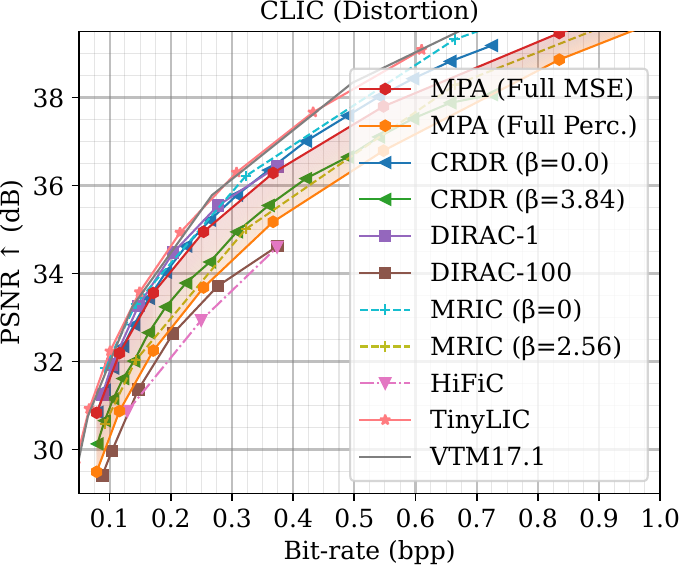}\label{subfig:clic_rd}}
\subfloat{\includegraphics[scale=0.403]{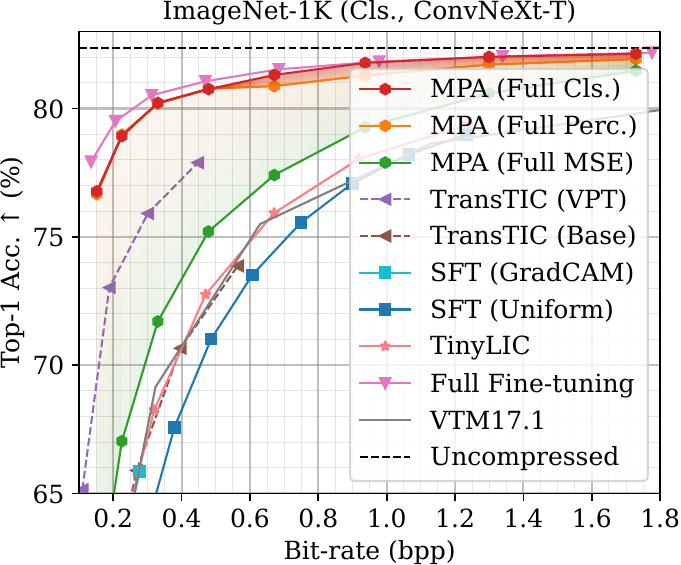}\label{subfig:convnext_t}}\vspace{-8pt}\\
\subfloat{\includegraphics[scale=0.391]{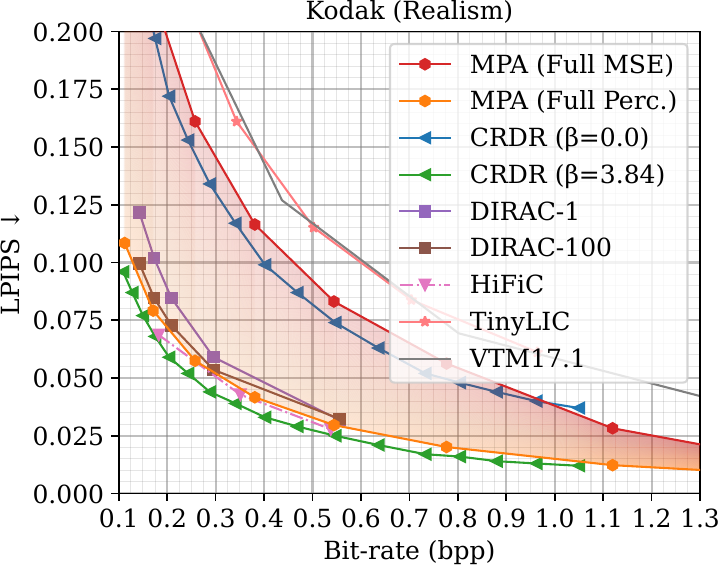}\label{subfig:kodak_lpips}}
\subfloat{\includegraphics[scale=0.391]{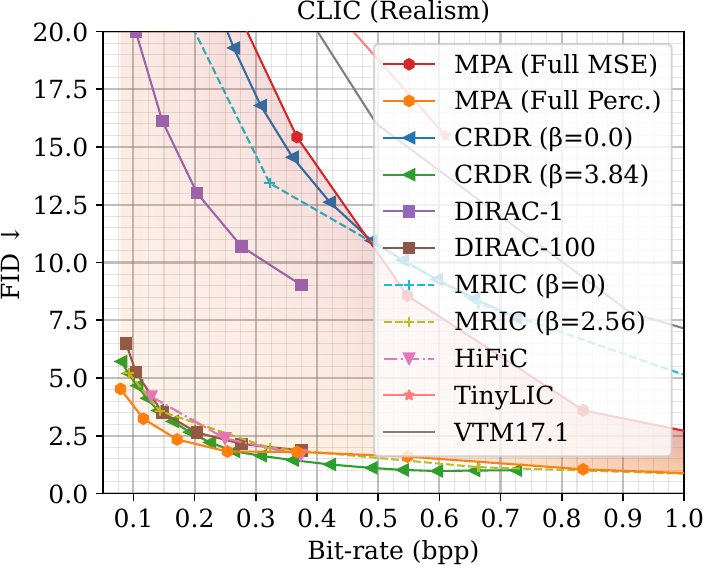}\label{subfig:clic_fid}}
\subfloat{\includegraphics[scale=0.391]{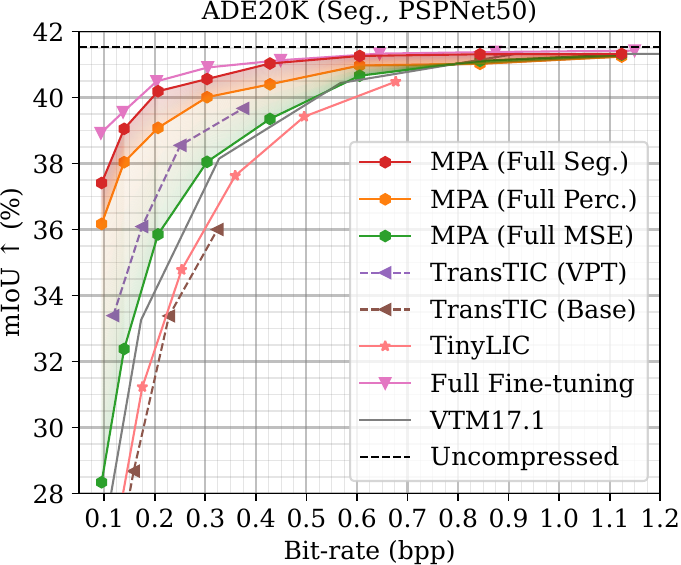}\label{subfig:pspnet50}}\hfill
\end{center}
\caption{Multi-task performance. The curves of variable-rate models are plotted as solid lines, while dashed lines are for single-rate models. Colored areas represent the adjustable range of MPA.}
\label{fig:performance}
\end{figure}

\subsection{Experimental settings}
\label{sec:settings}

\textbf{Dataset.} When training for perceptual quality and MSE, we use a combined dataset including Flicker2W~\cite{flicker2w}, DIV2K~\cite{div2k}, and CLIC~\cite{CLIC2020} training sets, about 23K images in total. ImageNet-1K~\cite{imagenet1k} and ADE20K~\cite{ade20k} are used to train $\phi_{\text{cls}}$ and $\phi_{\text{seg}}$, respectively. We evaluate the model on Kodak~\cite{KODAK} and CLIC test set~\cite{CLIC2020} for image compression, ImageNet validation set~\cite{imagenet1k} for classification, and ADE20K validation set~\cite{ade20k} for semantic segmentation.

\textbf{Training.} We set $\beta = 5$ in Eq.~\eqref{eq:rho}, $Q_{\text{max}} = 8$, $\lambda_{r}^{(q)} = \{18, 9.32, 4.83, 2.5, 1.3, 0.67, 0.35, 0.18\}$, $\lambda_{G} = 2.56$, $\lambda_{\text{perc}} = 4.26$, $\lambda_{\text{task}} = 1$, and $\lambda_{\text{ratio}} = 10$ for all training stages. $q$ and $\alpha$ are uniformly sampled from $\{1,2,3,4,5,6,7,8\}$ and $\{0,1,2,3,4,5,6,7\}/7$ respectively. The training steps of Stage 1 in Sec.~\ref{sec:training_stage1} are set to 3M, the first half without $\mathcal{L}_{G}$ and the last half with $\mathcal{L}_{G}$. The training steps of Stage 2 are set to 500K. In each training stage, the initial learning rate is set to $10^{-4}$, decayed to $10^{-5}$ for the last 25\% steps. When training $\phi_{\text{perc}}$ and $\phi_{\text{MSE}}$, images are randomly cropped to 256$\times$256$\times$3, and a random horizontal/vertical flip is applied. ConvNeXt-Tiny~\cite{liu2022convnet} is used for training $\phi_{\text{cls}}$, with images randomly resized and cropped to 256$\times$256$\times$3 followed by a random horizontal flip. PSPNet50~\cite{zhao2017pyramid} is used for training $\phi_{\text{seg}}$, with images randomly resized, flipped and cropped to 256$\times$256$\times$3. The batch size is set to 8 for all tasks. Adam~\cite{ADAM} is used for optimization.

\textbf{Evaluation.} For human vision evaluation, we use Peak Signal-to-Noise Ratio (PSNR) to measure distortion, and use Fréchet Inception Distance (FID)~\cite{fid} and LPIPS~\cite{zhang2018unreasonable} to measure realism. We use the same protocol in~\cite{mentzer2020highfidelity, agustsson2023multirealism, ghouse2023residual, iwai2024controlling} to calculate FID, i.e., cropping images to overlapped patches of size 256$\times$256$\times$3. Note that FID is not calculated on Kodak because it yields only 192 patches, which is not sufficient for measuring FID. For the classification task, we use top-1 accuracy (acc.) to present the performance, with images first resized to a short edge of 292 and then center-cropped to 256$\times$256$\times$3. For the segmentation task, we use mean Intersection over Union (mIoU) to present the performance, with images first resized to a short edge of 512 and then center-cropped to 512$\times$512$\times$3.

\begin{figure}[t]
\begin{center}
{
\resizebox{\linewidth}{!}{
\scriptsize
\begin{tabular}{@{\hskip 0mm}l@{\hskip 1.0mm}l@{\hskip 1.0mm}l@{\hskip 1.0mm}l@{\hskip 1.0mm}l@{\hskip 0mm}}

\toprule
Reconstruction &
FullGrad~\cite{srinivas2019fullgradient} &
Score Map (3-rd Stage) &
Score Map (2-nd Stage) &
Score Map (1-st Stage)
\\ \midrule

Cls. Path: 29.41dB
& Confidence: 81.26\%
& \multicolumn{3}{c}{The score maps are inverted for better visualization (higher for the side path).} \\
\includegraphics[width=0.16\linewidth]{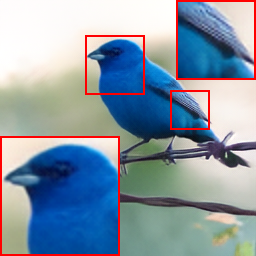} &
\includegraphics[width=0.16\linewidth]{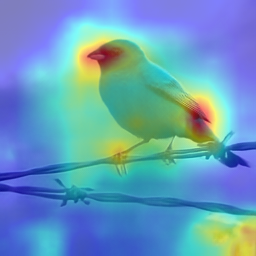} &
\includegraphics[width=0.2\linewidth]{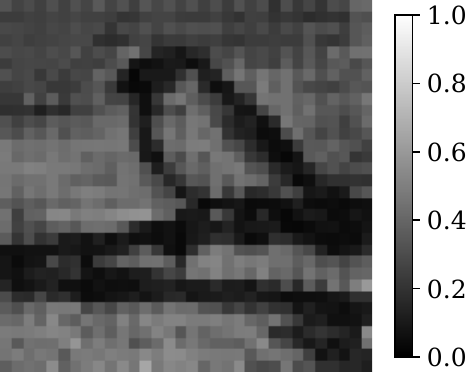} &
\includegraphics[width=0.2\linewidth]{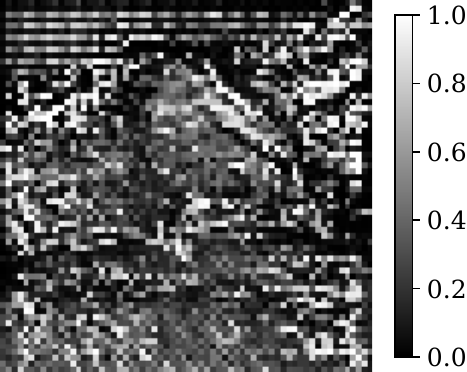} &
\includegraphics[width=0.2\linewidth]{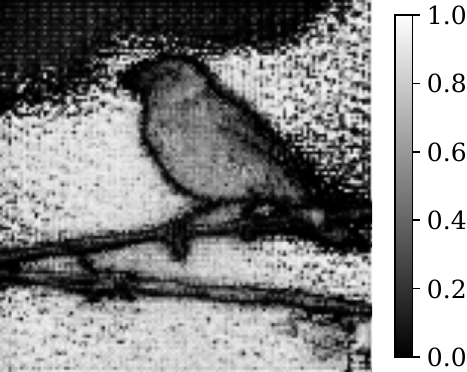} \\
MSE Path: 29.50dB
& Confidence: 37.55\%
& & & \\
\includegraphics[width=0.16\linewidth]{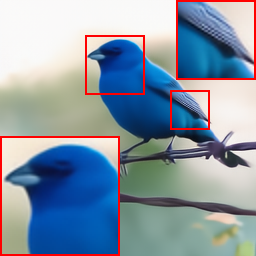} &
\includegraphics[width=0.16\linewidth]{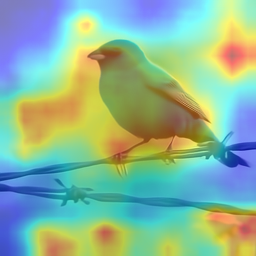} &
\includegraphics[width=0.2\linewidth]{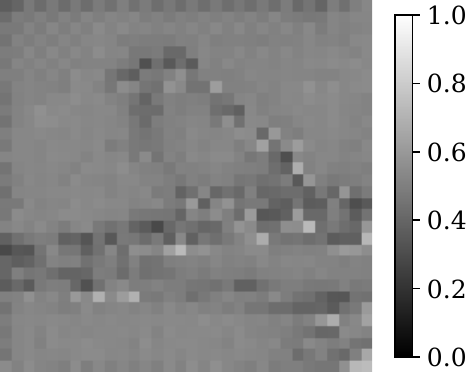} &
\includegraphics[width=0.2\linewidth]{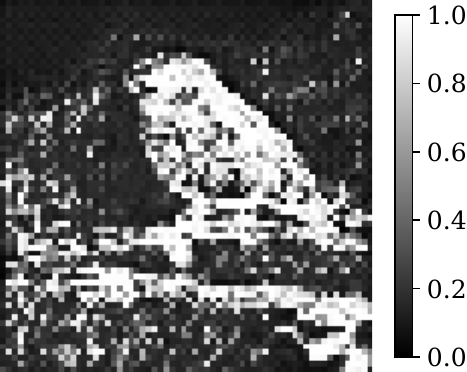} &
\includegraphics[width=0.2\linewidth]{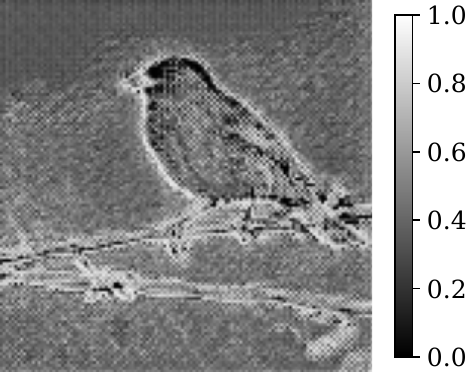} \\
\\[-2ex]\midrule

Perc. Path: 28.65dB
& Confidence: 77.46\%
& Bit Allocation
&
& Original: \texttt{val00035565} \\
\includegraphics[width=0.16\linewidth]{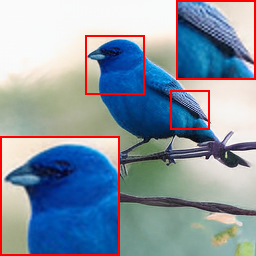} &
\includegraphics[width=0.16\linewidth]{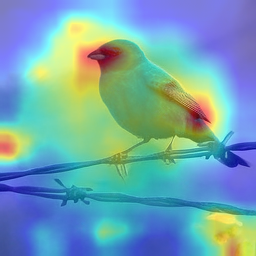} &
\includegraphics[width=0.2\linewidth]{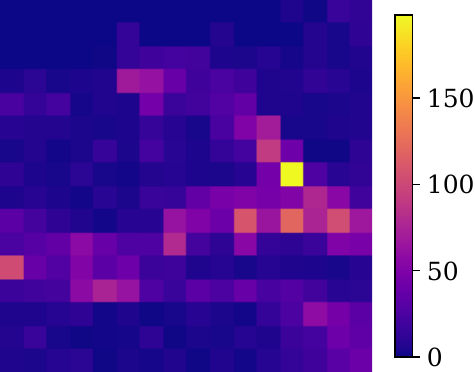} & &
\includegraphics[width=0.16\linewidth]{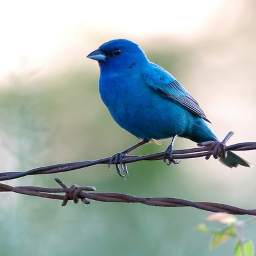} \\ \bottomrule
\end{tabular}
}
}
\caption{Visualization of the reconstructed images, FullGrad~\cite{srinivas2019fullgradient} and score maps. The image is from ImageNet~\cite{imagenet1k} and resized to 256$\times$256$\times$3. The regions with warmer colors in FullGrad have larger gradients, indicating a stronger impact on the classification decision. The bitrate is 0.0888bpp.}
\label{fig:cam}
\end{center}
\end{figure}

\begin{figure}[t]
\begin{center}
{
\resizebox{\linewidth}{!}{
\scriptsize
\begin{tabular}{@{\hskip 0mm}l@{\hskip 1.0mm}l@{\hskip 1.0mm}l@{\hskip 1.0mm}l@{\hskip 1.0mm}l@{\hskip 0mm}}

\toprule
Reconstruction &
Segmentation Map &
Score Map (3-rd Stage) &
Score Map (2-nd Stage) &
Score Map (1-st Stage)
\\ \midrule

Seg. Path: 31.62dB
& & \multicolumn{3}{c}{The score maps are inverted for better visualization (higher for the side path).} \\
\includegraphics[width=0.16\linewidth]{figures/visualization/ade20k/val00000269/val00000269_seg.png} &
\includegraphics[width=0.16\linewidth]{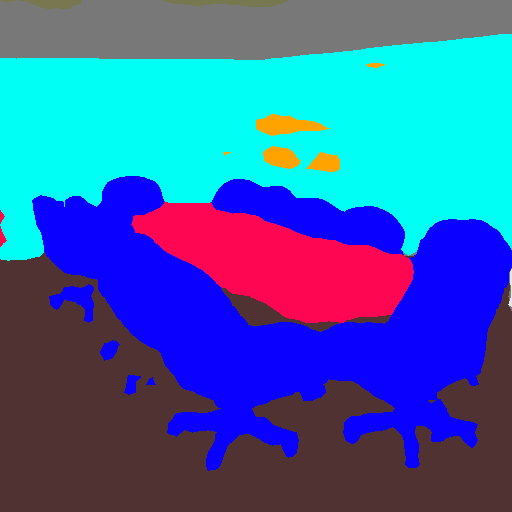} &
\includegraphics[width=0.2\linewidth]{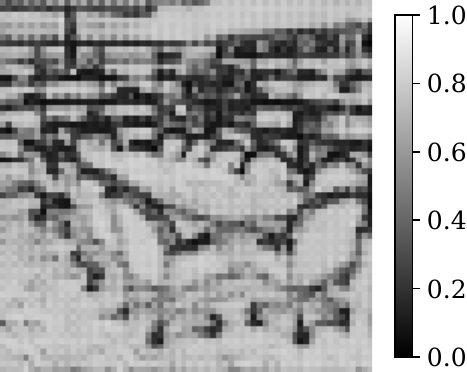} &
\includegraphics[width=0.2\linewidth]{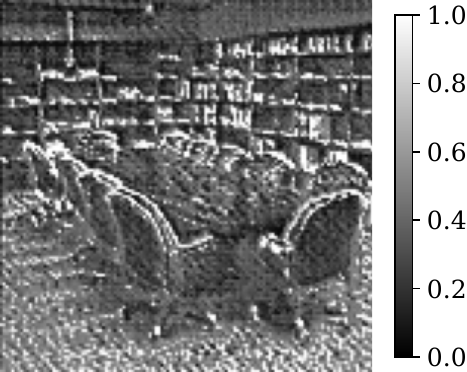} &
\includegraphics[width=0.2\linewidth]{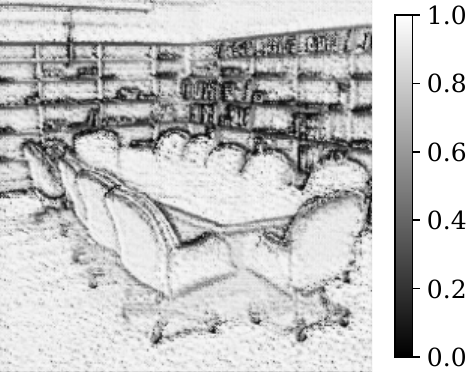} \\
MSE Path: 31.87dB
& & & & \\
\includegraphics[width=0.16\linewidth]{figures/visualization/ade20k/val00000269/val00000269_mse.png} &
\includegraphics[width=0.16\linewidth]{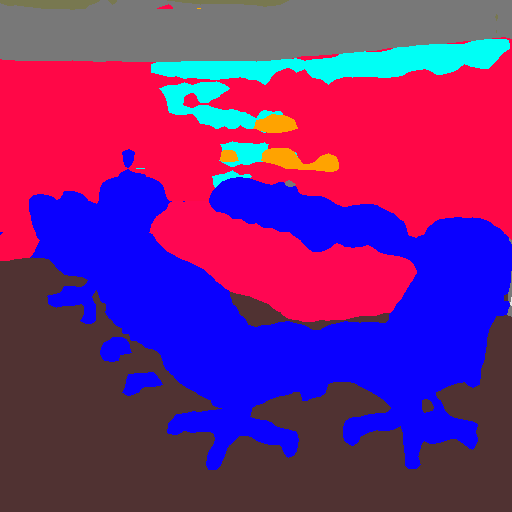} &
\includegraphics[width=0.2\linewidth]{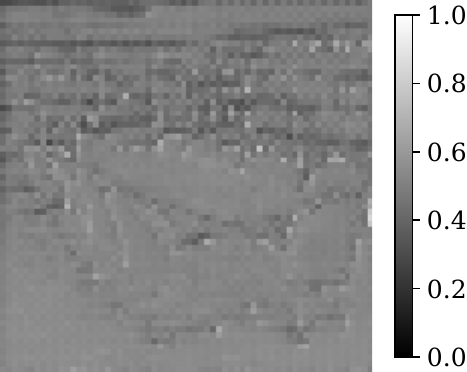} &
\includegraphics[width=0.2\linewidth]{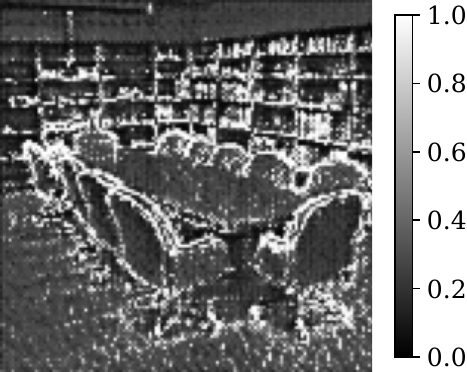} &
\includegraphics[width=0.2\linewidth]{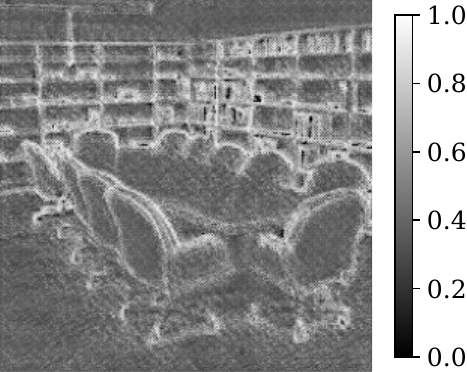} \\
\\[-2ex]\midrule

Perc. Path: 31.30dB
&
& Bit Allocation
& Original: \texttt{val00000269}
& Ground Truth \\
\includegraphics[width=0.16\linewidth]{figures/visualization/ade20k/val00000269/val00000269_perc.png} &
\includegraphics[width=0.16\linewidth]{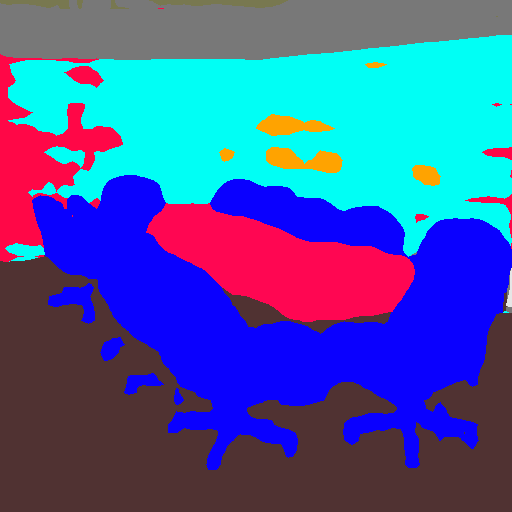} &
\includegraphics[width=0.2\linewidth]{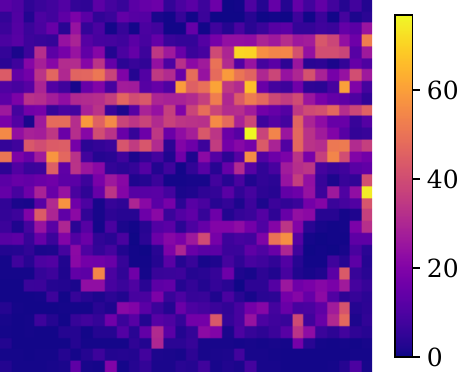} &
\includegraphics[width=0.16\linewidth]{figures/visualization/ade20k/val00000269/val00000269_original.png} &
\includegraphics[width=0.16\linewidth]{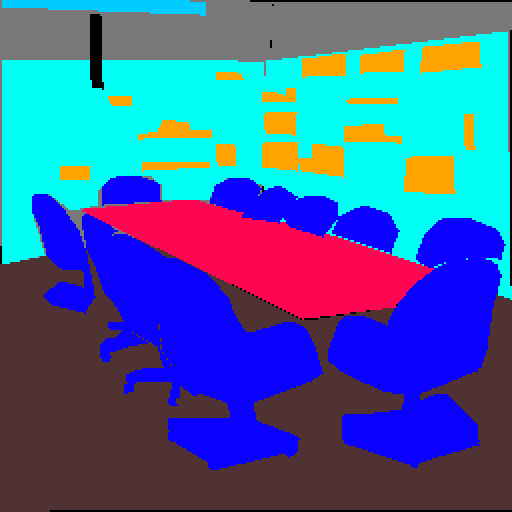} \\ \bottomrule
\end{tabular}
}
}
\caption{Visualization of the reconstructed images, segmentation maps and score maps. The image is from ADE20K~\cite{ade20k} and resized to 512$\times$512$\times$3. The bitrate is 0.0718bpp.}
\label{fig:seg}
\end{center}
\end{figure}

\subsection{Results of multi-task performance}
\label{sec:human_vision}

For human vision, we compare the proposed MPA to the SOTA baselines of the unified model to evaluate R-D and R-P performance, i.e., MRIC~\cite{agustsson2023multirealism}, DIRAC~\cite{ghouse2023residual}, CRDR~\cite{iwai2024controlling}. In addition, we add HiFiC~\cite{mentzer2020highfidelity} and VTM17.1~\cite{VTM_17_1} for comparison as the anchors of fully perceptual optimization and traditional coding method, respectively. For machine vision, the classification task is the understanding of the global semantic information of an image and represents a high-level vision task. In contrast, the segmentation task challenges the model's ability to understand pixel-level semantics and represents a low-level vision task. Thus, we test the reconstructed images on classification and semantic segmentation tasks to comprehensively evaluate the performance of MPA for machine vision. We compare MPA to SOTA baselines with task-specific optimization, SFT~\cite{song2021variablerate} and TransTIC~\cite{chen2023transtic}, along with VTM17.1~\cite{VTM_17_1}.

As shown in Fig.~\ref{fig:performance}, MPA can achieve performance comparable to existing SOTA models. For human vision, the FID of MPA is lower than that of the best-performing model, CRDR, at a low bitrate. Since perceptual performance is more reflective of the human visual system at a low bitrate, this part of the gain is significantly beneficial. Considering the distortion, even with only partial parameters optimized for MSE, MPA still has a wide adjustable range comparable to other fully fine-tuned models, especially at a low bitrate. Note that we achieve such performance with a smaller decoder (9.295M for MPA vs. 13.38M for CRDR), fewer training steps (3.5M steps for MPA vs. 5M steps for CRDR), and partial fine-tuning (7.27\% for MPA and 100\% for CRDR). In terms of machine vision, $\phi_{\text{perc}}$, $\phi_{\text{cls}}$ and $\phi_{\text{seg}}$ perform significantly better than models optimized for MSE, outperforming SFT and TransTIC which are specially optimized for vision tasks. With only 1.89\% parameters fine-tuned for machine vision, MPA can even achieve accuracy comparable to the fully fine-tuned models, showcasing its powerful task transition capability. Note that any point in the colored areas in Fig.~\ref{fig:performance} can be achieved by adjusting $q$ and $\alpha$, as MPA supports continuously variable-rate coding and seamless transitions between all tasks using a unified model.

\textbf{Discussion.} We choose the perceptually optimized path to serve as the main path due to its ability to preserve high-level semantic features that are useful across various tasks. This generalization is achieved by using pre-trained classification models for measurement (VGG~\cite{Simonyan2015} for training and AlexNet~\cite{alex2012imagenet} for testing). This method aligns well with both human and machine perceptions in the latent space~\cite{zhang2018unreasonable}. The pre-trained VGG implicitly incorporates label information, making the distribution of the reconstructed image semantically closer to the original one, thus reducing perceptual distance and enhancing accuracy. Additionally, $\phi_{\text{cls}}$ and $\phi_{\text{seg}}$ use task-specific pre-trained models to compute cross-entropy loss (image-wise for classification and pixel-wise for segmentation), directly targeting task-specific accuracy. These constraints, along with semantic information from the first training stage, allow $\phi_{\text{cls}}$ and $\phi_{\text{seg}}$ to achieve higher accuracy than $\phi_{\text{perc}}$ with lower computational complexity. Moreover, perceptually optimized features can be easily adapted for MSE optimization, achieving higher PSNR. Perceptual loss preserves high-level semantic features, which supports the minimization of pixel-wise distortion when fine-tuned with MSE loss especially at a low bitrate~\cite{duan2023lossy}. Thus, perceptual optimization not only unify human and machine vision tasks but also enhance performance in traditional metrics like PSNR, making it a suitable method for the generalized model.

\subsection{Visualization}

To investigate the effects of different paths, we visualize the qualitative results in Figs.~\ref{fig:cam} and~\ref{fig:seg}. MSE optimization prioritizes image textures, with fine-grained textures scoring higher, while analysis tasks focus on semantic regions. Score prediction differences at each stage highlight distinctions between MSE and machine vision perception. MSE-optimized features show significant score differences at lower-level stages (1st and 2nd), enhancing PSNR by improving textures. Conversely, machine vision-optimized features show significant score differences at all stages, indicating that both high- and low-level semantic features are crucial for analysis tasks. Consequently, images optimized for MSE without semantic supervision are visually smoother at low bitrate but may lose important semantic features, reducing accuracy. Paths optimized under LPIPS supervision retain more semantic information, ensuring the reconstructed image is semantically closer to the original one. By introducing task loss in Eq.~\eqref{eq:loss_rt}, images decoded by $\phi_{\text{cls}}$ and $\phi_{\text{seg}}$ have more task-specific features for classification and segmentation respectively, leading to higher accuracy than $\phi_{\text{perc}}$.

\begin{table}[t]
\begin{center}
\begin{minipage}{.46\linewidth}
\caption{Effects of path complexity}
\begin{center}
\resizebox{\linewidth}{!}{ 
\begin{tabular}{@{}llll@{}}
\toprule
MLP Type        & \begin{tabular}[c]{@{}l@{}}$\phi_{\text{MSE}}$\\(BD-Rate ↓)\end{tabular} & \begin{tabular}[c]{@{}l@{}}$\phi_{\text{cls}}$\\(Acc. ↑)\end{tabular} & \begin{tabular}[c]{@{}l@{}}$\phi_{\text{seg}}$\\(mIoU ↑)\end{tabular} \\ \midrule
Bottleneck      & 19.51\%               & 76.72\%                & 37.41\%              \\
Inv. Bottleneck & 16.04\%               & 77.16\%                & 37.76\%              \\ \bottomrule
\end{tabular}
}
\label{table:ablation_type}
\end{center}
\end{minipage}
\begin{minipage}{.53\linewidth}
\caption{Cross-validations on path choices}
\begin{center}
\resizebox{\linewidth}{!}{ 
\begin{tabular}{@{}lllll@{}}
\toprule
Task (Metric)   & $\phi_{\text{perc}}$ & $\phi_{\text{MSE}}$ & $\phi_{\text{cls}}$ & $\phi_{\text{seg}}$ \\ \midrule
MSE (BD-Rate ↓)     & 49.61\%              & \textbf{16.04\%}    & 32.81\%            & 34.19\%            \\
Cls. (Acc. ↑) & 76.66\%              & 60.59\%             & \textbf{76.77\%}    & 73.57\%             \\
Seg. (mIoU ↑)      & 36.17\%              & 28.34\%             & 35.34\%             & \textbf{37.41\%}    \\ \bottomrule
\end{tabular}
}
\label{table:ablation_path}
\end{center}
\end{minipage}
\end{center}
\end{table}

\begin{table}[t]
\begin{center}
\begin{minipage}{.35\linewidth}
\caption{Ablations on encoder}
\begin{center}
\resizebox{\linewidth}{!}{ 
\begin{tabular}{@{}ll@{}}
\toprule
Components    & \begin{tabular}[c]{@{}l@{}}BD-Rate ↓\\ against VTM\end{tabular} \\ \midrule
Full MPA      & 16.04\% \\
\hspace{2mm} w/o Predictors & 16.25\% \\
\hspace{2mm} w/o $\phi_{\text{hq}}$   & 17.05\% \\
\hspace{2mm} w/o $\phi_{\text{lq}}$   & 17.18\% \\ \bottomrule
\end{tabular}
}
\label{table:ablation_encoder}
\end{center}
\end{minipage}
\begin{minipage}{.635\linewidth}
\caption{Comparison of complexity}
\begin{center}
\resizebox{\linewidth}{!}{ 
\begin{tabular}{@{}llll@{}}
\toprule
Models                                                           & \#Param. & KFLOPs per pixel   & Latency (ms)     \\ \midrule
\textcolor{gray}{MRIC~\cite{agustsson2023multirealism}} & \textcolor{gray}{69.14M} & \textcolor{gray}{1118.17} & \textcolor{gray}{11.89} \\
TinyLIC~\cite{lu2022highefficiency}                             & 28.46M   & 439.29             & 12.68            \\
\hspace{2mm} + MLPs                                    & +0.51M$\sim$+2.04M & -56.68$\sim$+0     & -0.33$\sim$+0    \\
\hspace{2mm} + Predictors                                        & +0.03M   & +2.23              & +0.09            \\
\hspace{2mm} + \textcircled{\small{S}} \& \textcircled{\small{A}} & +0      & +0                 & +0.28            \\ \midrule
MPA                                                    & 29.00M$\sim$30.53M & 384.84$\sim$441.52 & 12.72$\sim$13.05 \\ \bottomrule
\end{tabular}
}
\label{table:complexity}
\end{center}
\end{minipage}
\end{center}
\end{table}

\subsection{Diving into MPA}
\label{sec:dive_into}

To further explore the configuration of the MPA, we perform a series of evaluations. Kodak~\cite{KODAK}, ImageNet-1K~\cite{imagenet1k} and ADE20K~\cite{ade20k} are used for evaluate BD-rate~\cite{bjontegaard2001calculation}, top-1 accuracy and mIoU respectively. BD-rate is computed over the whole R-D curve. Top-1 accuracy and mIoU are computed at the lowest bitrate, i.e., 0.1521bpp on ImageNet-1K and 0.0948bpp on ADE20K. For the effects of complexity, Table~\ref{table:ablation_type} reveals that human vision-oriented tasks are more sensitive to the complexity of the paths, which suggests the use of inverted bottleneck MLPs, while machine vision-oriented paths can use bottleneck MLPs because the complexity has relatively less impact on the accuracy. As for the choice of paths, we conduct cross-validations as shown in Table~\ref{table:ablation_path}, which demonstrates that the paths do learn the task-specific features to make their corresponding tasks perform optimally. To evaluate the effect of MPA in the encoder, we conduct ablations on each MPA component. We replace the predicted mask with a random mask and disable $\phi_{\text{hq}}$ and $\phi_{\text{lq}}$ during encoding. The results in Table~\ref{table:ablation_encoder} demonstrate that the predictors capture different features critical to high and low bitrate compression, and $\phi_{\text{hq}}$ and $\phi_{\text{lq}}$ are specialized to corresponding features respectively.

\begin{wrapfigure}[14]{r}{0.4\textwidth}
\vspace{-2.2em}
\begin{minipage}{0.4\textwidth}
\begin{figure}[H]
\begin{center}
\resizebox{\linewidth}{!}{ 
\includegraphics{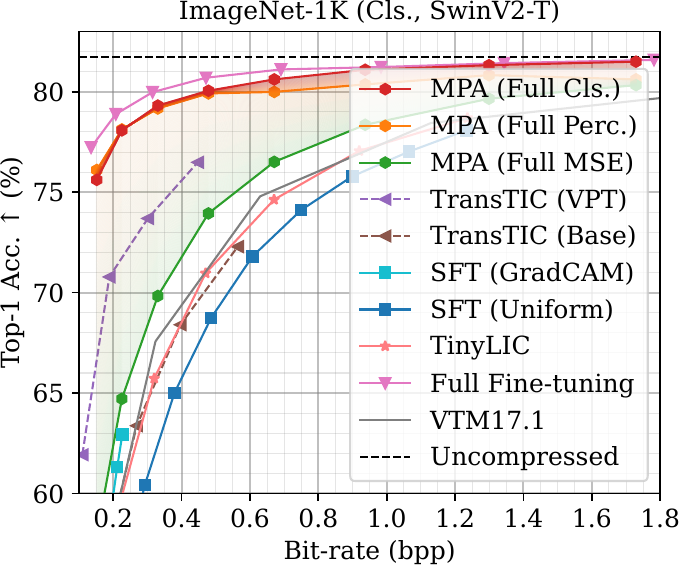}
}
\caption{Validation on SwinV2-T~\cite{liu2022swin}}
\label{fig:swinv2_t}
\end{center}
\end{figure}
\end{minipage}
\end{wrapfigure}

To evaluate the generalization of the learned features for the same task, we test $\phi_{\text{cls}}$ on Swin Transformer V2~\cite{liu2022swin}. Fig.~\ref{fig:swinv2_t} shows that $\phi_{\text{cls}}$ has a similar performance gain comparable to that on ConvNeXt~\cite{liu2022convnet} which is originally used for training. Note that $\phi_{\text{cls}}$ is only trained once without fine-tuning on SwinV2, reflecting the generalization of the learned features in MPA.

The complexity of MPA is shown in Table~\ref{table:complexity}. Here, the input size is 256$\times$256$\times$3. Latency is averaged over a batch of 16 images running 5000 times on a RTX3090 GPU through an encoder, a hyperprior entropy model, and a decoder. With the negligible computational overhead, MPA achieves comparable performance to the full fine-tuning model and has seamless transitions between tasks. The main limitations of MPA are its increased latency compared to the baseline and the need for separate training for each task. The former can be mitigated by developing a specialized operator to eliminate the frequent flattening, selection, and reassembly operations, while the latter can be addressed by leveraging multi-task learning techniques~\cite{vandenhende2021multitask}, which will be explored in future work.

\section{Conclusion}

In this paper, we propose Multi-Path Aggregation (MPA) to tackle the problem of coding for multi-task applications in an all-in-one manner. By aggregating task-specific paths, MPA can support a variety of tasks with seamless transitions between them for joint human-machine vision using a unified model and a unified representation. Merely fine-tuning as low as 1.89\% additional parameters, MPA can achieve comparable performance to that of separable models optimized using dedicated criteria, showcasing its strong versatility and scalability. Future work will explore joint optimization of multiple paths to further unify training process and improve multi-task performance.

\begin{ack}
This work was supported in part by Natural Science Foundation of Jiangsu Province under Grant BK20241226, and Jiangsu Provincial Key Research and Development Program under Grant BE2022155. The authors would like to express their sincere gratitude to the Interdisciplinary Research Center for Future Intelligent Chips (Chip-X) and Yachen Foundation for their invaluable support.
\end{ack}

{
\small

\bibliographystyle{plain}
\bibliography{references}
}


\clearpage
\newpage
\appendix

\section*{Appendix}

\section{Broader impacts}
\label{sec:impacts}

MPA provides a flexible and task-oriented optimization approach for a unified coding framework. By leveraging the different importance of features for different tasks and feature correlations across tasks, MPA can empower existing learned image compression models with the ability to reconstruct images with various orientations based on the needs of users or machines (such as fidelity and accuracy), significantly enhancing their versatility. This research not only proposes and sets a new baseline for multi-task coding but also offers a new avenue for the exploration of joint human-machine vision. By exploiting generalized features and few-step fine-tuning for a tiny proportion of parameters, MPA enables joint human-machine vision for multiple tasks in an all-in-one manner, without massive handcrafted techniques. This advancement represents a significant step toward creating a versatile coding framework that can adapt to a wide range of applications and requirements. This can have a positive societal impact by saving a massive amount of storage space, as only a unified model and a unified representation need to be stored rather than multiple coding models and representations. By reducing storage requirements, it can lower the cost of data storage and transmission, making it more accessible for a wide range of applications. Additionally, this reduction in storage can lead to more efficient use of resources and energy, contributing to environmental sustainability. Lower storage costs and improved resource efficiency can also facilitate broader adoption of advanced coding technologies in various industries, enhancing scalability and operational efficiency. Although there are limitations in scenarios requiring extremely high fidelity or accuracy, which may lead to potential negative societal impacts such as misjudgments on reconstructed images, these issues can be addressed by future work aimed at performance improvements or the development of solutions tailored to specific scenarios.

\section{Implementation details}
\label{sec:impl}

\subsection{Architecture}
\label{sec:arch}

\begin{figure}[htbp]
\begin{center}
\includegraphics[width=\textwidth]{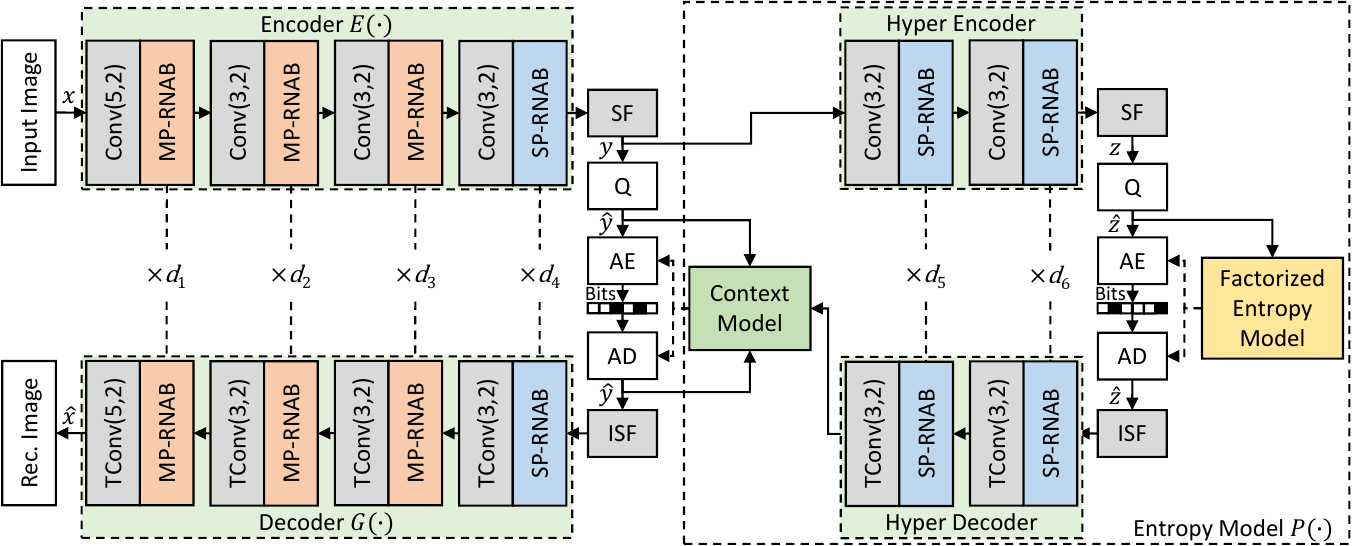}\hfill
\caption{MPA in TinyLIC~\cite{lu2022highefficiency}. $d_i$ is the number of RNABs used at $i$-th stage. Convolution Conv($k$,$s$) and its transposed version TConv($k$,$s$) apply the kernel at a size of $k \times k$ and a stride of $s$. Uniform Quantization is used in Q; AE and AD stand for respective Arithmetic Encoding and Decoding.}
\label{fig:architecture}
\end{center}
\end{figure}

Our implementation is based on TinyLIC~\cite{lu2022highefficiency}. The multi-path mechanism is fulfilled by replacing the default Single-Path Residual Neighborhood Attention Block (SP-RNAB) with Multi-Path RNAB (MP-RNAB, i.e., MP-Block in Fig.~\ref{fig:MPA}) at the first three stages of the main encoder and decoder (see Fig.~\ref{fig:architecture}). To build a continuously variable-rate baseline, we introduce the scaling factor (SF) modulation $\bm{s}_q$~\cite{Chen2020} and non-linear interpolation~\cite{cui2021asymmetric, lee2022selective} as Eq.~\eqref{eq:scaling_factor}. The SF layer and its inverse operation (ISF) are embedded for the latent representation and each Transformer Block. \textit{Other settings are consistent with TinyLIC.}
\begin{equation}
\label{eq:scaling_factor}
\bm{s}_q=
\begin{cases}
  \bm{s}_q, & \text{if}\ q \in \{1,2,...,Q_{\text{max}}\}, \\
  {\bm{s}_{\floor q}^ {1-(q-{\floor q})}}  \cdot {\bm{s}_{\ceil q}^ {q-{\floor q}}}, & \text{otherwise}.
\end{cases}
\end{equation}

\begin{figure}[htbp]
\begin{center}
\subfloat[Bottleneck MLP]{\includegraphics[scale=0.7]{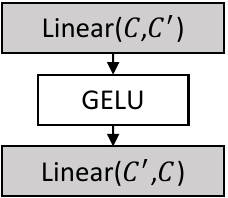}\label{subfig:bottleneck}}\hspace{30pt}
\subfloat[Inverted Bottleneck MLP]{\includegraphics[scale=0.7]{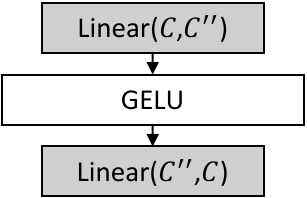}\label{subfig:inv_bottleneck}}\hfill
\caption{Two types of MLP. $\text{Linear}(C, C')$ denotes a fully-connected layer with $C$ input channels and $C'$ output channels. We set $C' = C / 2$ and $C'' = 2C$ in our implementation.}
\label{fig:mlp}
\end{center}
\end{figure}

In terms of path configuration, we use inverted bottleneck MLPs for human vision (i.e., $\phi_{\text{perc}}$ and $\phi_{\text{MSE}}$) and bottleneck MLPs for machine vision (i.e., $\phi_{\text{cls}}$ and $\phi_{\text{seg}}$), as shown in Fig~\ref{fig:mlp}. The bottleneck MLP is used to address low complexity demands, while the inverted bottleneck MLP is used to address high complexity demands. Specifically, the bottleneck MLP first down-samples input channels to $C'$ followed by GELU activation, then up-samples it to $C$, while the inverted one first up-samples $C$ to $C''$ with activation and then down-samples it to $C$. In our implementation, we set $C'' = 2C$ as TinyLIC~\cite{lu2022highefficiency} does, and set $C' = C / 2$ to align the up/down-sampling ratio.

\subsection{MPA in encoder}
\label{sec:encoder}

\begin{figure}[htbp]
\begin{center}
\subfloat{\includegraphics[scale=0.5]{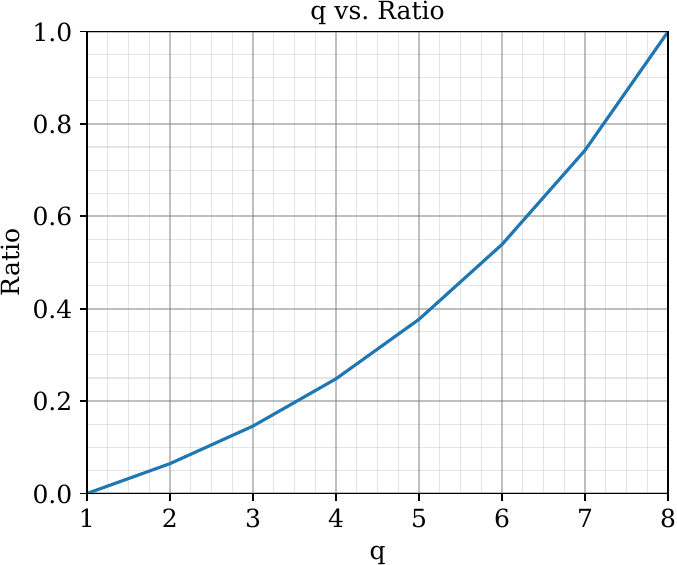}\label{subfig:q_vs_ratio}}\hspace{10pt}
\subfloat{\includegraphics[scale=0.5]{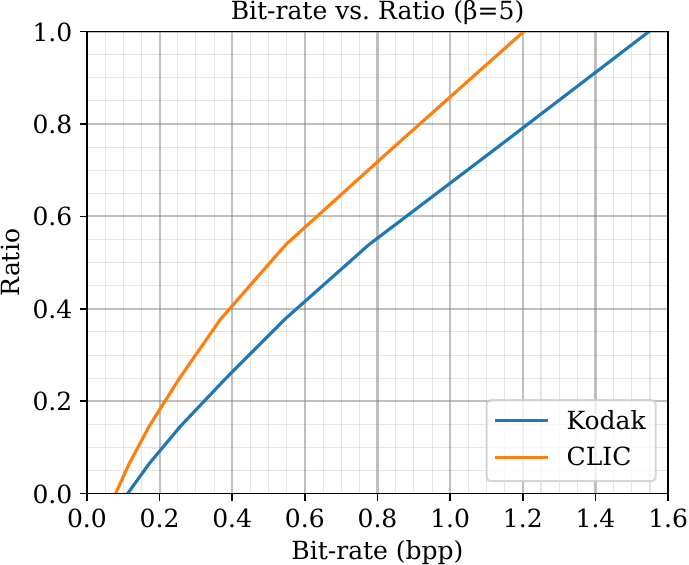}\label{subfig:bitrate_vs_ratio}}\hfill
\caption{The relationship between $q$ and the aggregation ratio $\rho_{\text{enc}}$.}
\label{fig:ratio}
\end{center}
\end{figure}

As we mentioned in Sec.~\ref{sec:mpa}, we also implement MPA in the encoder, i.e., $\phi_{\text{hq}}$ and $\phi_{\text{lq}}$. Setting $\beta = 5$ in Eq.~\ref{eq:rho}, the relationship between $q$ and the aggregation ratio $\rho_{\text{enc}}$ is visualized in Fig.~\ref{fig:ratio}, which fits an approximate linear relationship between bitrate and $\rho_{\text{enc}}$ on both Kodak~\cite{KODAK} and CLIC~\cite{CLIC2020}.

\subsection{Reproducibility}
\label{sec:reproducibility}

All experiments are conducted on a server with two Intel Xeon Silver 4210 CPUs and a single NVIDIA RTX3090 GPU. Our implementation relies on the open-source repository of TinyLIC~\cite{lu2022highefficiency}, which is based on PyTorch~\cite{pytorch} and CompressAI~\cite{begaint2020compressai}. The open-source baselines, i.e., SFT~\cite{song2021variablerate} and TransTIC~\cite{chen2023transtic}, are re-tested using their public repository. For the comparison to baselines without public code and weights, we align our experimental setup with theirs. ConvNeXt Tiny~\cite{liu2022convnet} and Swin Transformer V2 Tiny~\cite{liu2022swin} are created using timm~\cite{rw2019timm} with officially published pre-trained weights. PSPNet50~\cite{liu2022convnet} and its pre-trained weights are obtained from the authors' official repository. FullGrad~\cite{srinivas2019fullgradient} is visualized using an open-source PyTorch library~\cite{jacobgilpytorchcam}.

\section{Comparison to other related work}

Mixture-of-Experts (MoE)~\cite{jacobs1991adaptive, shazeer2017} enables Transformer models~\cite{vaswani2017attention, dosovitskiy2021vit} to have non-uniform computing patterns, improving downstream task performance~\cite{lepikhin2021gshard, du2022glam, fedus2022switch, xue2022go}, gaining multi-modality modelling capabilities~\cite{bao2022vlmo, wang2022image}, or enabling multi-task learning~\cite{chen2023adamvmoe}. They replace the Feed-Forward Network (FFN) with multiple experts and routing tokens via gating networks to the appropriate experts for inference. Unfortunately, the strengths of MoE are most predominantly represented in large language models due to the scaling law~\cite{kaplan2020scaling}, which is not suitable for storage- and complexity-sensitive scenarios. In comparison, MPA is quite different from the common practice of MoE, as listed in Table~\ref{table:mpa_vs_moe}.

\begin{table}[htbp]
\caption{Comparison between MPA and MoE}
\begin{center}
\begin{tabular}{@{}p{0.14\linewidth}p{0.395\linewidth}p{0.395\linewidth}@{}}
\toprule
\textbf{Aspect} & \textbf{MPA} & \textbf{MoE} \\ \midrule
\textbf{Architecture} & Unified all-in-one model with multiple task-specific paths & Multiple expert models selected by a gating network \\ \midrule
\textbf{Optimization} & Two-stage optimization with partial parameter fine-tuning & Joint optimization with expert selection via gating network \\ \midrule
\textbf{Routing} & Leverages feature importance and correlations across tasks & Utilizes the expertise of different experts for specific inputs \\ \midrule
\textbf{Efficiency} & Designed specifically for storage- and computation-sensitive multi-task scenarios & Specially used in large models to significantly increase parameter count for performance boost \\ \midrule
\textbf{Usage} & User-defined selection of task paths & Expert allocation entirely decided by the gating network \\ \bottomrule
\end{tabular}
\label{table:mpa_vs_moe}
\end{center}
\end{table}

Low-Rank Adaptation (LoRA)~\cite{hu2022lora} is a technique that reduces the number of trainable parameters in large pre-trained models by factorizing weight matrices into low-rank forms, enabling efficient fine-tuning with minimal computational cost while maintaining performance across various tasks. Our experiments recognize the effectiveness of LoRA in Table~\ref{table:mpa_vs_lora}. Although MPA involves more fine-tuning parameters, it also achieves better performance. Moreover, our work has distinct features. We utilize predictors to support coding for multi-task applications and smooth transitions between tasks within an all-in-one framework. The low-rank structure design of LoRA inspires the future work to consider improvements for the side path.

\begin{table}[htbp]
\caption{R-D performance comparison between MPA and LoRA. The ranks $r$ of LoRA are set to 64, 16, and 4, respectively, and the \texttt{lora\_alpha} (NOT the $\alpha$ in MPA) is fixed to 1. We optimize MSE paths for low distortion. Other experimental settings are consistent with those in Sec.~\ref{sec:training}.}
\begin{center}
\begin{tabular}{@{}lllll@{}}
\toprule
\multirow{2}{*}{Dataset} & \multicolumn{4}{c}{BD-Rate against VTM ↓}                                    \\ \cmidrule(lr){2-5}
                         & MPA (7.27\% ft.) & LoRA (3.04\% ft.) & LoRA (0.83\% ft.) & LoRA (0.28\% ft.) \\ \midrule
Kodak~\cite{KODAK}      & 16.04\%          & 16.26\%           & 16.39\%           & 16.83\%           \\
CLIC~\cite{CLIC2020}    & 21.43\%          & 21.80\%           & 21.95\%           & 22.59\%           \\ \bottomrule
\end{tabular}
\label{table:mpa_vs_lora}
\end{center}
\end{table}

\section{Loss functions}

Our loss functions follow the same forms and hyperparameters as those in~\cite{agustsson2023multirealism}, which have been thoroughly evaluated. For the task loss $\mathcal{L}_{\text{task}}$ we propose, we provide detailed ablation studies in Table~\ref{table:ablation_loss} to demonstrate that our current combination achieves a competitive trade-off. Regarding the role of each term, LPIPS enriches the semantic information of the reconstructed images, improving generalization; MSE loss enforces pixel value consistency between the reconstructed and original images; and the cross-entropy loss directly optimizes classification accuracy.

\begin{table}[htbp]
\caption{Ablations on loss terms in $\mathcal{L}_{\text{task}}$. We use the same experimental settings as in Sec.~\ref{sec:training}, and re-train the classification path for each case. Metrics are evaluated at 0.1521bpp on ImageNet-1K~\cite{imagenet1k}. The top-3 results are underlined. $\mathcal{L}_{\text{MSE}}$, $\mathcal{L}_{\text{perc}}$ and $\mathcal{L}_{\text{ce}}$ denote MSE loss, perceptual loss and cross-entropy loss respectively.}
\begin{center}
\begin{tabular}{@{}llllllll@{}}
\toprule
No. & $\mathcal{L}_{\text{MSE}}$ & $\mathcal{L}_{\text{perc}}$ & $\mathcal{L}_{\text{ce}}$ & PSNR ↑ & LPIPS ↓ & \begin{tabular}[c]{@{}l@{}}Top-1 Acc. ↑\\ on ConvNeXt-T~\cite{liu2022convnet}\end{tabular} & \begin{tabular}[c]{@{}l@{}}Top-1 Acc. ↑\\ on SwinV2-T~\cite{liu2022swin}\end{tabular} \\ \midrule
1   & \checkmark &            &            & \underline{26.7dB} & 0.330             & 62.22\%              & 59.27\%             \\
2   &            & \checkmark &            & 25.7dB             & \underline{0.229} & 74.21\%              & 72.25\%             \\
3   &            &            & \checkmark & 24.2dB             & 0.305             & \underline{76.79\%}  & \underline{75.16\%} \\
4   &            & \checkmark & \checkmark & 25.7dB             & \underline{0.244} & \underline{77.01\%}  & \underline{75.96\%} \\
5   & \checkmark &            & \checkmark & \underline{26.5dB} & 0.275             & 76.17\%              & 74.90\%             \\
6   & \checkmark & \checkmark &            & \underline{26.6dB} & \underline{0.238} & 74.28\%              & 73.01\%             \\
7   & \checkmark & \checkmark & \checkmark & \underline{26.5dB} & \underline{0.244} & \underline{76.77\%}  & \underline{75.61\%} \\ \bottomrule
\end{tabular}
\label{table:ablation_loss}
\end{center}
\end{table}

\section{Licenses for released assets}
\label{sec:licenses}

We list the licenses for the released assets we use in Table~\ref{table:licenses}.

\begin{table}[htbp]
\caption{Licenses for released assets}
\begin{center}
\begin{tabular}{@{}p{0.4\linewidth}p{0.45\linewidth}@{}}
\toprule
\textbf{Asset} & \textbf{License} \\ \midrule
TinyLIC~\cite{lu2022highefficiency} & Apache-2.0 license \\ \midrule
TransTIC~\cite{chen2023transtic} & Apache-2.0 license \\ \midrule
SFT~\cite{song2021variablerate} & Not specified \\ \midrule
ConvNeXt~\cite{liu2022convnet} & MIT license \\ \midrule
PSPNet~\cite{zhao2017pyramid} & MIT license \\ \midrule
Swin Transformer V2~\cite{liu2022swin} & MIT license \\ \midrule
ResNet~\cite{he2016deep} (in TorchVision~\cite{torchvision2016}) & BSD-3-Clause license \\ \midrule
Faster R-CNN~\cite{ren2017faster} (in Detectron2~\cite{wu2019detectron2}) & Apache-2.0 license \\ \midrule
VTM17.1~\cite{VTM_17_1} & BSD license \\ \midrule
Flicker2W~\cite{flicker2w} & Custom (research-only) \\ \midrule
DIV2K~\cite{div2k} & Custom (research-only) \\ \midrule
CLIC~\cite{CLIC2020} & Not specified \\ \midrule
Kodak~\cite{KODAK} & Not specified \\ \midrule
ImageNet-1K~\cite{imagenet1k} & Custom (research-only, non-commercial) \\ \midrule
ADE20K~\cite{ade20k} & Custom (research-only, non-commercial) \\ \midrule
MS-COCO 2017~\cite{coco} & Custom \\ \bottomrule
\end{tabular}
\label{table:licenses}
\end{center}
\end{table}

\clearpage
\newpage

\section{More results}

\subsection{Quantitative results}

\begin{figure}[htbp]
\begin{center}
\subfloat{\includegraphics[scale=0.5]{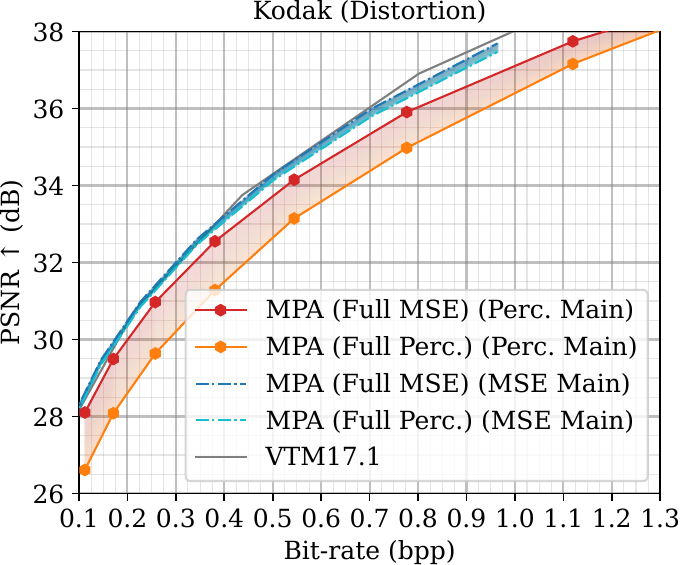}\label{subfig:kodak_rd_mse_main}}
\subfloat{\includegraphics[scale=0.5]{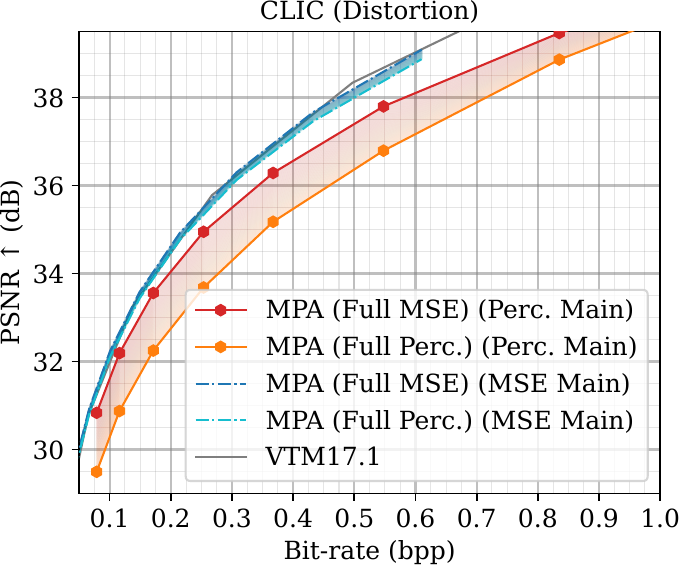}\label{subfig:clic_rd_mse_main}}\vspace{-8pt}\\
\subfloat{\includegraphics[scale=0.5]{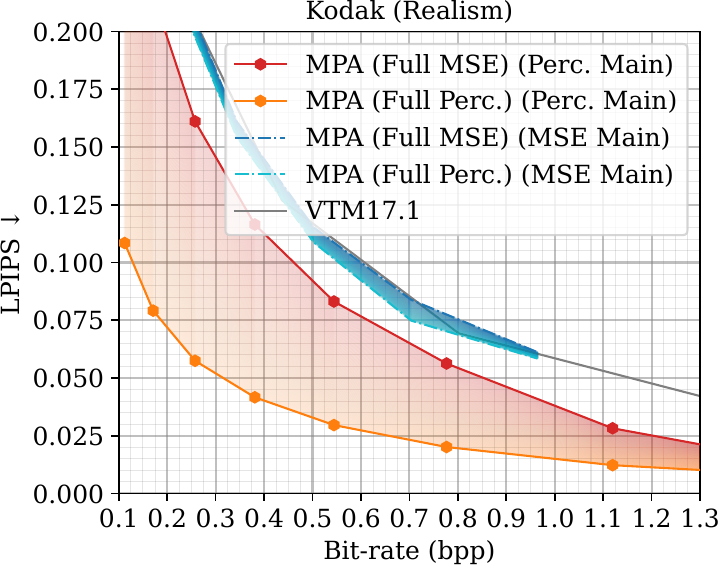}\label{subfig:kodak_lpips_mse_main}}
\subfloat{\includegraphics[scale=0.5]{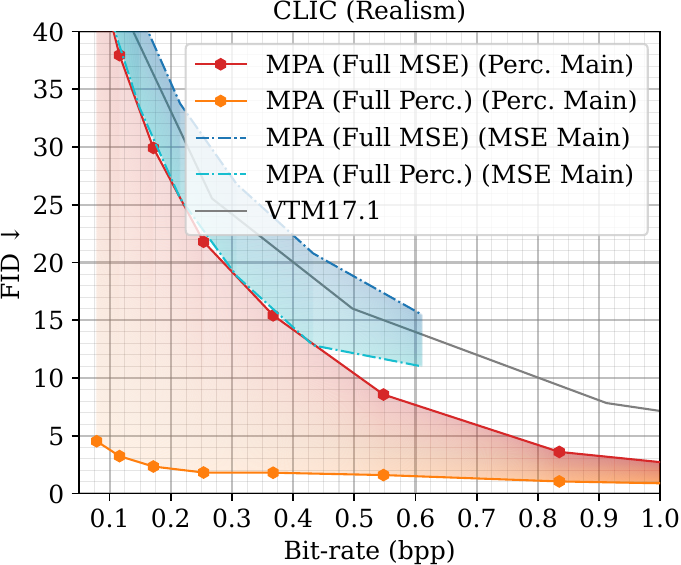}\label{subfig:clic_fid_mse_main}}\hfill
\caption{Multi-task performance with optimizing MSE path as the main path. The results reveal that perceptual optimization is more generalized than MSE optimization for multi-task applications. Although MSE optimization achieves lower distortion, its perceptual quality is more severely degraded compared to that of perceptual optimization, with a significantly narrower adjustable range.}
\label{fig:performance_mse_main}
\end{center}
\end{figure}

\begin{figure}[htbp]
\begin{center}
\subfloat{\includegraphics[scale=0.39]{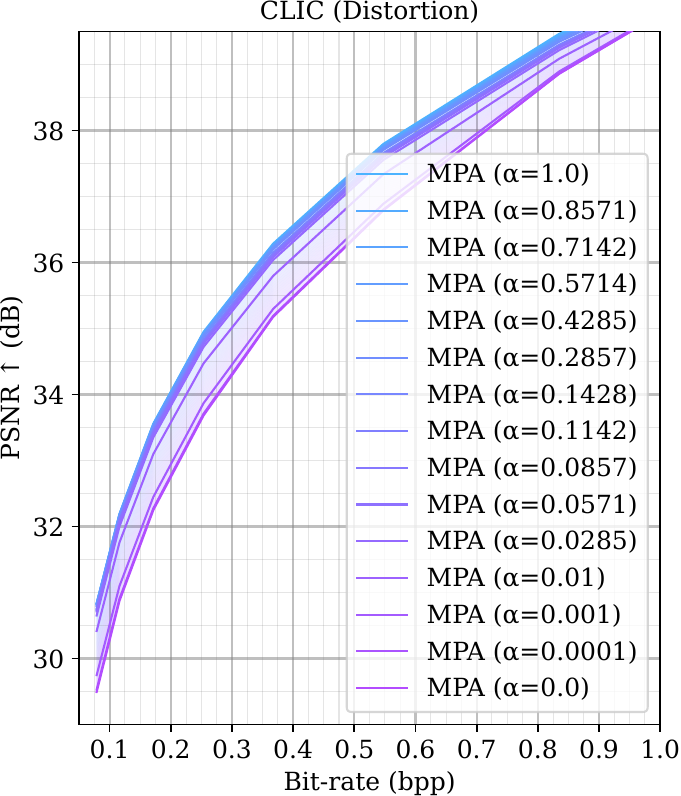}\label{subfig:clic_rd_alpha}}
\subfloat{\includegraphics[scale=0.39]{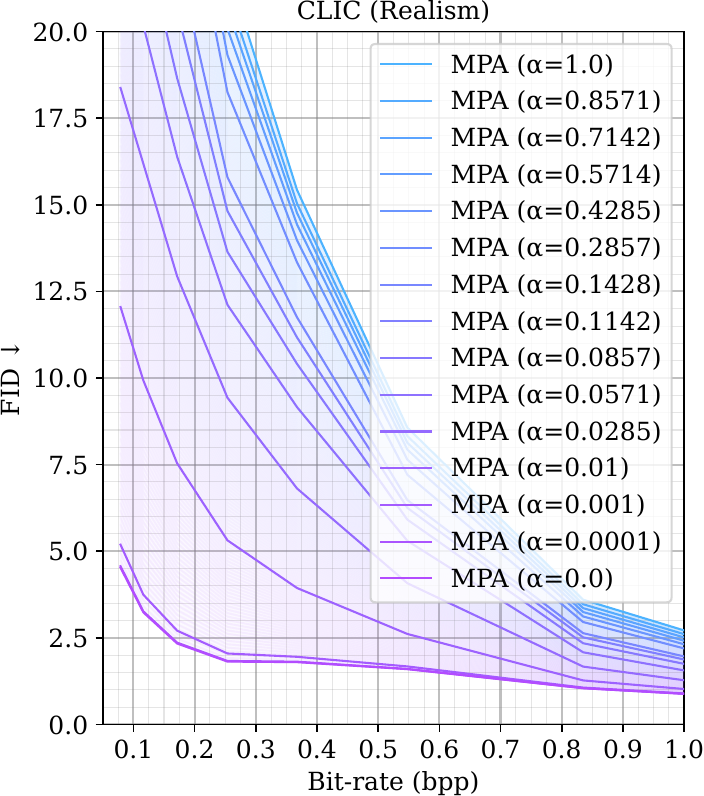}\label{subfig:clic_fid_alpha}}
\subfloat{\includegraphics[scale=0.39]{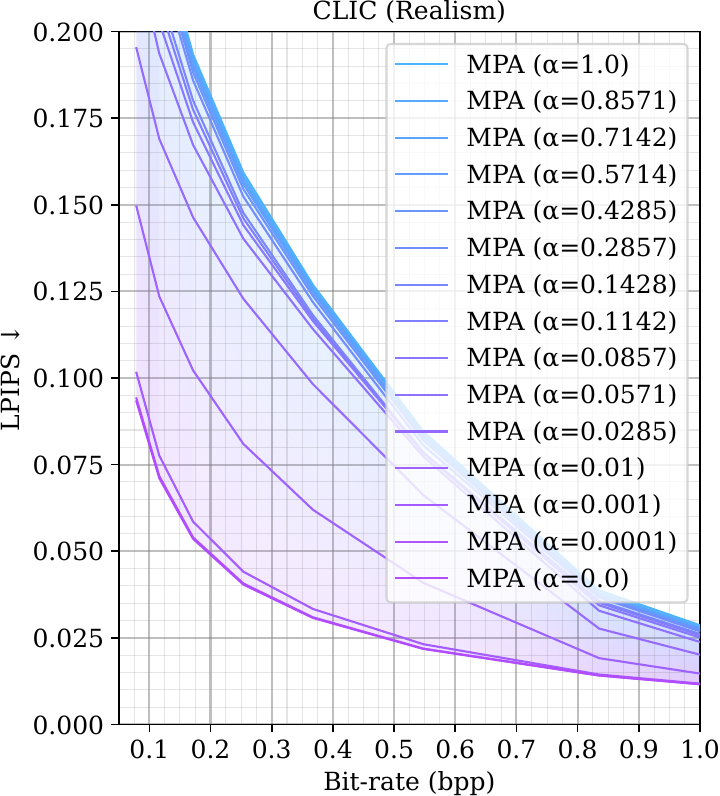}\label{subfig:clic_lpips_alpha}}\hfill
\caption{R-D and R-P performance of MPA with variable $\alpha$ on CLIC dataset~\cite{CLIC2020}.}
\label{fig:performance_clic_alpha}
\end{center}
\end{figure}

\begin{figure}[htbp]
\begin{center}
\subfloat{\includegraphics[scale=0.5]{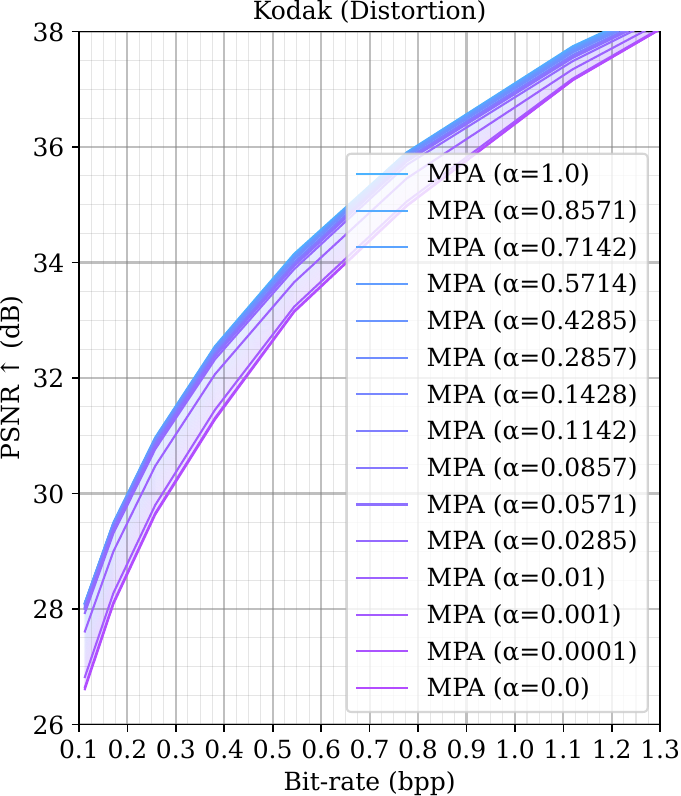}\label{subfig:kodak_rd_alpha}}\hspace{10pt}
\subfloat{\includegraphics[scale=0.5]{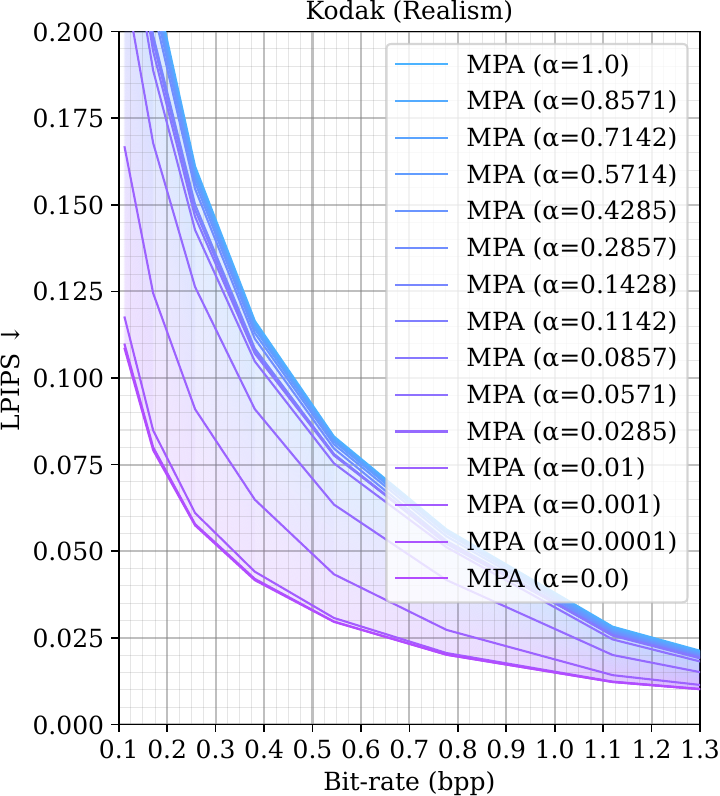}\label{subfig:kodak_lpips_alpha}}\hfill
\caption{R-D and R-P performance of MPA with variable $\alpha$ on Kodak dataset~\cite{KODAK}.}
\label{fig:performance_kodak_alpha}
\end{center}
\end{figure}

\begin{figure}[htbp]
\begin{center}
\subfloat{\includegraphics[scale=0.5]{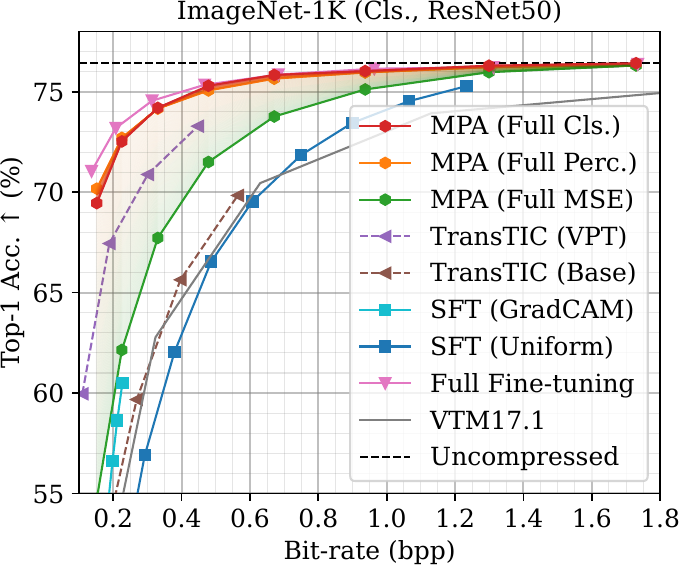}\label{subfig:tv_resnet50}}\hspace{10pt}
\subfloat{\includegraphics[scale=0.5]{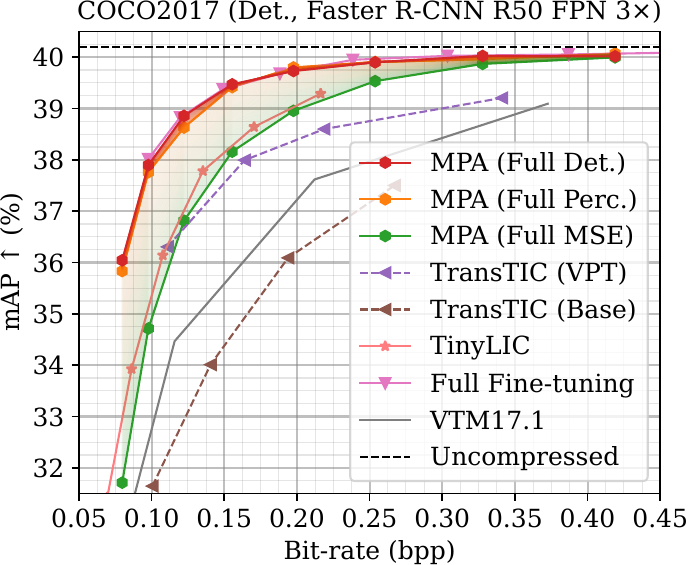}\label{subfig:fasterrcnn}}\hfill
\caption{Classification and objection detection performance of MPA on ImageNet-1K~\cite{imagenet1k} and MS-COCO 2017~\cite{coco}. ResNet50~\cite{he2016deep} and Faster R-CNN R50~\cite{ren2017faster} are used for inference. Top-1 accuracy and mAP (mean Average Precision) are used as metrics for comparison.}
\label{fig:performance_cls_det}
\end{center}
\end{figure}

\begin{figure}[htbp]
\begin{center}
\subfloat{\includegraphics[scale=0.5]{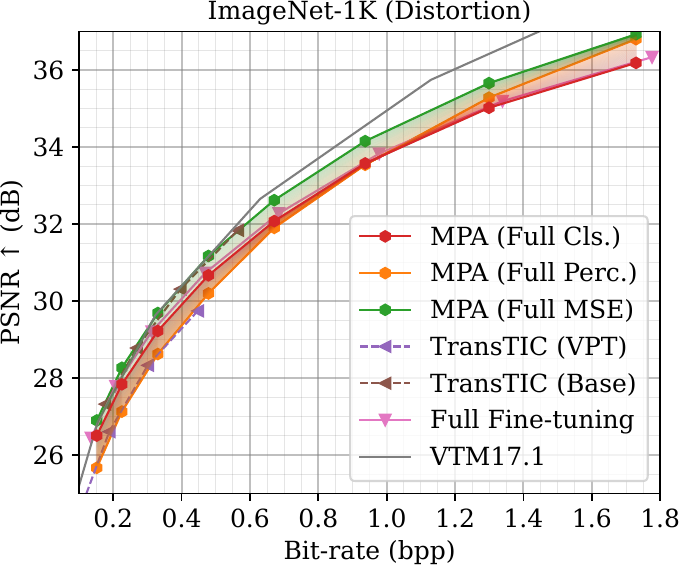}\label{subfig:imagenet_rd}}\hspace{10pt}
\subfloat{\includegraphics[scale=0.5]{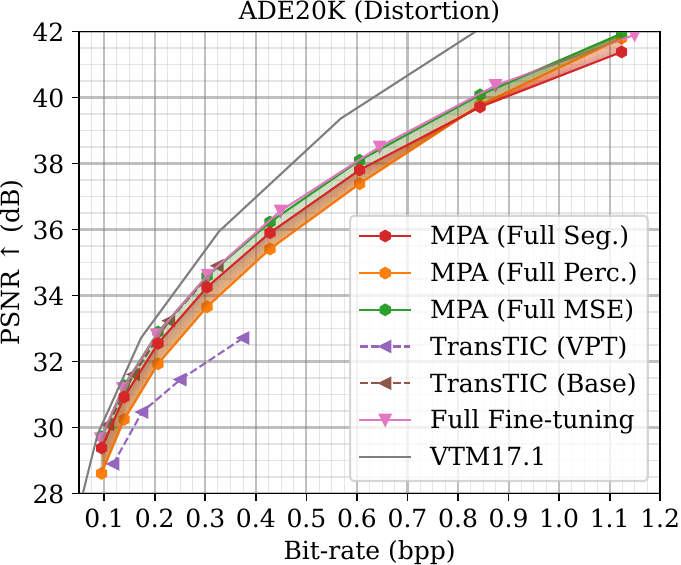}\label{subfig:ade20k_rd}}\hfill
\caption{R-D performance of MPA on ImageNet-1K~\cite{imagenet1k} and ADE20K~\cite{ade20k}.}
\label{fig:performance_imagenet1k_ade20k}
\end{center}
\end{figure}

\clearpage
\newpage

\subsection{Qualitative results}


\begin{figure}[htbp]
\begin{center}
{
\resizebox{\linewidth}{!}{
\scriptsize
\begin{tabular}{@{\hskip 0mm}l@{\hskip 1.0mm}l@{\hskip 1.0mm}l@{\hskip 1.0mm}l@{\hskip 1.0mm}l@{\hskip 1.0mm}l@{\hskip 1.0mm}l@{\hskip 1.0mm}l@{\hskip 1.0mm}l@{\hskip 0mm}}

\toprule
Original &
\begin{tabular}[c]{@{}l@{}} MRIC~\cite{agustsson2023multirealism} \\ ($\beta=2.56$) \end{tabular} &
\begin{tabular}[c]{@{}l@{}} MPA \\ ($\alpha=0.0$)    \end{tabular} &
\begin{tabular}[c]{@{}l@{}} MPA \\ ($\alpha=0.0001$) \end{tabular} &
\begin{tabular}[c]{@{}l@{}} MPA \\ ($\alpha=0.001$)  \end{tabular} &
\begin{tabular}[c]{@{}l@{}} MPA \\ ($\alpha=0.01$)   \end{tabular} &
\begin{tabular}[c]{@{}l@{}} MPA \\ ($\alpha=0.1$)    \end{tabular} &
\begin{tabular}[c]{@{}l@{}} MPA \\ ($\alpha=1.0$)    \end{tabular} &
\begin{tabular}[c]{@{}l@{}} MRIC~\cite{agustsson2023multirealism} \\ ($\beta=0.0$) \end{tabular}
\\ \midrule

\texttt{3f273} &
\begin{tabular}[c]{@{}l@{}} 0.0540bpp \\ 31.16dB \end{tabular} &
\begin{tabular}[c]{@{}l@{}} 0.0509bpp \\ 30.00dB \end{tabular} &
\begin{tabular}[c]{@{}l@{}} 0.0509bpp \\ 30.07dB \end{tabular} &
\begin{tabular}[c]{@{}l@{}} 0.0509bpp \\ 30.50dB \end{tabular} &
\begin{tabular}[c]{@{}l@{}} 0.0509bpp \\ 31.08dB \end{tabular} &
\begin{tabular}[c]{@{}l@{}} 0.0509bpp \\ 31.32dB \end{tabular} &
\begin{tabular}[c]{@{}l@{}} 0.0509bpp \\ 31.40dB \end{tabular} &
\begin{tabular}[c]{@{}l@{}} 0.0540bpp \\ 32.17dB \end{tabular} \\

\includegraphics[width=0.111\linewidth]{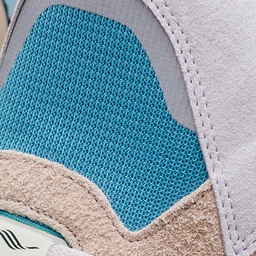} &
\includegraphics[width=0.111\linewidth]{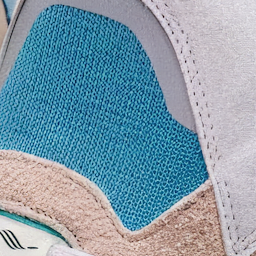} &
\includegraphics[width=0.111\linewidth]{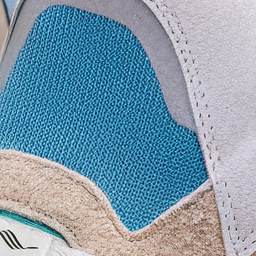} &
\includegraphics[width=0.111\linewidth]{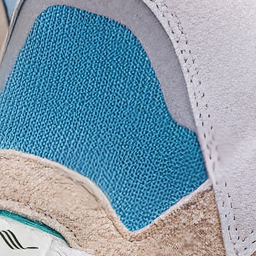} &
\includegraphics[width=0.111\linewidth]{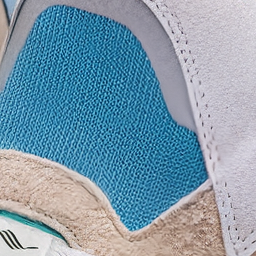} &
\includegraphics[width=0.111\linewidth]{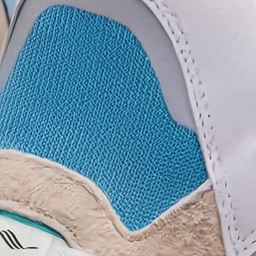} &
\includegraphics[width=0.111\linewidth]{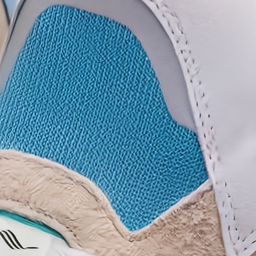} &
\includegraphics[width=0.111\linewidth]{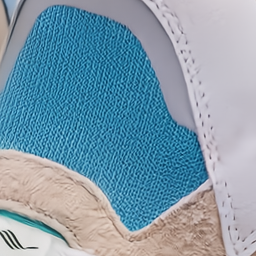} &
\includegraphics[width=0.111\linewidth]{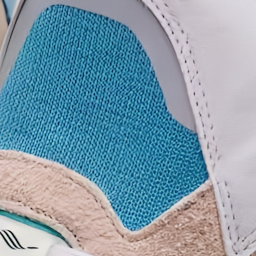} \\

\texttt{88c58} &
\begin{tabular}[c]{@{}l@{}} 0.0475bpp \\ 32.29dB \end{tabular} &
\begin{tabular}[c]{@{}l@{}} 0.0459bpp \\ 31.35dB \end{tabular} &
\begin{tabular}[c]{@{}l@{}} 0.0459bpp \\ 31.42dB \end{tabular} &
\begin{tabular}[c]{@{}l@{}} 0.0459bpp \\ 31.82dB \end{tabular} &
\begin{tabular}[c]{@{}l@{}} 0.0459bpp \\ 32.52dB \end{tabular} &
\begin{tabular}[c]{@{}l@{}} 0.0459bpp \\ 32.82dB \end{tabular} &
\begin{tabular}[c]{@{}l@{}} 0.0459bpp \\ 32.94dB \end{tabular} &
\begin{tabular}[c]{@{}l@{}} 0.0475bpp \\ 33.73dB \end{tabular} \\

\includegraphics[width=0.111\linewidth]{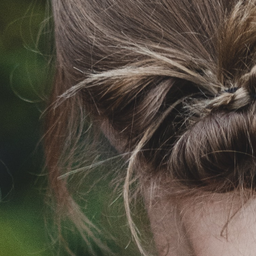} &
\includegraphics[width=0.111\linewidth]{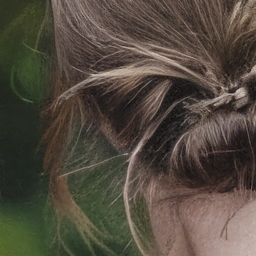} &
\includegraphics[width=0.111\linewidth]{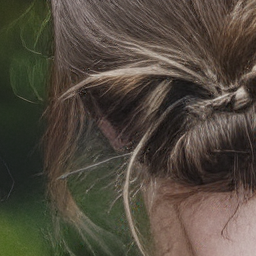} &
\includegraphics[width=0.111\linewidth]{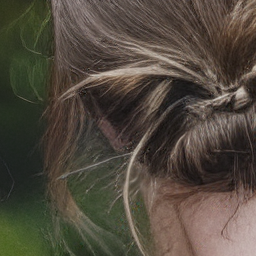} &
\includegraphics[width=0.111\linewidth]{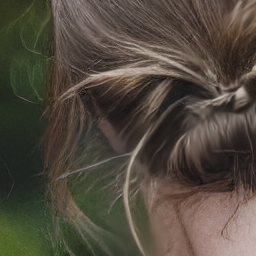} &
\includegraphics[width=0.111\linewidth]{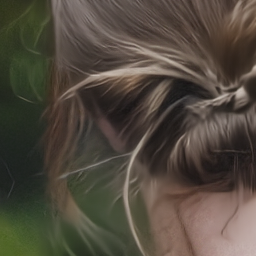} &
\includegraphics[width=0.111\linewidth]{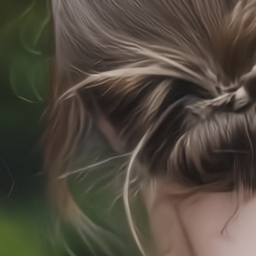} &
\includegraphics[width=0.111\linewidth]{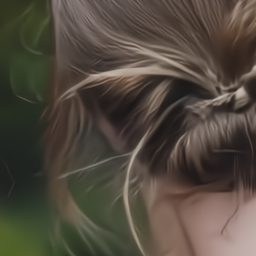} &
\includegraphics[width=0.111\linewidth]{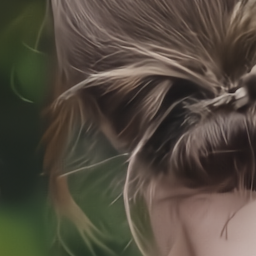} \\

\texttt{1487a} &
\begin{tabular}[c]{@{}l@{}} 0.0755bpp \\ 29.36dB \end{tabular} &
\begin{tabular}[c]{@{}l@{}} 0.0636bpp \\ 28.44dB \end{tabular} &
\begin{tabular}[c]{@{}l@{}} 0.0636bpp \\ 28.48dB \end{tabular} &
\begin{tabular}[c]{@{}l@{}} 0.0636bpp \\ 28.76dB \end{tabular} &
\begin{tabular}[c]{@{}l@{}} 0.0636bpp \\ 29.52dB \end{tabular} &
\begin{tabular}[c]{@{}l@{}} 0.0636bpp \\ 30.01dB \end{tabular} &
\begin{tabular}[c]{@{}l@{}} 0.0636bpp \\ 30.09dB \end{tabular} &
\begin{tabular}[c]{@{}l@{}} 0.0755bpp \\ 30.96dB \end{tabular} \\

\includegraphics[width=0.111\linewidth]{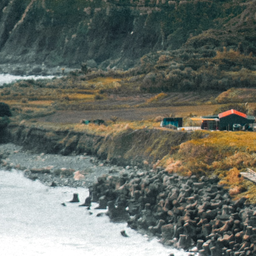} &
\includegraphics[width=0.111\linewidth]{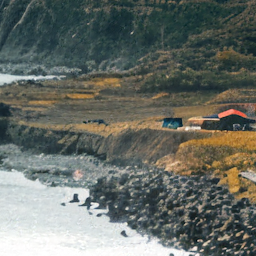} &
\includegraphics[width=0.111\linewidth]{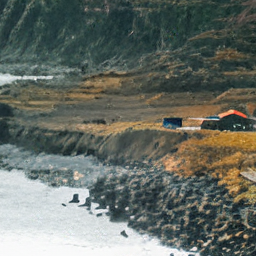} &
\includegraphics[width=0.111\linewidth]{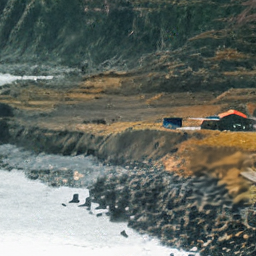} &
\includegraphics[width=0.111\linewidth]{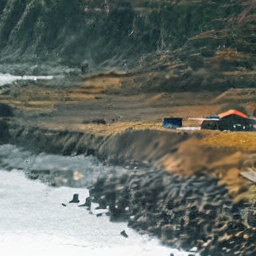} &
\includegraphics[width=0.111\linewidth]{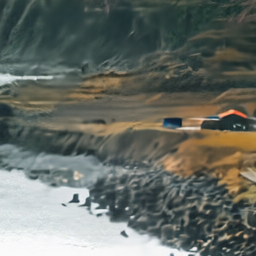} &
\includegraphics[width=0.111\linewidth]{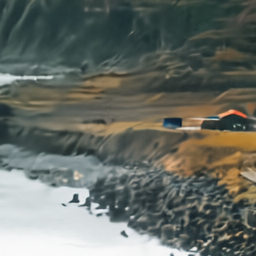} &
\includegraphics[width=0.111\linewidth]{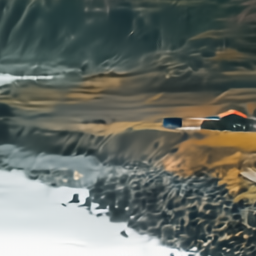} &
\includegraphics[width=0.111\linewidth]{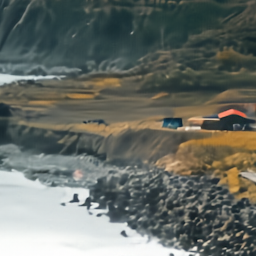} \\

\texttt{f5003} &
\begin{tabular}[c]{@{}l@{}} 0.0750bpp \\ 30.34dB \end{tabular} &
\begin{tabular}[c]{@{}l@{}} 0.0660bpp \\ 29.11dB \end{tabular} &
\begin{tabular}[c]{@{}l@{}} 0.0660bpp \\ 29.14dB \end{tabular} &
\begin{tabular}[c]{@{}l@{}} 0.0660bpp \\ 29.40dB \end{tabular} &
\begin{tabular}[c]{@{}l@{}} 0.0660bpp \\ 30.15dB \end{tabular} &
\begin{tabular}[c]{@{}l@{}} 0.0660bpp \\ 30.48dB \end{tabular} &
\begin{tabular}[c]{@{}l@{}} 0.0660bpp \\ 30.61dB \end{tabular} &
\begin{tabular}[c]{@{}l@{}} 0.0750bpp \\ 31.71dB \end{tabular} \\

\includegraphics[width=0.111\linewidth]{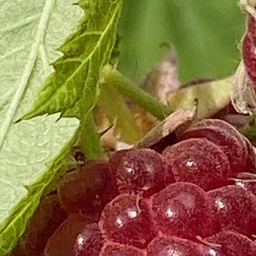} &
\includegraphics[width=0.111\linewidth]{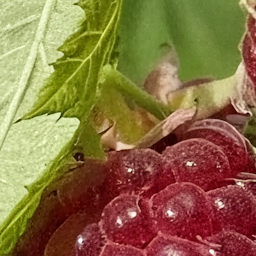} &
\includegraphics[width=0.111\linewidth]{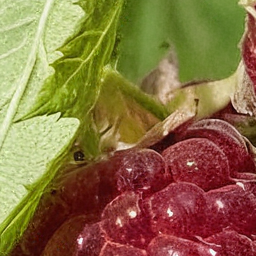} &
\includegraphics[width=0.111\linewidth]{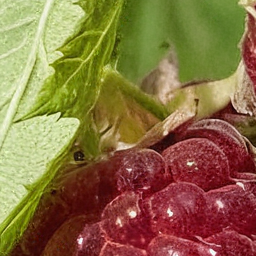} &
\includegraphics[width=0.111\linewidth]{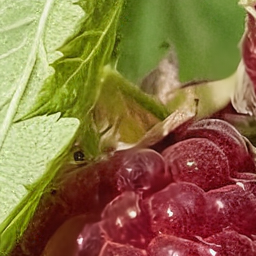} &
\includegraphics[width=0.111\linewidth]{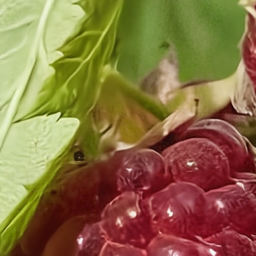} &
\includegraphics[width=0.111\linewidth]{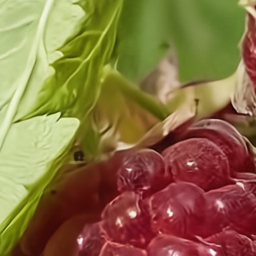} &
\includegraphics[width=0.111\linewidth]{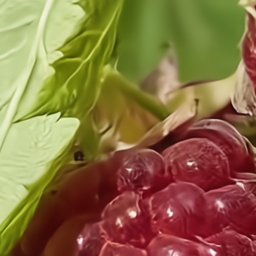} &
\includegraphics[width=0.111\linewidth]{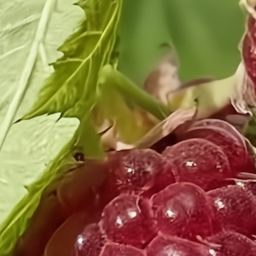} \\

\bottomrule
\end{tabular}
}
}
\caption{We visualize the reconstructed images of MPA with variable $\alpha$ compared to MRIC~\cite{agustsson2023multirealism}. The image is from CLIC test set~\cite{CLIC2020} and is cropped to 256$\times$256$\times$3. The selected paths are $\phi_{\text{perc}}$ and $\phi_{\text{MSE}}$. Larger $\beta$ for MRIC and smaller $\alpha$ for MPA stand for higher realism. Note that both the bitrate and the PSNR are calculated on uncropped images.}
\label{fig:clic_vis}
\end{center}
\end{figure}


\begin{figure}[htbp]
\begin{center}
{
\resizebox{\linewidth}{!}{
\scriptsize
\begin{tabular}{@{\hskip 0mm}l@{\hskip 1.0mm}l@{\hskip 1.0mm}l@{\hskip 1.0mm}l@{\hskip 0mm}}

\toprule
Original: \texttt{val00035565} &
Score Map (1-st Stage) &
Score Map (2-nd Stage) &
Score Map (3-rd Stage) \\

\includegraphics[width=0.16\linewidth]{figures/visualization/imagenet1k/val00035565/val00035565_original.png} &
\includegraphics[width=0.2\linewidth]{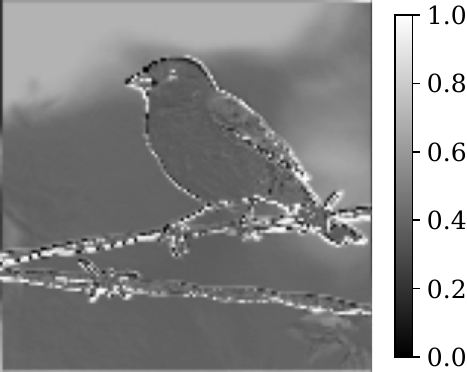} &
\includegraphics[width=0.2\linewidth]{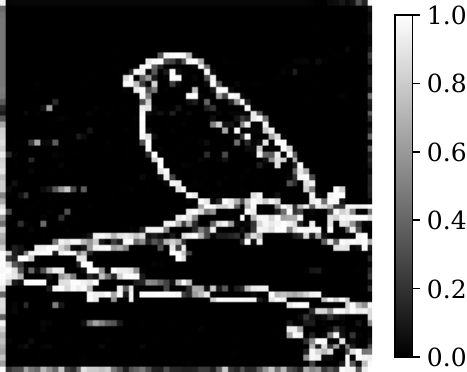} &
\includegraphics[width=0.2\linewidth]{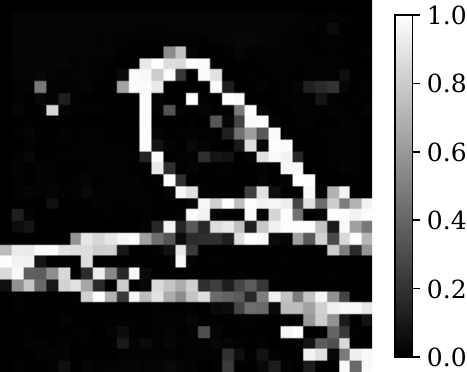} \\
\\[-2ex]\midrule

\multicolumn{4}{c}{Bit Allocation} \\

$q=1$ (0.0888bpp) & $q=2$ (0.1166bpp) & $q=3$ (0.1630bpp) & $q=4$ (0.2294bpp) \\
\includegraphics[width=0.2\linewidth]{figures/visualization/imagenet1k/val00035565/val00035565_bit_allocation_q1.pdf} &
\includegraphics[width=0.2\linewidth]{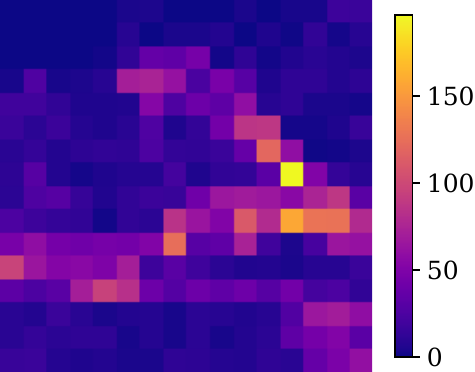} &
\includegraphics[width=0.2\linewidth]{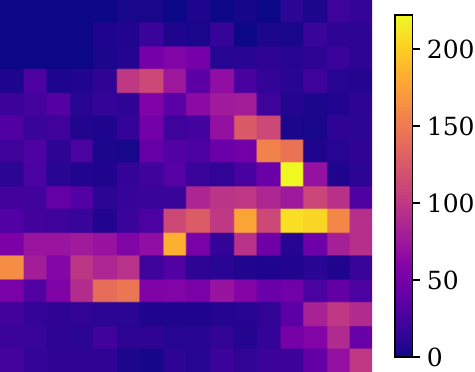} &
\includegraphics[width=0.2\linewidth]{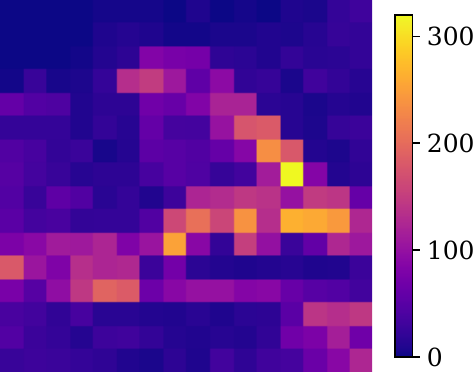} \\
$q=5$ (0.3168bpp) & $q=6$ (0.4560bpp) & $q=7$ (0.6562bpp) & $q=8$ (0.9238bpp) \\
\includegraphics[width=0.2\linewidth]{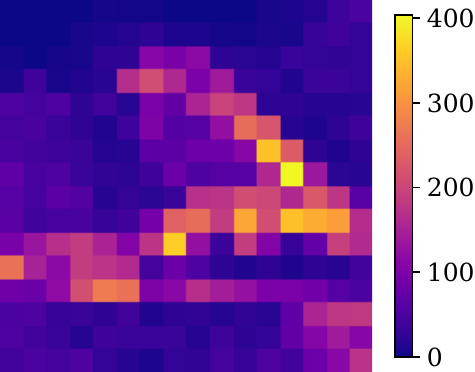} &
\includegraphics[width=0.2\linewidth]{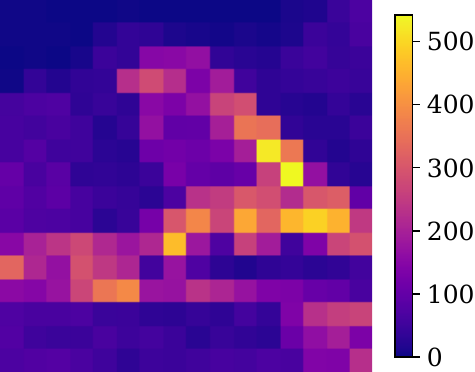} &
\includegraphics[width=0.2\linewidth]{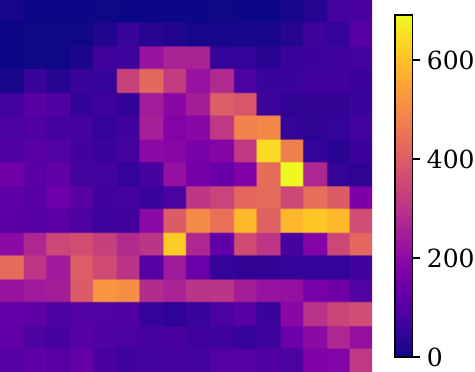} &
\includegraphics[width=0.2\linewidth]{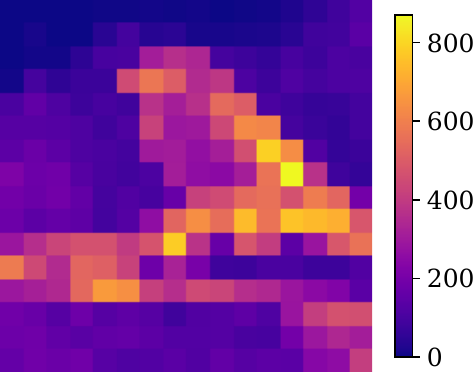} \\ \bottomrule
\end{tabular}
}
}
\caption{We visualize the bit allocation and score maps in the encoder. The image is from ImageNet validation set~\cite{imagenet1k} and resized to 256$\times$256$\times$3. The 1st, 2nd, and 3rd stages are the first three stages of the encoder, consecutively.}
\label{fig:enc_val00035565}
\end{center}
\end{figure}

\begin{figure}[htbp]
\begin{center}
{
\resizebox{\linewidth}{!}{
\scriptsize
\begin{tabular}{@{\hskip 0mm}l@{\hskip 1.0mm}l@{\hskip 1.0mm}l@{\hskip 1.0mm}l@{\hskip 1.0mm}l@{\hskip 0mm}}

\toprule
Reconstruction &
FullGrad~\cite{srinivas2019fullgradient} &
Score Map (3-rd Stage) &
Score Map (2-nd Stage) &
Score Map (1-st Stage)
\\ \midrule

Cls. Path: 29.41dB
& Confidence: 81.26\%
& \multicolumn{3}{c}{The score maps are inverted for better visualization (higher for the side path).} \\
\includegraphics[width=0.16\linewidth]{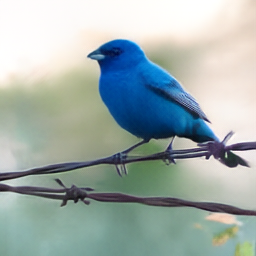} &
\includegraphics[width=0.16\linewidth]{figures/visualization/imagenet1k/val00035565/val00035565_cls_grad_cam.png} &
\includegraphics[width=0.2\linewidth]{figures/visualization/imagenet1k/val00035565/val00035565_predictor_cls_g_s2.pdf} &
\includegraphics[width=0.2\linewidth]{figures/visualization/imagenet1k/val00035565/val00035565_predictor_cls_g_s4.pdf} &
\includegraphics[width=0.2\linewidth]{figures/visualization/imagenet1k/val00035565/val00035565_predictor_cls_g_s6.pdf} \\
Seg. Path: 29.41dB
& Confidence: 68.26\%
& & & \\
\includegraphics[width=0.16\linewidth]{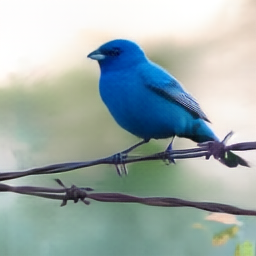} &
\includegraphics[width=0.16\linewidth]{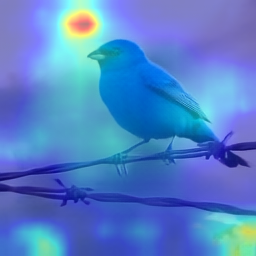} &
\includegraphics[width=0.2\linewidth]{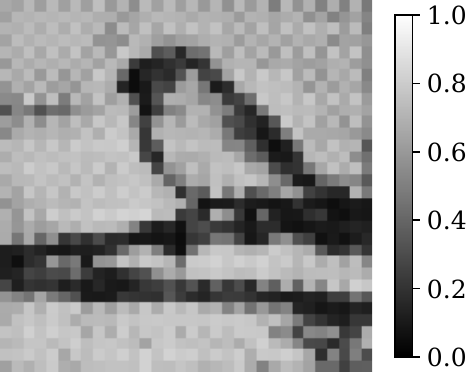} &
\includegraphics[width=0.2\linewidth]{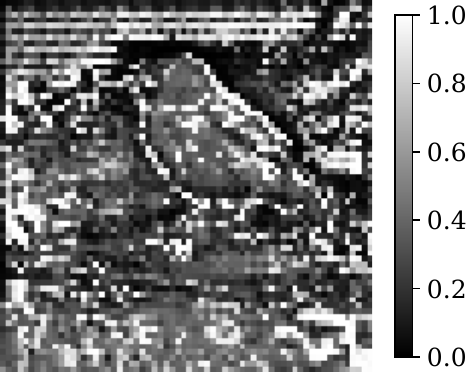} &
\includegraphics[width=0.2\linewidth]{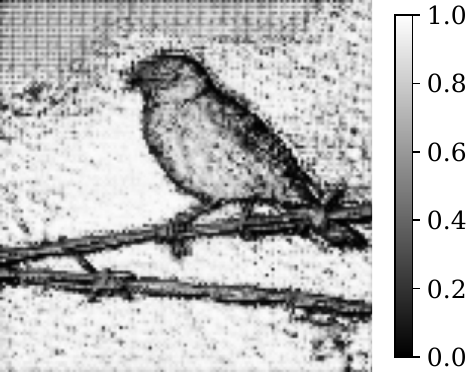} \\
MSE Path: 29.50dB
& Confidence: 37.55\%
& & & \\
\includegraphics[width=0.16\linewidth]{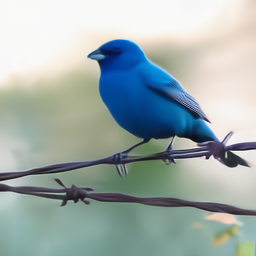} &
\includegraphics[width=0.16\linewidth]{figures/visualization/imagenet1k/val00035565/val00035565_mse_grad_cam.png} &
\includegraphics[width=0.2\linewidth]{figures/visualization/imagenet1k/val00035565/val00035565_predictor_mse_g_s2.pdf} &
\includegraphics[width=0.2\linewidth]{figures/visualization/imagenet1k/val00035565/val00035565_predictor_mse_g_s4.pdf} &
\includegraphics[width=0.2\linewidth]{figures/visualization/imagenet1k/val00035565/val00035565_predictor_mse_g_s6.pdf} \\
\\[-2ex]\midrule

Perc. Path: 28.65dB
& Confidence: 77.46\%
& &
& Original: \texttt{val00035565} \\
\includegraphics[width=0.16\linewidth]{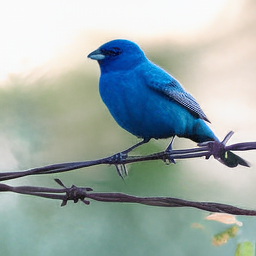} &
\includegraphics[width=0.16\linewidth]{figures/visualization/imagenet1k/val00035565/val00035565_perc_grad_cam.png} & & &
\includegraphics[width=0.16\linewidth]{figures/visualization/imagenet1k/val00035565/val00035565_original.png} \\ \bottomrule
\end{tabular}
}
}
\caption{We visualize the reconstructed images, FullGrad~\cite{srinivas2019fullgradient} and score maps predicted by importance predictors in each path. The image is from ImageNet validation set~\cite{imagenet1k} and resized to 256$\times$256$\times$3. The 3rd, 2nd, and 1st stages are the last three stages of the decoder, consecutively. The regions with warmer colors in FullGrad represent areas with larger gradients, indicating a stronger impact on the classification decision. $q$ is set to 1 for a more distinct comparison. Note that the score maps are inverted for better visualization, i.e., larger scores indicate prioritized for entering the selected side path. The bitrate is 0.0888bpp.}
\label{fig:cam_val00035565}
\end{center}
\end{figure}


\begin{figure}[htbp]
\begin{center}
{
\resizebox{\linewidth}{!}{
\scriptsize
\begin{tabular}{@{\hskip 0mm}l@{\hskip 1.0mm}l@{\hskip 1.0mm}l@{\hskip 1.0mm}l@{\hskip 0mm}}

\toprule
Original: \texttt{val00017577} &
Score Map (1-st Stage) &
Score Map (2-nd Stage) &
Score Map (3-rd Stage) \\

\includegraphics[width=0.16\linewidth]{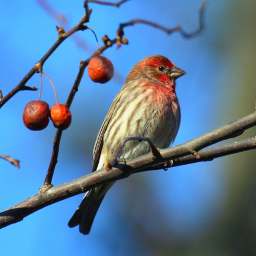} &
\includegraphics[width=0.2\linewidth]{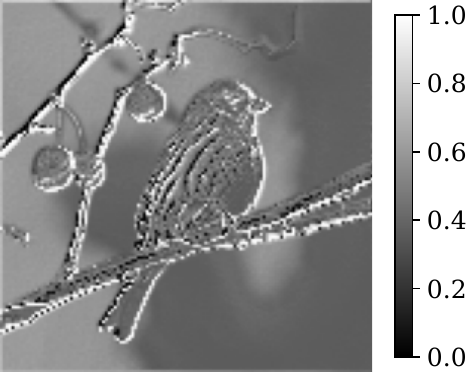} &
\includegraphics[width=0.2\linewidth]{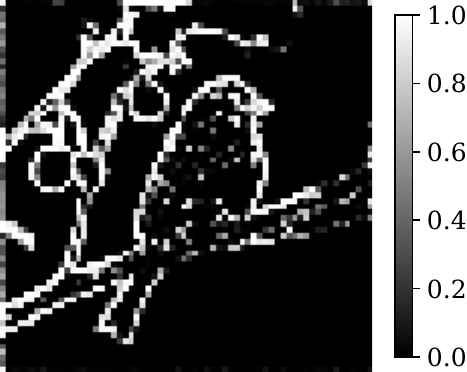} &
\includegraphics[width=0.2\linewidth]{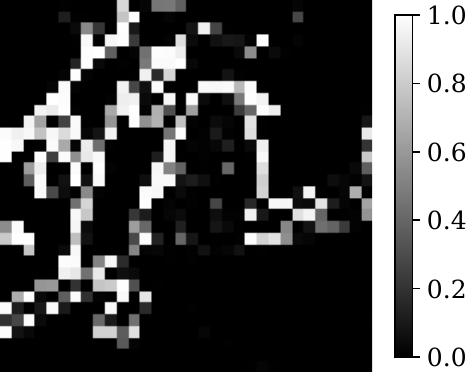} \\
\\[-2ex]\midrule

\multicolumn{4}{c}{Bit Allocation} \\

$q=1$ (0.1152bpp) & $q=2$ (0.1616bpp) & $q=3$ (0.2163bpp) & $q=4$ (0.2998bpp) \\
\includegraphics[width=0.2\linewidth]{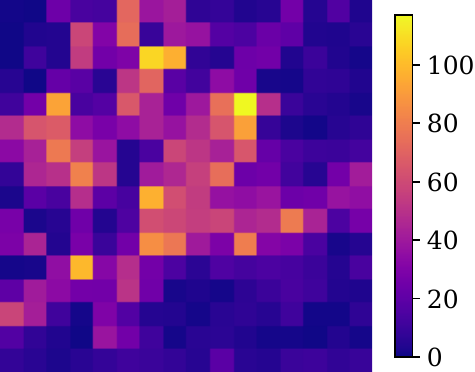} &
\includegraphics[width=0.2\linewidth]{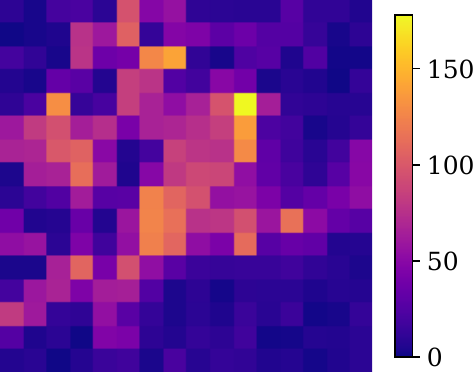} &
\includegraphics[width=0.2\linewidth]{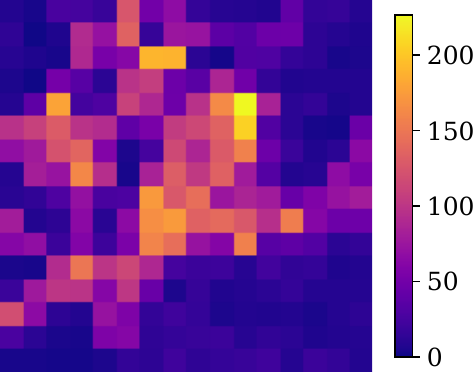} &
\includegraphics[width=0.2\linewidth]{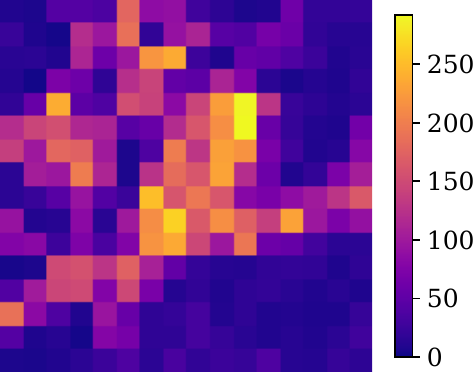} \\
$q=5$ (0.4150bpp) & $q=6$ (0.5874bpp) & $q=7$ (0.8256bpp) & $q=8$ (1.1650bpp) \\
\includegraphics[width=0.2\linewidth]{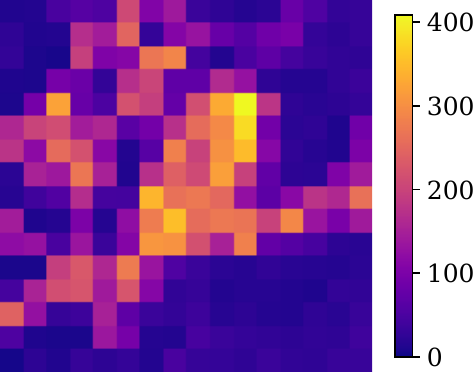} &
\includegraphics[width=0.2\linewidth]{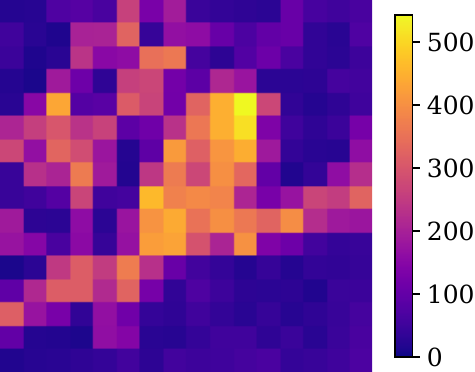} &
\includegraphics[width=0.2\linewidth]{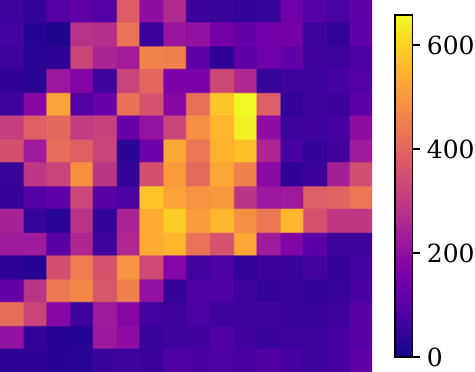} &
\includegraphics[width=0.2\linewidth]{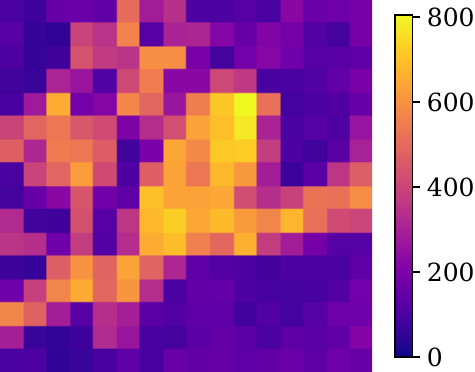} \\ \bottomrule
\end{tabular}
}
}
\caption{We visualize the bit allocation and score maps in the encoder. The image is from ImageNet validation set~\cite{imagenet1k} and resized to 256$\times$256$\times$3. The 1st, 2nd, and 3rd stages are the first three stages of the encoder, consecutively.}
\label{fig:enc_val00017577}
\end{center}
\end{figure}

\begin{figure}[htbp]
\begin{center}
{
\resizebox{\linewidth}{!}{
\scriptsize
\begin{tabular}{@{\hskip 0mm}l@{\hskip 1.0mm}l@{\hskip 1.0mm}l@{\hskip 1.0mm}l@{\hskip 1.0mm}l@{\hskip 0mm}}

\toprule
Reconstruction &
FullGrad~\cite{srinivas2019fullgradient} &
Score Map (3-rd Stage) &
Score Map (2-nd Stage) &
Score Map (1-st Stage)
\\ \midrule

Cls. Path: 27.46dB
& Confidence: 76.22\%
& \multicolumn{3}{c}{The score maps are inverted for better visualization (higher for the side path).} \\
\includegraphics[width=0.16\linewidth]{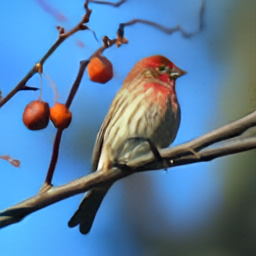} &
\includegraphics[width=0.16\linewidth]{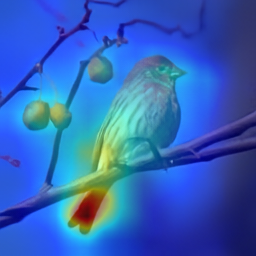} &
\includegraphics[width=0.2\linewidth]{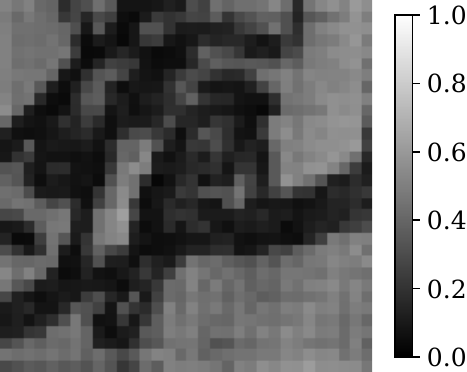} &
\includegraphics[width=0.2\linewidth]{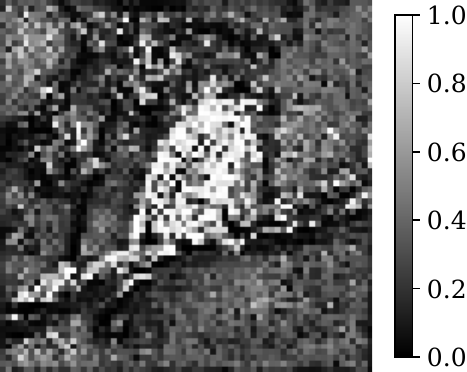} &
\includegraphics[width=0.2\linewidth]{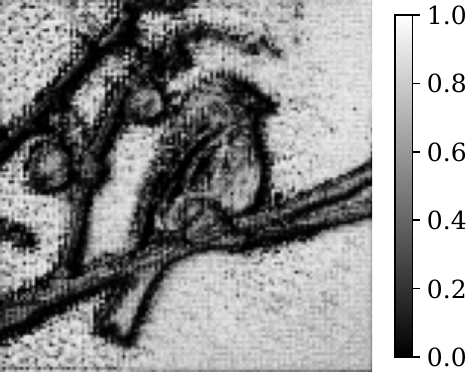} \\
Seg. Path: 27.18dB
& Confidence: 62.71\%
& & & \\
\includegraphics[width=0.16\linewidth]{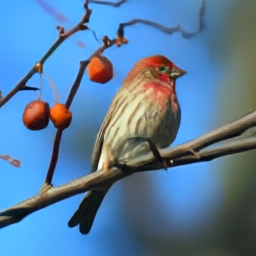} &
\includegraphics[width=0.16\linewidth]{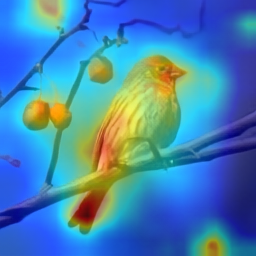} &
\includegraphics[width=0.2\linewidth]{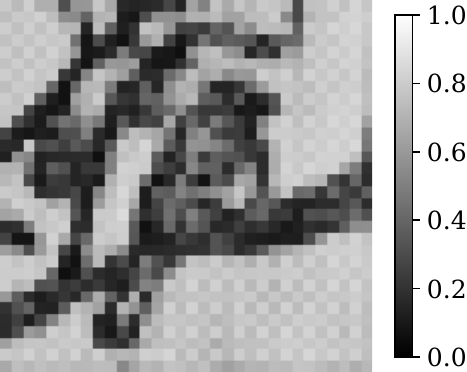} &
\includegraphics[width=0.2\linewidth]{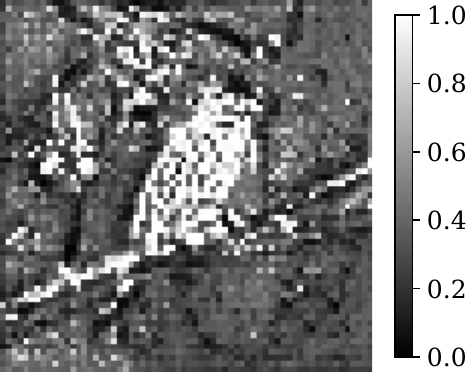} &
\includegraphics[width=0.2\linewidth]{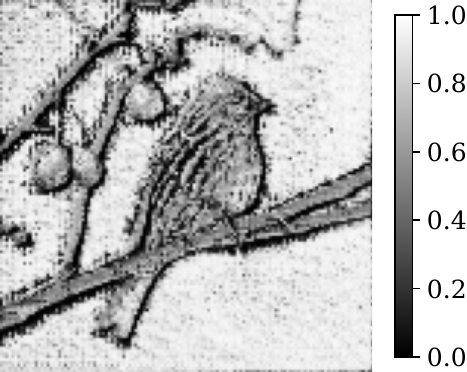} \\
MSE Path: 27.85dB
& Confidence: 44.31\%
& & & \\
\includegraphics[width=0.16\linewidth]{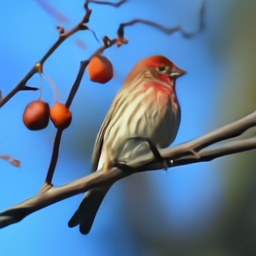} &
\includegraphics[width=0.16\linewidth]{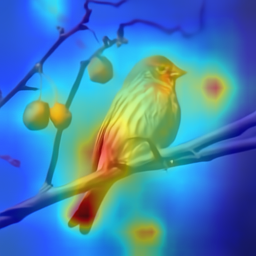} &
\includegraphics[width=0.2\linewidth]{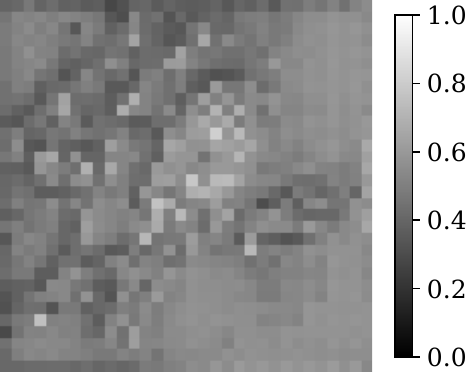} &
\includegraphics[width=0.2\linewidth]{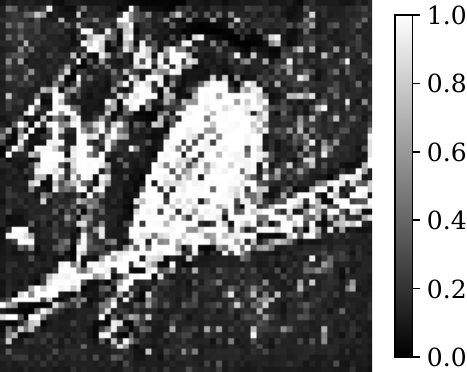} &
\includegraphics[width=0.2\linewidth]{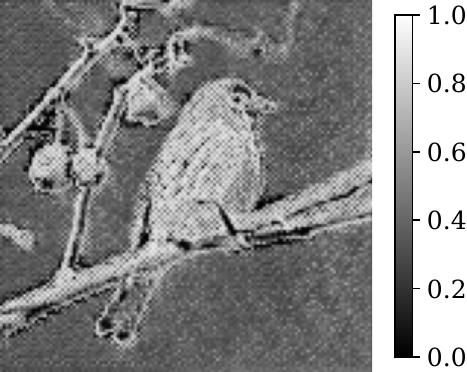} \\
\\[-2ex]\midrule

Perc. Path: 26.91dB
& Confidence: 75.23\%
& &
& Original: \texttt{val00017577} \\
\includegraphics[width=0.16\linewidth]{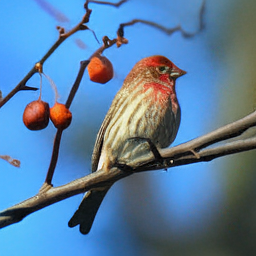} &
\includegraphics[width=0.16\linewidth]{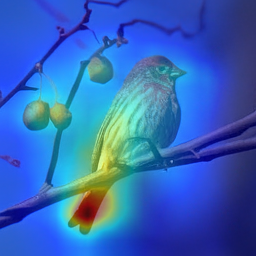} & & &
\includegraphics[width=0.16\linewidth]{figures/visualization/imagenet1k/val00017577/val00017577_original.png} \\ \bottomrule
\end{tabular}
}
}
\caption{We visualize the reconstructed images, FullGrad~\cite{srinivas2019fullgradient} and score maps predicted by importance predictors in each path. The image is from ImageNet validation set~\cite{imagenet1k} and resized to 256$\times$256$\times$3. The 3rd, 2nd, and 1st stages are the last three stages of the decoder, consecutively. The regions with warmer colors in FullGrad represent areas with larger gradients, indicating a stronger impact on the classification decision. $q$ is set to 1 for a more distinct comparison. Note that the score maps are inverted for better visualization, i.e., larger scores indicate prioritized for entering the selected side path. The bitrate is 0.1152bpp.}
\label{fig:cam_val00017577}
\end{center}
\end{figure}


\begin{figure}[htbp]
\begin{center}
{
\resizebox{\linewidth}{!}{
\scriptsize
\begin{tabular}{@{\hskip 0mm}l@{\hskip 1.0mm}l@{\hskip 1.0mm}l@{\hskip 1.0mm}l@{\hskip 0mm}}

\toprule
Original: \texttt{val00022926} &
Score Map (1-st Stage) &
Score Map (2-nd Stage) &
Score Map (3-rd Stage) \\

\includegraphics[width=0.16\linewidth]{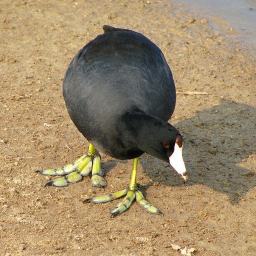} &
\includegraphics[width=0.2\linewidth]{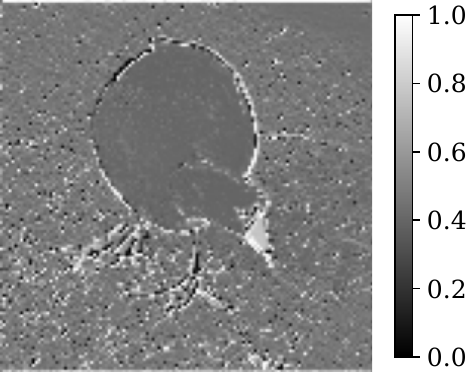} &
\includegraphics[width=0.2\linewidth]{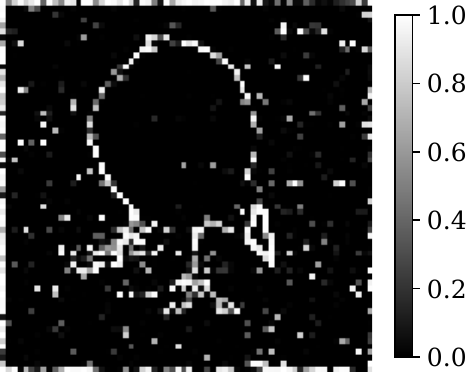} &
\includegraphics[width=0.2\linewidth]{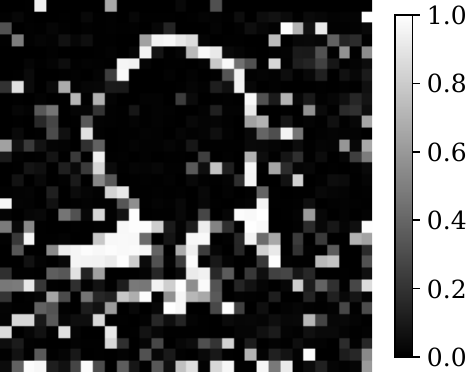} \\
\\[-2ex]\midrule

\multicolumn{4}{c}{Bit Allocation} \\

$q=1$ (0.1279bpp) & $q=2$ (0.2343bpp) & $q=3$ (0.3984bpp) & $q=4$ (0.6533bpp) \\
\includegraphics[width=0.2\linewidth]{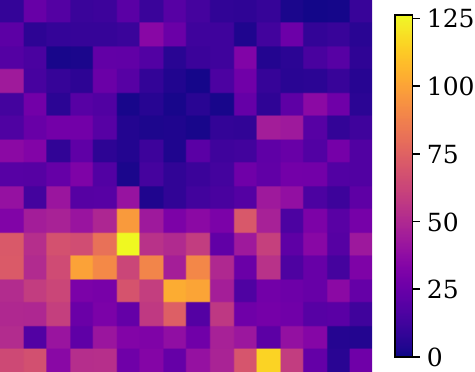} &
\includegraphics[width=0.2\linewidth]{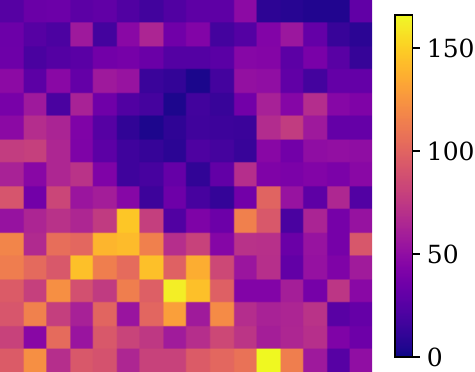} &
\includegraphics[width=0.2\linewidth]{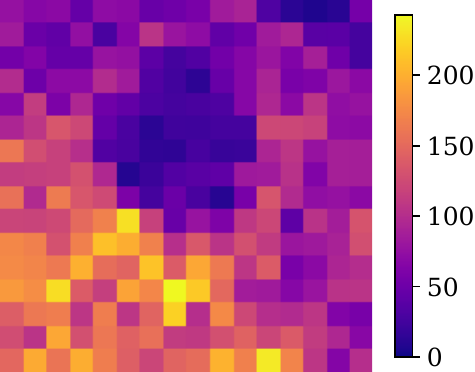} &
\includegraphics[width=0.2\linewidth]{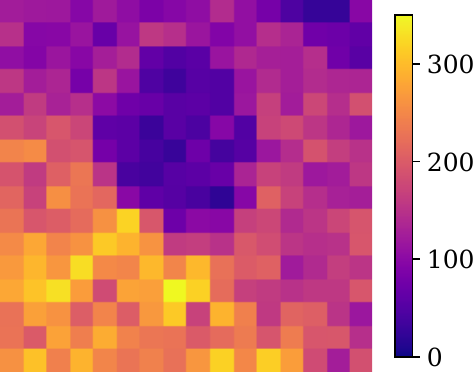} \\
$q=5$ (0.9980bpp) & $q=6$ (1.4140bpp) & $q=7$ (1.9096bpp) & $q=8$ (2.4282bpp) \\
\includegraphics[width=0.2\linewidth]{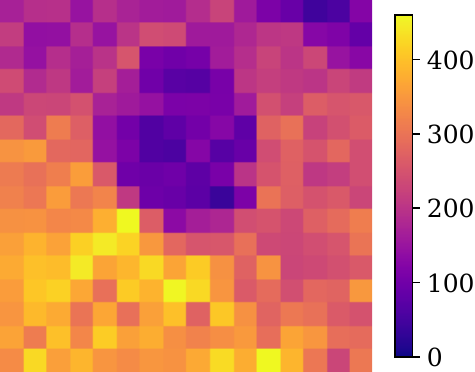} &
\includegraphics[width=0.2\linewidth]{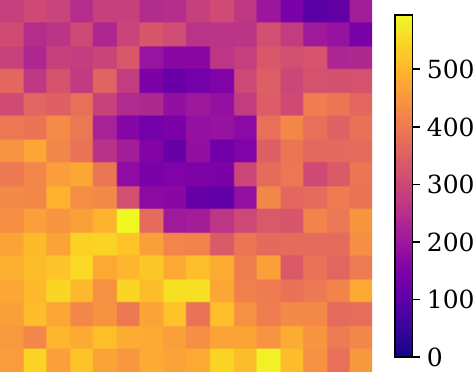} &
\includegraphics[width=0.2\linewidth]{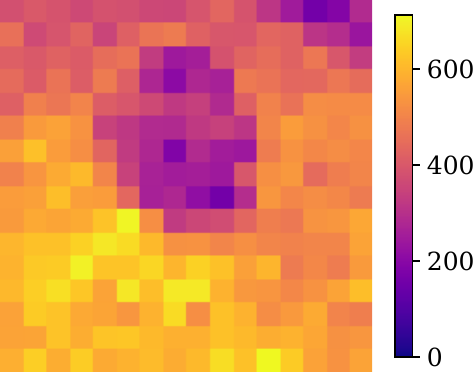} &
\includegraphics[width=0.2\linewidth]{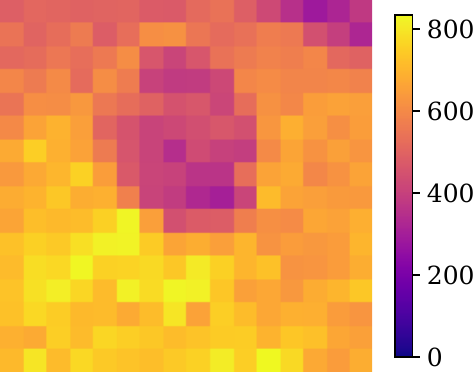} \\ \bottomrule
\end{tabular}
}
}
\caption{We visualize the bit allocation and score maps in the encoder. The image is from ImageNet validation set~\cite{imagenet1k} and resized to 256$\times$256$\times$3. The 1st, 2nd, and 3rd stages are the first three stages of the encoder, consecutively.}
\label{fig:enc_val00022926}
\end{center}
\end{figure}

\begin{figure}[htbp]
\begin{center}
{
\resizebox{\linewidth}{!}{
\scriptsize
\begin{tabular}{@{\hskip 0mm}l@{\hskip 1.0mm}l@{\hskip 1.0mm}l@{\hskip 1.0mm}l@{\hskip 1.0mm}l@{\hskip 0mm}}

\toprule
Reconstruction &
FullGrad~\cite{srinivas2019fullgradient} &
Score Map (3-rd Stage) &
Score Map (2-nd Stage) &
Score Map (1-st Stage)
\\ \midrule

Cls. Path: 24.80dB
& Confidence: 83.21\%
& \multicolumn{3}{c}{The score maps are inverted for better visualization (higher for the side path).} \\
\includegraphics[width=0.16\linewidth]{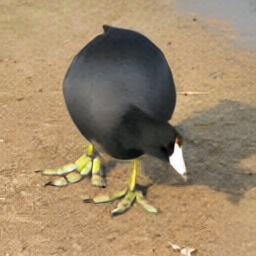} &
\includegraphics[width=0.16\linewidth]{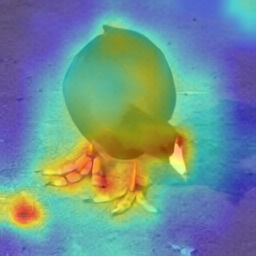} &
\includegraphics[width=0.2\linewidth]{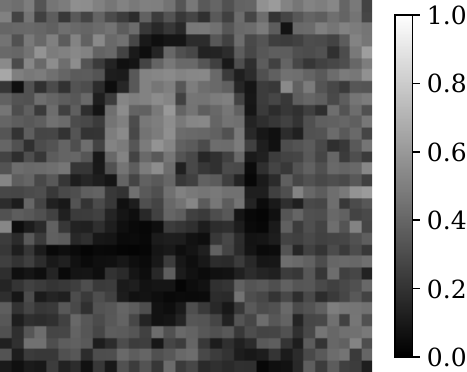} &
\includegraphics[width=0.2\linewidth]{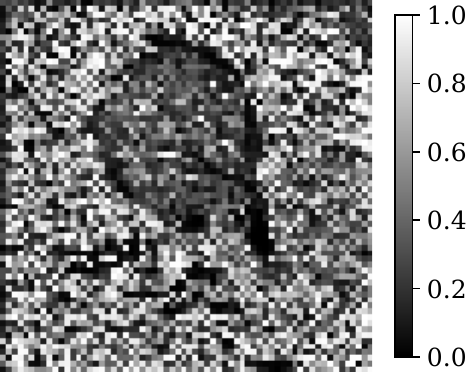} &
\includegraphics[width=0.2\linewidth]{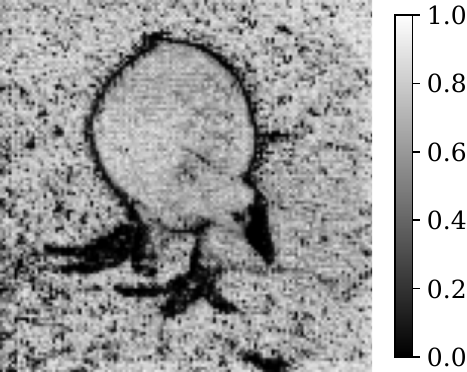} \\
Seg. Path: 24.80dB
& Confidence: 64.81\%
& & & \\
\includegraphics[width=0.16\linewidth]{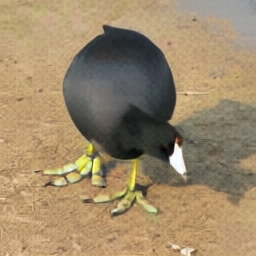} &
\includegraphics[width=0.16\linewidth]{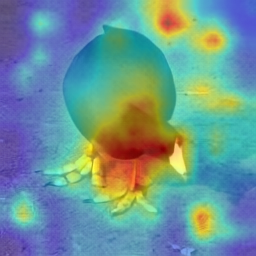} &
\includegraphics[width=0.2\linewidth]{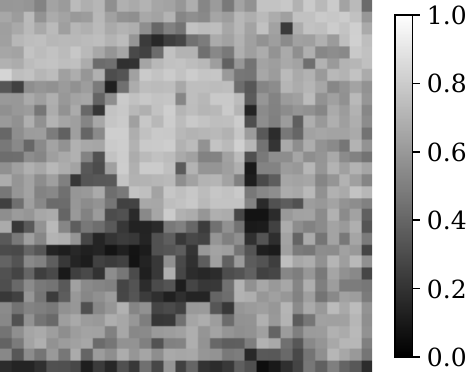} &
\includegraphics[width=0.2\linewidth]{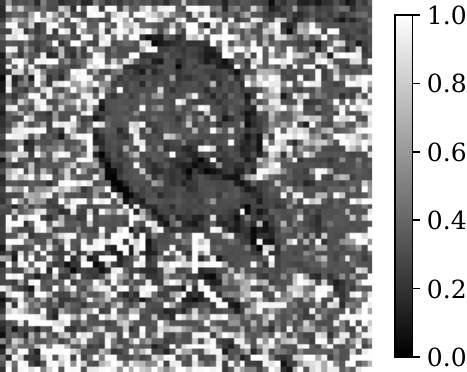} &
\includegraphics[width=0.2\linewidth]{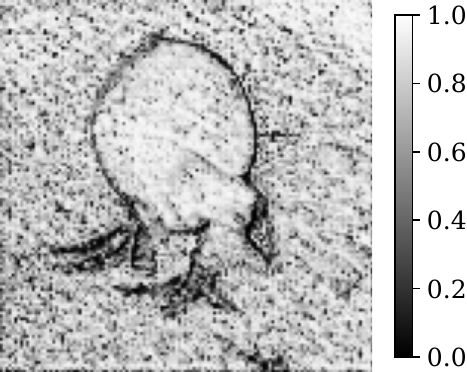} \\
MSE Path: 25.18dB
& Confidence: 61.06\%
& & & \\
\includegraphics[width=0.16\linewidth]{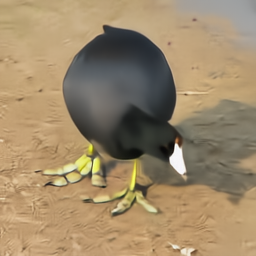} &
\includegraphics[width=0.16\linewidth]{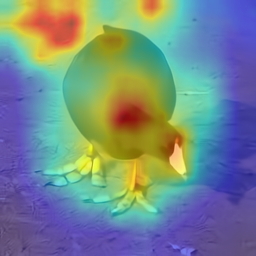} &
\includegraphics[width=0.2\linewidth]{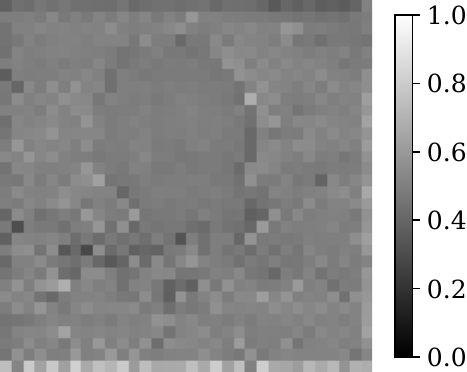} &
\includegraphics[width=0.2\linewidth]{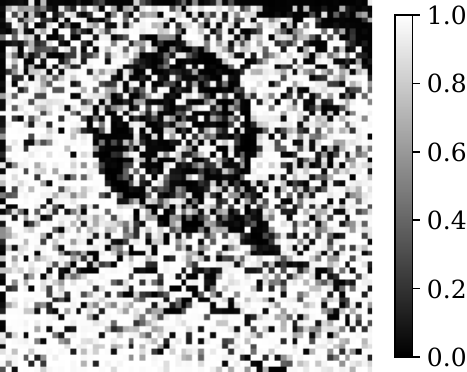} &
\includegraphics[width=0.2\linewidth]{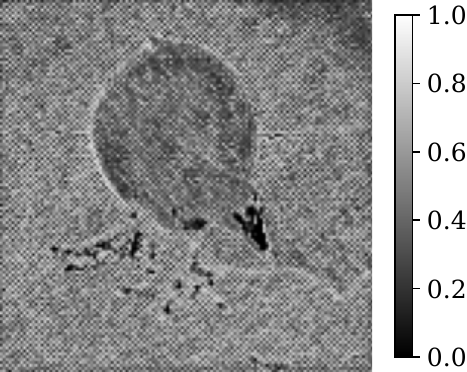} \\
\\[-2ex]\midrule

Perc. Path: 23.29dB
& Confidence: 73.79\%
& &
& Original: \texttt{val00022926} \\
\includegraphics[width=0.16\linewidth]{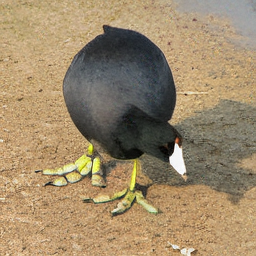} &
\includegraphics[width=0.16\linewidth]{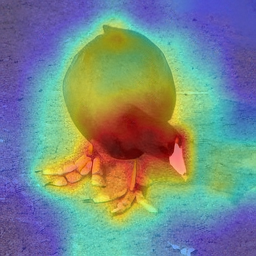} & & &
\includegraphics[width=0.16\linewidth]{figures/visualization/imagenet1k/val00022926/val00022926_original.png} \\ \bottomrule
\end{tabular}
}
}
\caption{We visualize the reconstructed images, FullGrad~\cite{srinivas2019fullgradient} and score maps predicted by importance predictors in each path. The image is from ImageNet validation set~\cite{imagenet1k} and resized to 256$\times$256$\times$3. The 3rd, 2nd, and 1st stages are the last three stages of the decoder, consecutively. The regions with warmer colors in FullGrad represent areas with larger gradients, indicating a stronger impact on the classification decision. $q$ is set to 1 for a more distinct comparison. Note that the score maps are inverted for better visualization, i.e., larger scores indicate prioritized for entering the selected side path. The bitrate is 0.1279bpp.}
\label{fig:cam_val00022926}
\end{center}
\end{figure}


\begin{figure}[htbp]
\begin{center}
{
\resizebox{\linewidth}{!}{
\scriptsize
\begin{tabular}{@{\hskip 0mm}l@{\hskip 1.0mm}l@{\hskip 1.0mm}l@{\hskip 1.0mm}l@{\hskip 0mm}}

\toprule
Original: \texttt{val00000269} &
Score Map (1-st Stage) &
Score Map (2-nd Stage) &
Score Map (3-rd Stage) \\

\includegraphics[width=0.16\linewidth]{figures/visualization/ade20k/val00000269/val00000269_original.png} &
\includegraphics[width=0.2\linewidth]{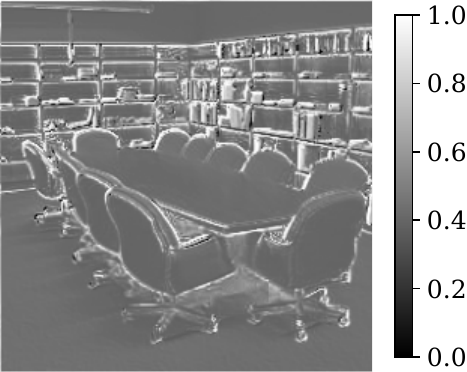} &
\includegraphics[width=0.2\linewidth]{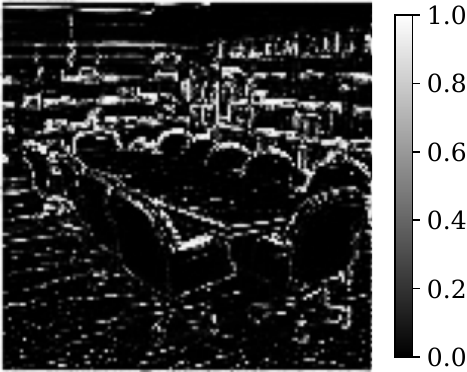} &
\includegraphics[width=0.2\linewidth]{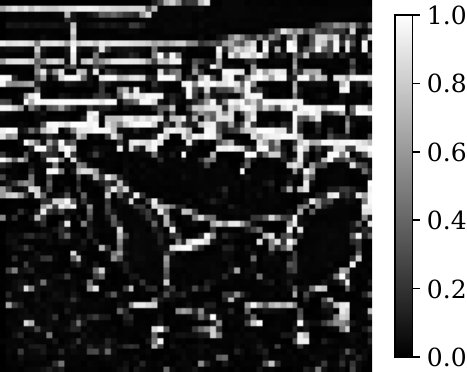} \\
\\[-2ex]\midrule

\multicolumn{4}{c}{Bit Allocation} \\

$q=1$ (0.0718bpp) & $q=2$ (0.0972bpp) & $q=3$ (0.1383bpp) & $q=4$ (0.1998bpp) \\
\includegraphics[width=0.2\linewidth]{figures/visualization/ade20k/val00000269/val00000269_bit_allocation_q1.pdf} &
\includegraphics[width=0.2\linewidth]{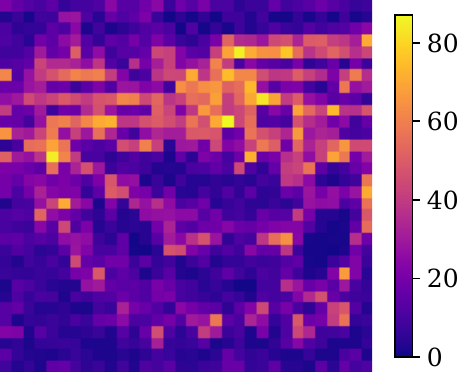} &
\includegraphics[width=0.2\linewidth]{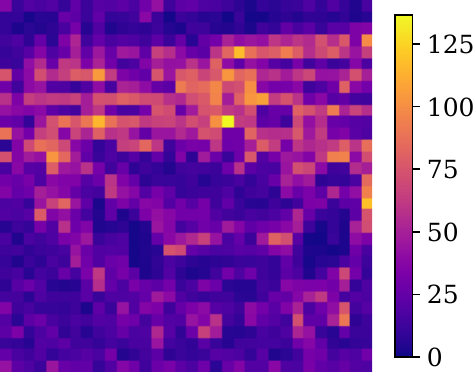} &
\includegraphics[width=0.2\linewidth]{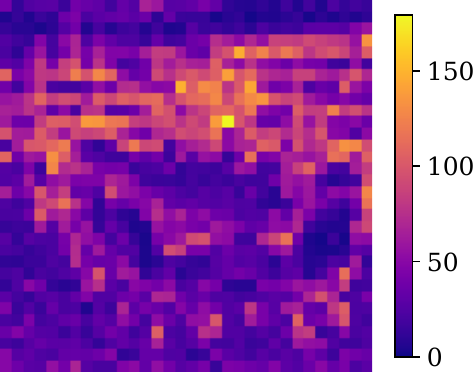} \\
$q=5$ (0.2739bpp) & $q=6$ (0.3784bpp) & $q=7$ (0.5108bpp) & $q=8$ (0.6706bpp) \\
\includegraphics[width=0.2\linewidth]{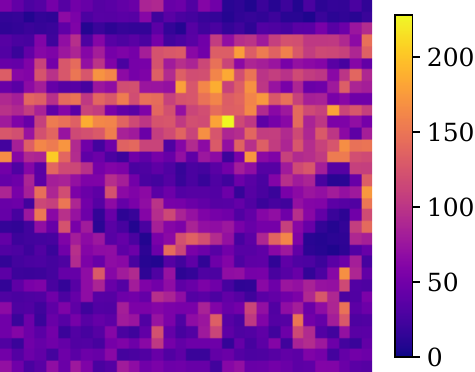} &
\includegraphics[width=0.2\linewidth]{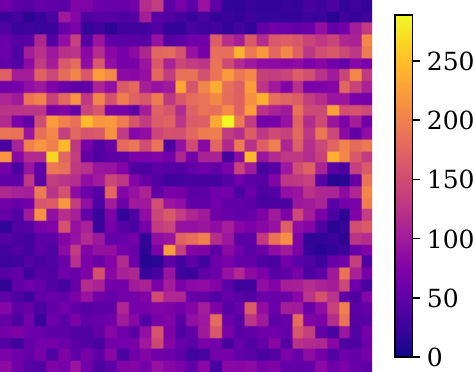} &
\includegraphics[width=0.2\linewidth]{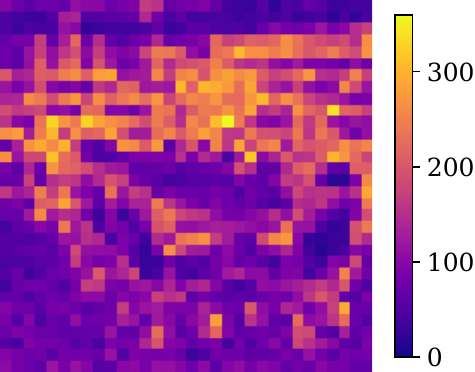} &
\includegraphics[width=0.2\linewidth]{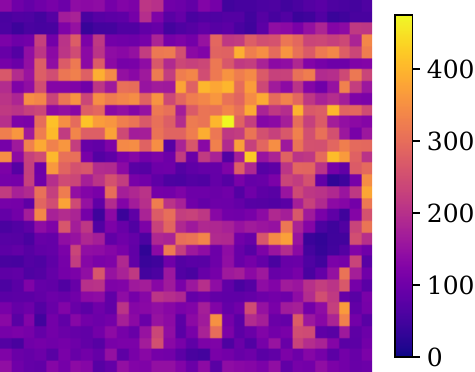} \\ \bottomrule
\end{tabular}
}
}
\caption{We visualize the bit allocation and score maps in the encoder. The image is from ADE20K validation set~\cite{ade20k} and resized to 512$\times$512$\times$3. The 1st, 2nd, and 3rd stages are the first three stages of the encoder, consecutively.}
\label{fig:enc_val00000269}
\end{center}
\end{figure}

\begin{figure}[htbp]
\begin{center}
{
\resizebox{\linewidth}{!}{
\scriptsize
\begin{tabular}{@{\hskip 0mm}l@{\hskip 1.0mm}l@{\hskip 1.0mm}l@{\hskip 1.0mm}l@{\hskip 1.0mm}l@{\hskip 0mm}}

\toprule
Reconstruction &
Segmentation Map &
Score Map (3-rd Stage) &
Score Map (2-nd Stage) &
Score Map (1-st Stage)
\\ \midrule

Cls. Path: 31.53dB
& & \multicolumn{3}{c}{The score maps are inverted for better visualization (higher for the side path).} \\
\includegraphics[width=0.16\linewidth]{figures/visualization/ade20k/val00000269/val00000269_cls.png} &
\includegraphics[width=0.16\linewidth]{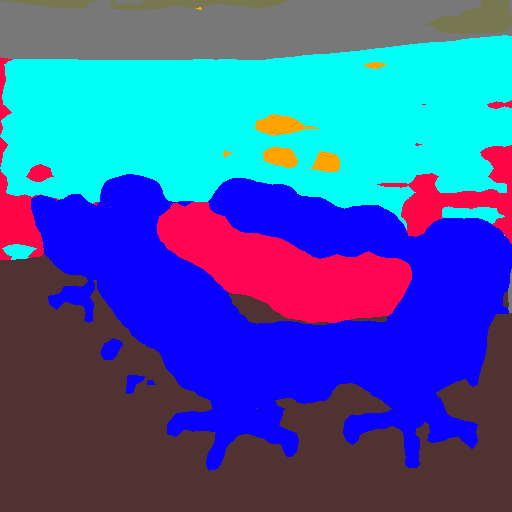} &
\includegraphics[width=0.2\linewidth]{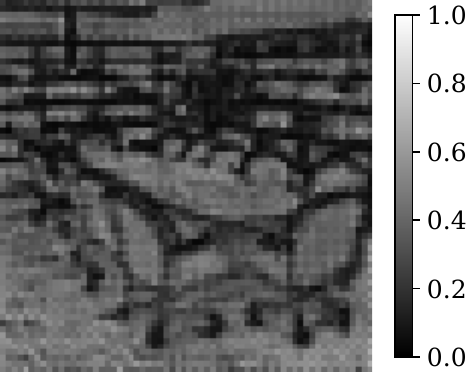} &
\includegraphics[width=0.2\linewidth]{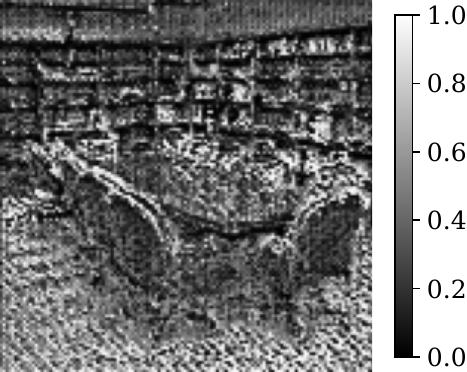} &
\includegraphics[width=0.2\linewidth]{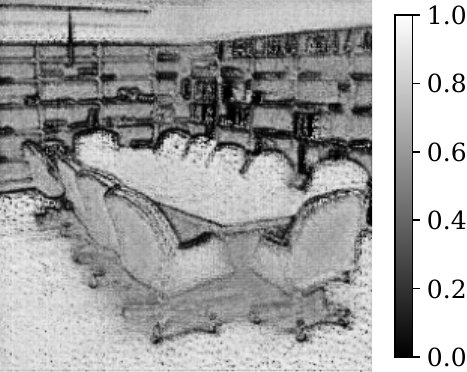} \\
Seg. Path: 31.62dB
& & & & \\
\includegraphics[width=0.16\linewidth]{figures/visualization/ade20k/val00000269/val00000269_seg.png} &
\includegraphics[width=0.16\linewidth]{figures/visualization/ade20k/val00000269/val00000269_seg_color.png} &
\includegraphics[width=0.2\linewidth]{figures/visualization/ade20k/val00000269/val00000269_predictor_seg_g_s2.pdf} &
\includegraphics[width=0.2\linewidth]{figures/visualization/ade20k/val00000269/val00000269_predictor_seg_g_s4.pdf} &
\includegraphics[width=0.2\linewidth]{figures/visualization/ade20k/val00000269/val00000269_predictor_seg_g_s6.pdf} \\
MSE Path: 31.87dB
& & & & \\
\includegraphics[width=0.16\linewidth]{figures/visualization/ade20k/val00000269/val00000269_mse.png} &
\includegraphics[width=0.16\linewidth]{figures/visualization/ade20k/val00000269/val00000269_mse_color.png} &
\includegraphics[width=0.2\linewidth]{figures/visualization/ade20k/val00000269/val00000269_predictor_mse_g_s2.pdf} &
\includegraphics[width=0.2\linewidth]{figures/visualization/ade20k/val00000269/val00000269_predictor_mse_g_s4.pdf} &
\includegraphics[width=0.2\linewidth]{figures/visualization/ade20k/val00000269/val00000269_predictor_mse_g_s6.pdf} \\
\\[-2ex]\midrule

Perc. Path: 31.30dB
& &
& Original: \texttt{val00000269}
& Ground Truth \\
\includegraphics[width=0.16\linewidth]{figures/visualization/ade20k/val00000269/val00000269_perc.png} &
\includegraphics[width=0.16\linewidth]{figures/visualization/ade20k/val00000269/val00000269_perc_color.png} & &
\includegraphics[width=0.16\linewidth]{figures/visualization/ade20k/val00000269/val00000269_original.png} &
\includegraphics[width=0.16\linewidth]{figures/visualization/ade20k/val00000269/val00000269_ground_truth_color.png} \\ \bottomrule
\end{tabular}
}
}
\caption{We visualize the reconstructed images, segmentation map and score maps predicted by importance predictors in each path. The image is from ADE20K validation set~\cite{ade20k} and resized to 512$\times$512$\times$3. The 3rd, 2nd, and 1st stages are the last three stages of the decoder, consecutively. $q$ is set to 1 for a more distinct comparison. Note that the score maps are inverted for better visualization, i.e., larger scores indicate prioritized for entering the selected side path. The bitrate is 0.0718bpp.}
\label{fig:seg_val00000269}
\end{center}
\end{figure}


\begin{figure}[htbp]
\begin{center}
{
\resizebox{\linewidth}{!}{
\scriptsize
\begin{tabular}{@{\hskip 0mm}l@{\hskip 1.0mm}l@{\hskip 1.0mm}l@{\hskip 1.0mm}l@{\hskip 0mm}}

\toprule
Original: \texttt{val00001200} &
Score Map (1-st Stage) &
Score Map (2-nd Stage) &
Score Map (3-rd Stage) \\

\includegraphics[width=0.16\linewidth]{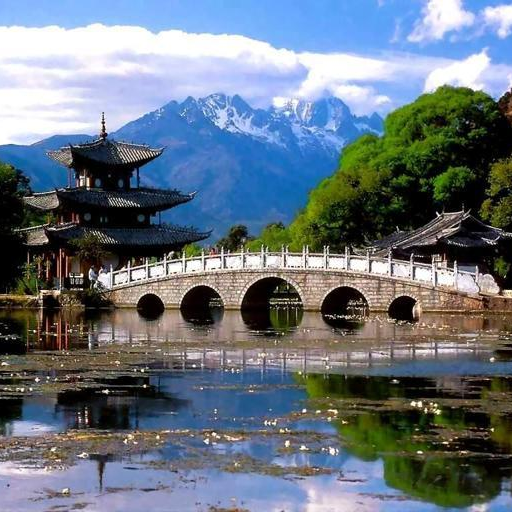} &
\includegraphics[width=0.2\linewidth]{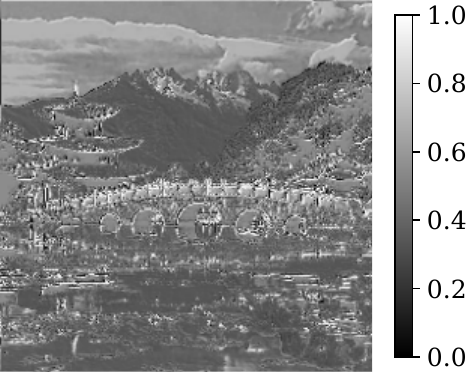} &
\includegraphics[width=0.2\linewidth]{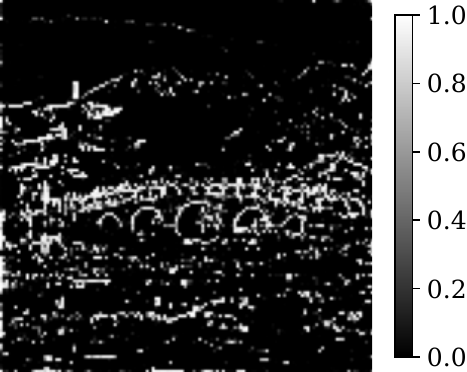} &
\includegraphics[width=0.2\linewidth]{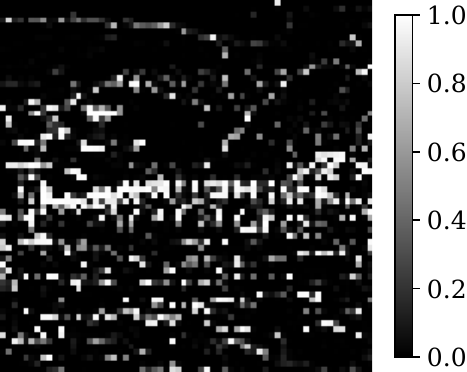} \\
\\[-2ex]\midrule

\multicolumn{4}{c}{Bit Allocation} \\

$q=1$ (0.1806bpp) & $q=2$ (0.2667bpp) & $q=3$ (0.3870bpp) & $q=4$ (0.5521bpp) \\
\includegraphics[width=0.2\linewidth]{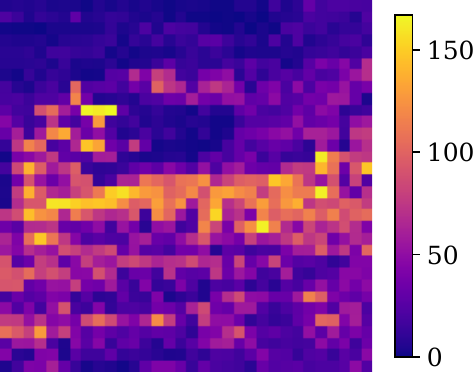} &
\includegraphics[width=0.2\linewidth]{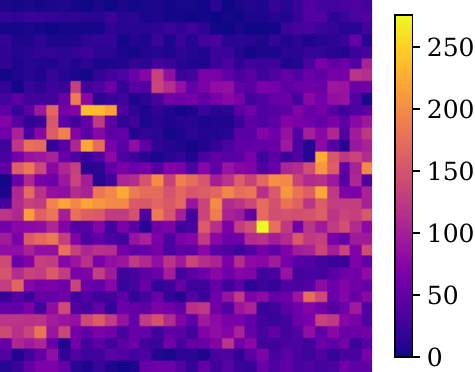} &
\includegraphics[width=0.2\linewidth]{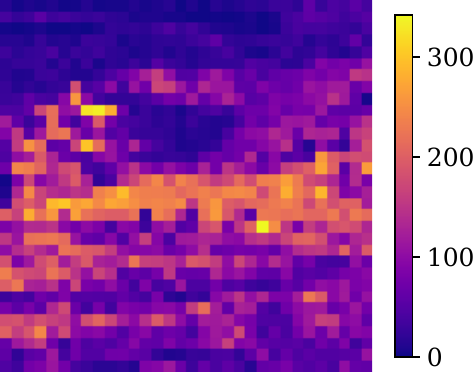} &
\includegraphics[width=0.2\linewidth]{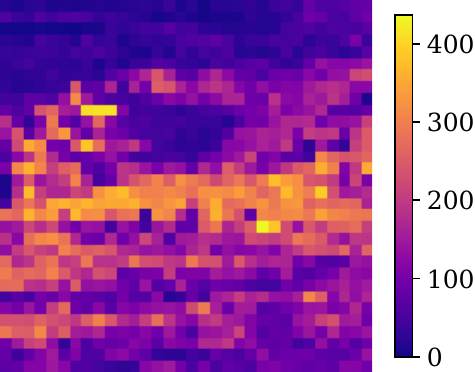} \\
$q=5$ (0.7644bpp) & $q=6$ (1.0471bpp) & $q=7$ (1.4075bpp) & $q=8$ (1.8221bpp) \\
\includegraphics[width=0.2\linewidth]{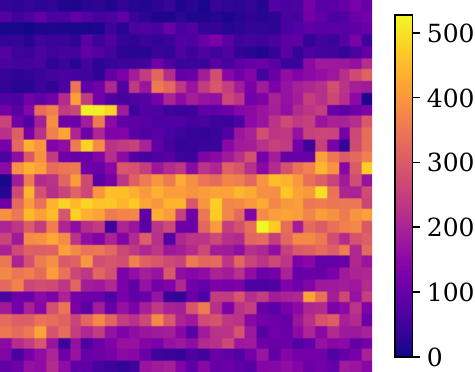} &
\includegraphics[width=0.2\linewidth]{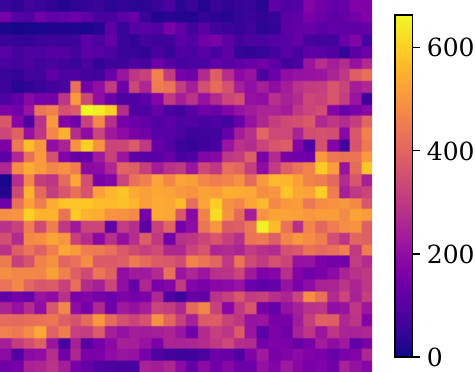} &
\includegraphics[width=0.2\linewidth]{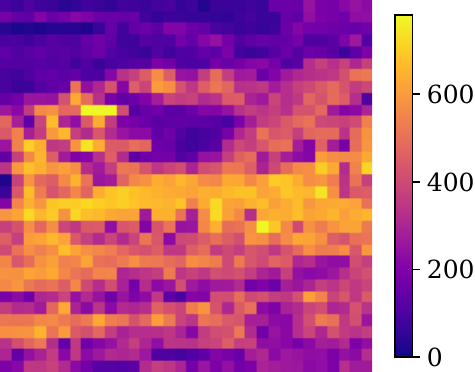} &
\includegraphics[width=0.2\linewidth]{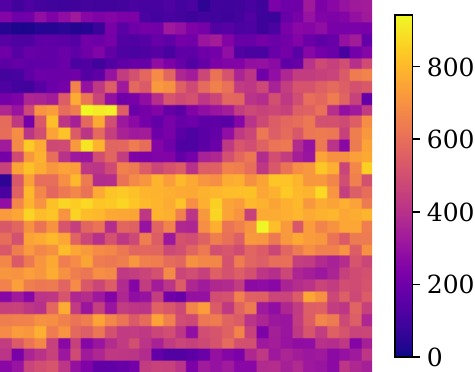} \\ \bottomrule
\end{tabular}
}
}
\caption{We visualize the bit allocation and score maps in the encoder. The image is from ADE20K validation set~\cite{ade20k} and resized to 512$\times$512$\times$3. The 1st, 2nd, and 3rd stages are the first three stages of the encoder, consecutively.}
\label{fig:enc_val00001200}
\end{center}
\end{figure}

\begin{figure}[htbp]
\begin{center}
{
\resizebox{\linewidth}{!}{
\scriptsize
\begin{tabular}{@{\hskip 0mm}l@{\hskip 1.0mm}l@{\hskip 1.0mm}l@{\hskip 1.0mm}l@{\hskip 1.0mm}l@{\hskip 0mm}}

\toprule
Reconstruction &
Segmentation Map &
Score Map (3-rd Stage) &
Score Map (2-nd Stage) &
Score Map (1-st Stage)
\\ \midrule

Cls. Path: 25.13dB
& & \multicolumn{3}{c}{The score maps are inverted for better visualization (higher for the side path).} \\
\includegraphics[width=0.16\linewidth]{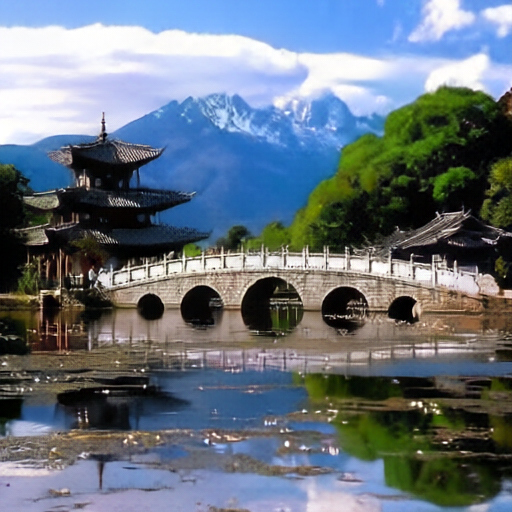} &
\includegraphics[width=0.16\linewidth]{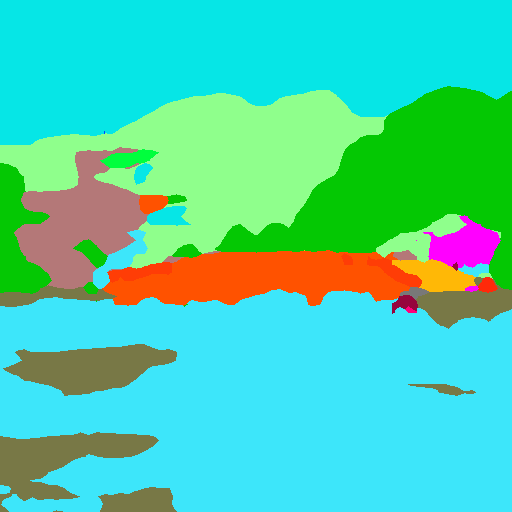} &
\includegraphics[width=0.2\linewidth]{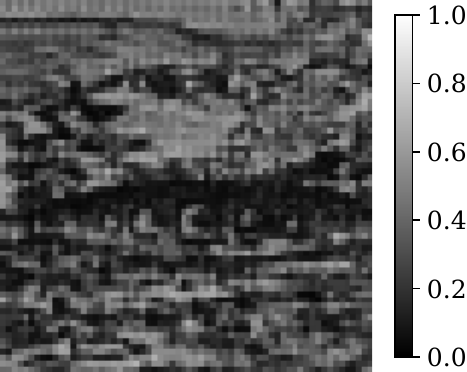} &
\includegraphics[width=0.2\linewidth]{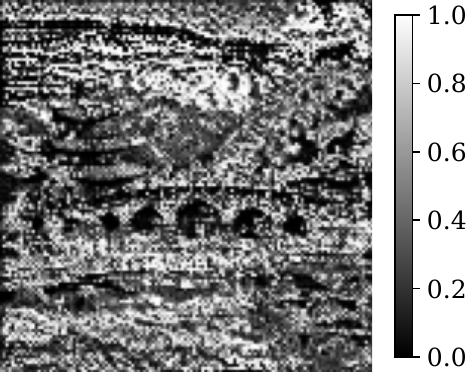} &
\includegraphics[width=0.2\linewidth]{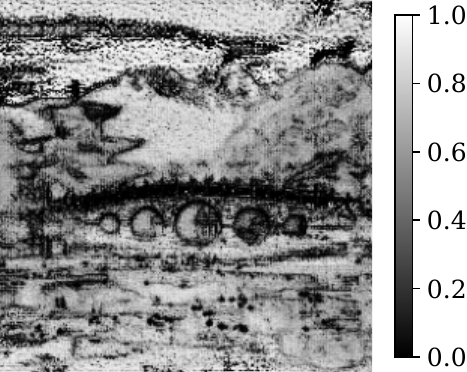} \\
Seg. Path: 25.17dB
& & & & \\
\includegraphics[width=0.16\linewidth]{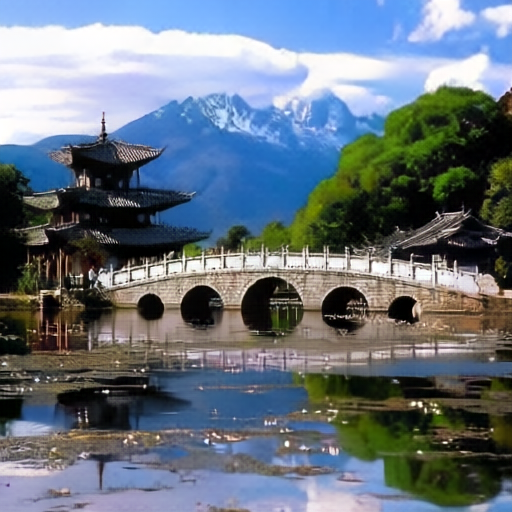} &
\includegraphics[width=0.16\linewidth]{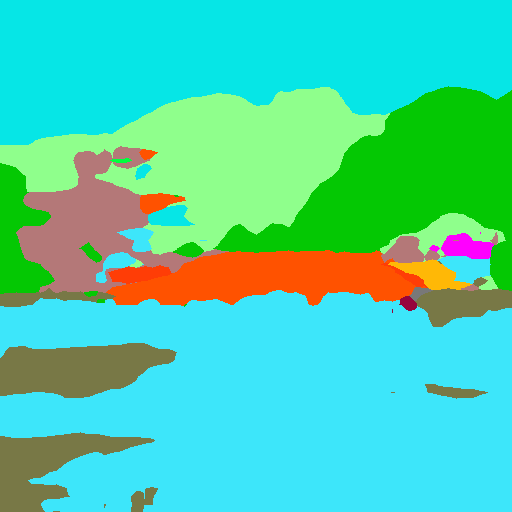} &
\includegraphics[width=0.2\linewidth]{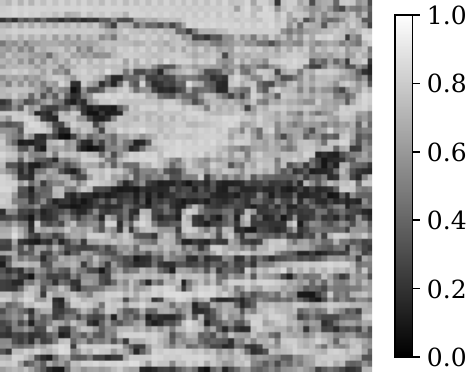} &
\includegraphics[width=0.2\linewidth]{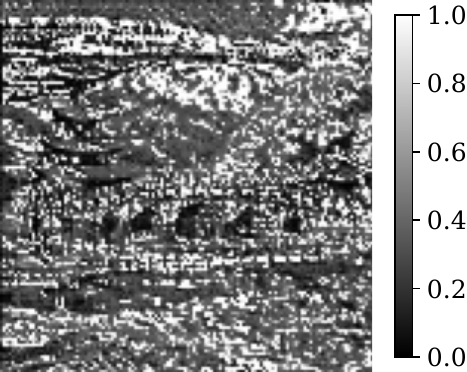} &
\includegraphics[width=0.2\linewidth]{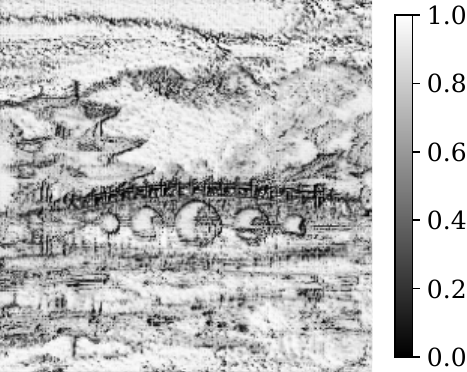} \\
MSE Path: 25.45dB
& & & & \\
\includegraphics[width=0.16\linewidth]{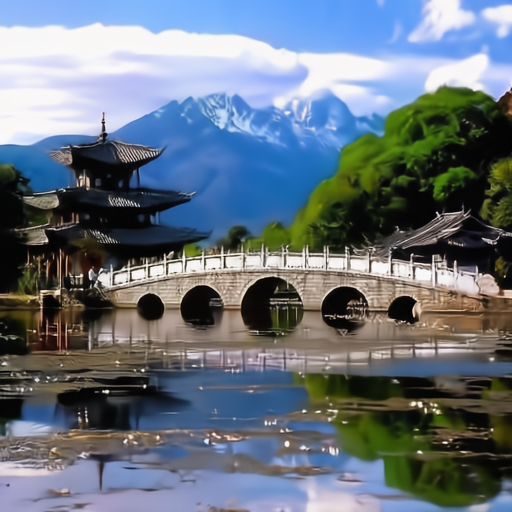} &
\includegraphics[width=0.16\linewidth]{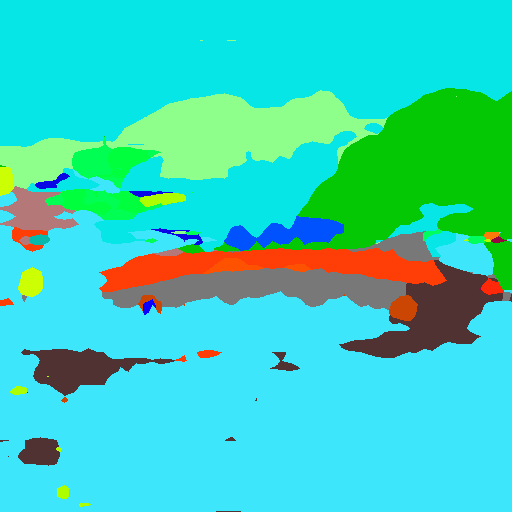} &
\includegraphics[width=0.2\linewidth]{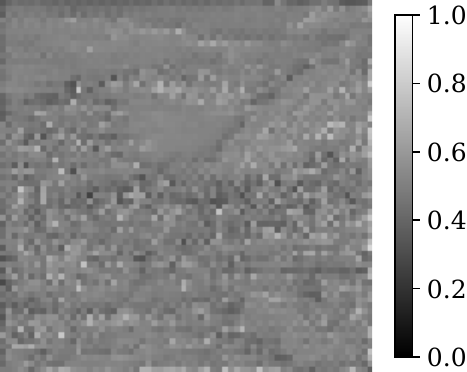} &
\includegraphics[width=0.2\linewidth]{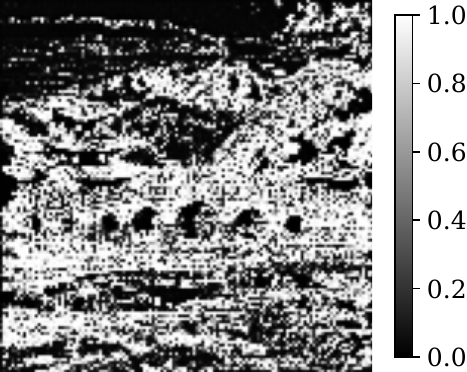} &
\includegraphics[width=0.2\linewidth]{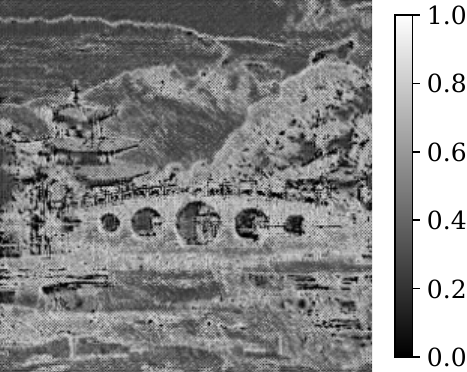} \\
\\[-2ex]\midrule

Perc. Path: 24.28dB
& &
& Original: \texttt{val00001200}
& Ground Truth \\
\includegraphics[width=0.16\linewidth]{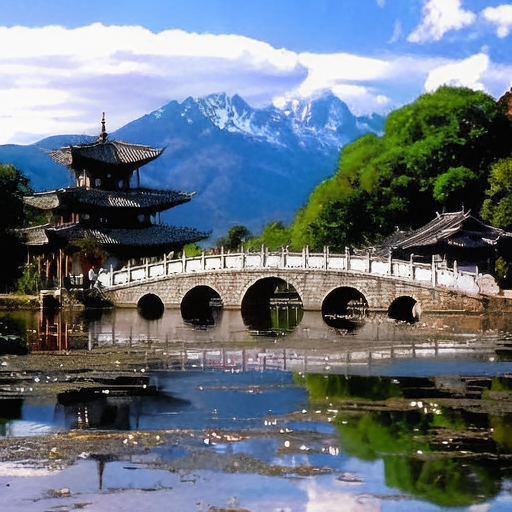} &
\includegraphics[width=0.16\linewidth]{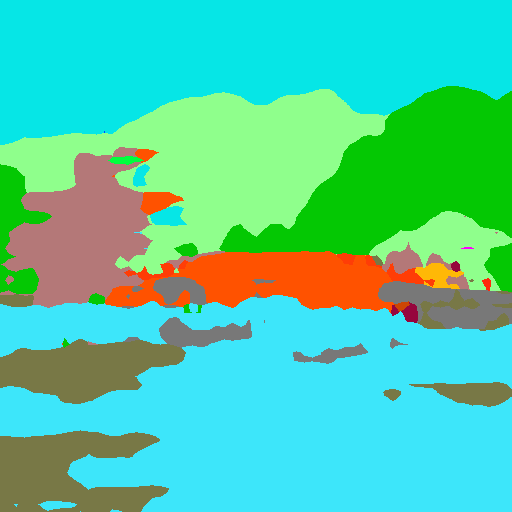} & &
\includegraphics[width=0.16\linewidth]{figures/visualization/ade20k/val00001200/val00001200_original.png} &
\includegraphics[width=0.16\linewidth]{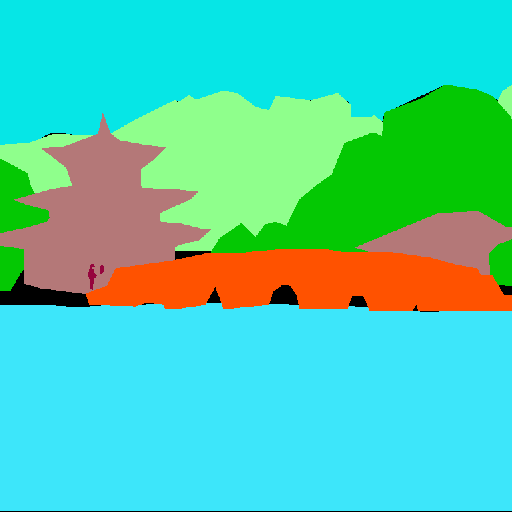} \\ \bottomrule
\end{tabular}
}
}
\caption{We visualize the reconstructed images, segmentation map and score maps predicted by importance predictors in each path. The image is from ADE20K validation set~\cite{ade20k} and resized to 512$\times$512$\times$3. The 3rd, 2nd, and 1st stages are the last three stages of the decoder, consecutively. $q$ is set to 1 for a more distinct comparison. Note that the score maps are inverted for better visualization, i.e., larger scores indicate prioritized for entering the selected side path. The bitrate is 0.1806bpp.}
\label{fig:seg_val00001200}
\end{center}
\end{figure}


\begin{figure}[htbp]
\begin{center}
{
\resizebox{\linewidth}{!}{
\scriptsize
\begin{tabular}{@{\hskip 0mm}l@{\hskip 1.0mm}l@{\hskip 1.0mm}l@{\hskip 1.0mm}l@{\hskip 0mm}}

\toprule
Original: \texttt{val00001969} &
Score Map (1-st Stage) &
Score Map (2-nd Stage) &
Score Map (3-rd Stage) \\

\includegraphics[width=0.16\linewidth]{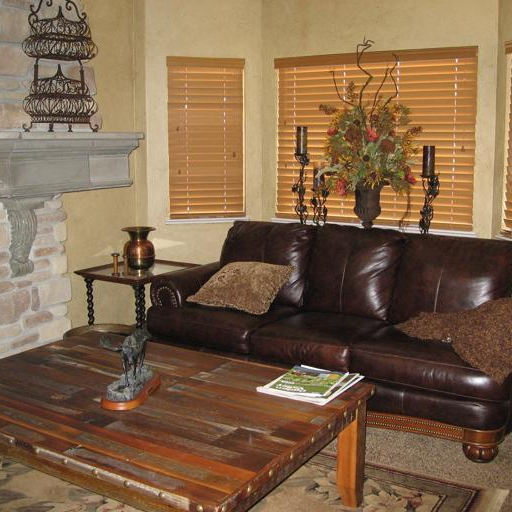} &
\includegraphics[width=0.2\linewidth]{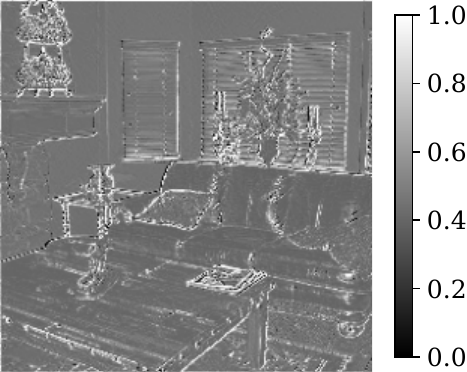} &
\includegraphics[width=0.2\linewidth]{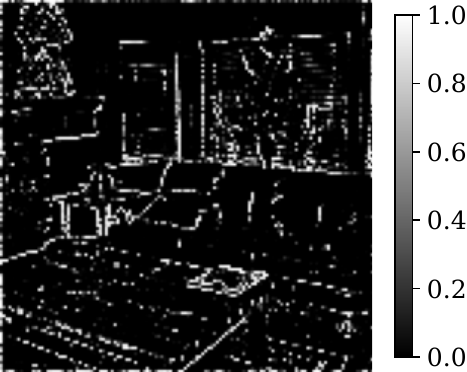} &
\includegraphics[width=0.2\linewidth]{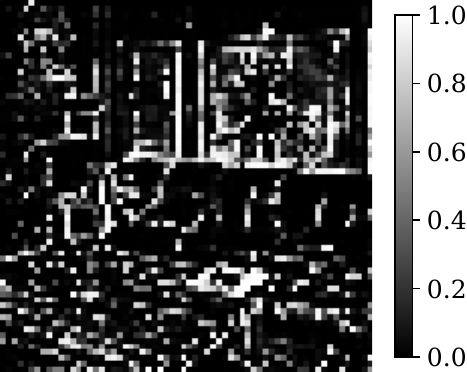} \\
\\[-2ex]\midrule

\multicolumn{4}{c}{Bit Allocation} \\

$q=1$ (0.1164bpp) & $q=2$ (0.1744bpp) & $q=3$ (0.2586bpp) & $q=4$ (0.3856bpp) \\
\includegraphics[width=0.2\linewidth]{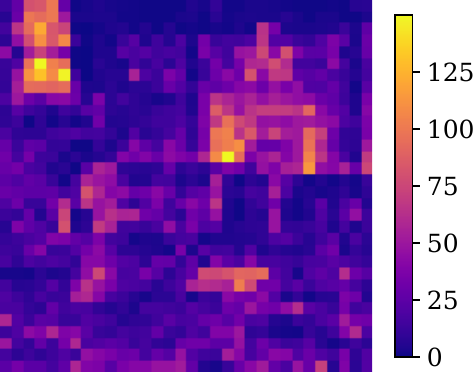} &
\includegraphics[width=0.2\linewidth]{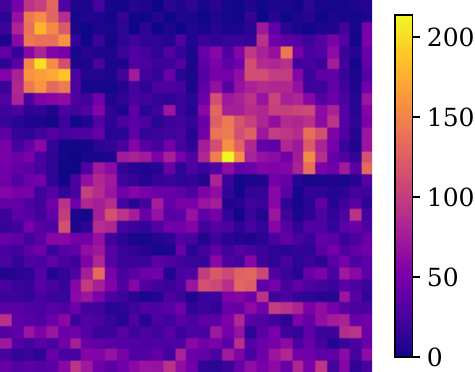} &
\includegraphics[width=0.2\linewidth]{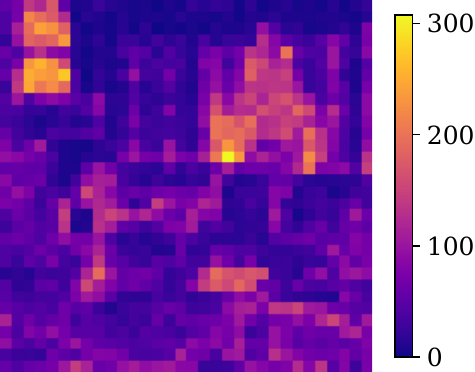} &
\includegraphics[width=0.2\linewidth]{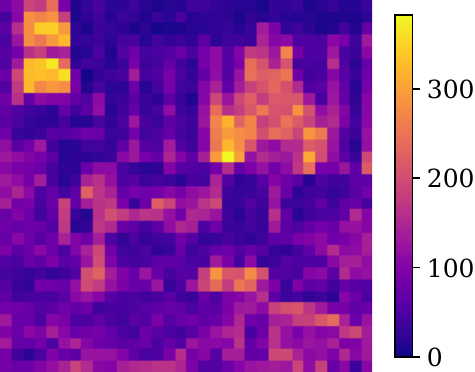} \\
$q=5$ (0.5589bpp) & $q=6$ (0.8149bpp) & $q=7$ (1.1666bpp) & $q=8$ (1.5661bpp) \\
\includegraphics[width=0.2\linewidth]{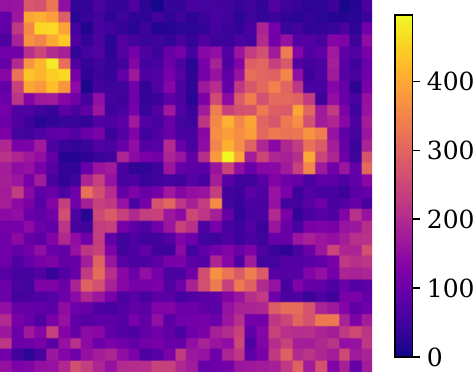} &
\includegraphics[width=0.2\linewidth]{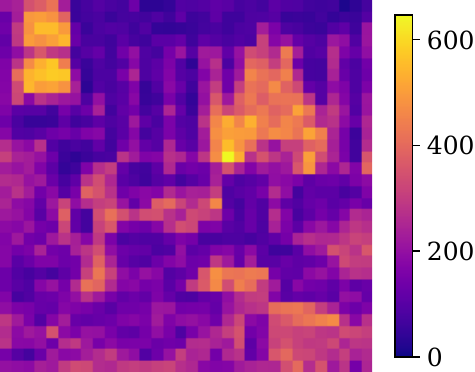} &
\includegraphics[width=0.2\linewidth]{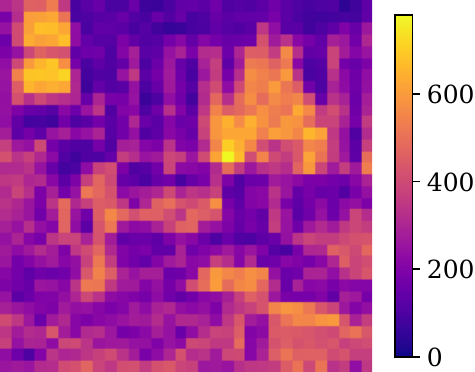} &
\includegraphics[width=0.2\linewidth]{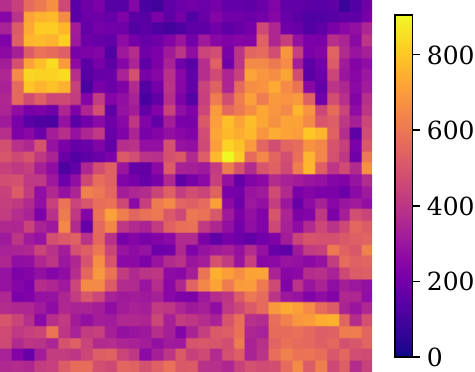} \\ \bottomrule
\end{tabular}
}
}
\caption{We visualize the bit allocation and score maps in the encoder. The image is from ADE20K validation set~\cite{ade20k} and resized to 512$\times$512$\times$3. The 1st, 2nd, and 3rd stages are the first three stages of the encoder, consecutively.}
\label{fig:enc_val00001969}
\end{center}
\end{figure}

\begin{figure}[htbp]
\begin{center}
{
\resizebox{\linewidth}{!}{
\scriptsize
\begin{tabular}{@{\hskip 0mm}l@{\hskip 1.0mm}l@{\hskip 1.0mm}l@{\hskip 1.0mm}l@{\hskip 1.0mm}l@{\hskip 0mm}}

\toprule
Reconstruction &
Segmentation Map &
Score Map (3-rd Stage) &
Score Map (2-nd Stage) &
Score Map (1-st Stage)
\\ \midrule

Cls. Path: 27.03dB
& & \multicolumn{3}{c}{The score maps are inverted for better visualization (higher for the side path).} \\
\includegraphics[width=0.16\linewidth]{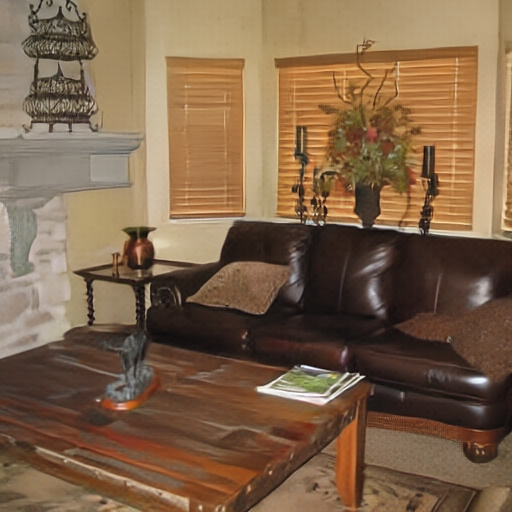} &
\includegraphics[width=0.16\linewidth]{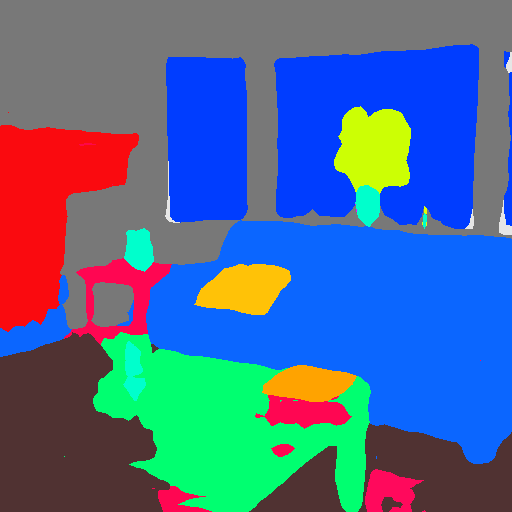} &
\includegraphics[width=0.2\linewidth]{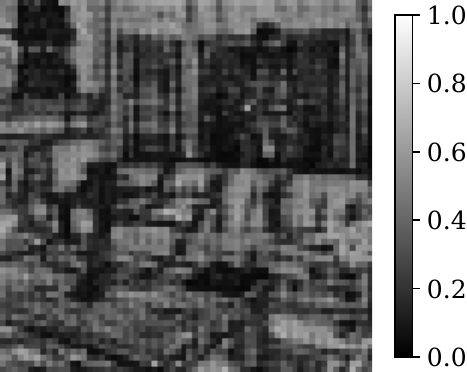} &
\includegraphics[width=0.2\linewidth]{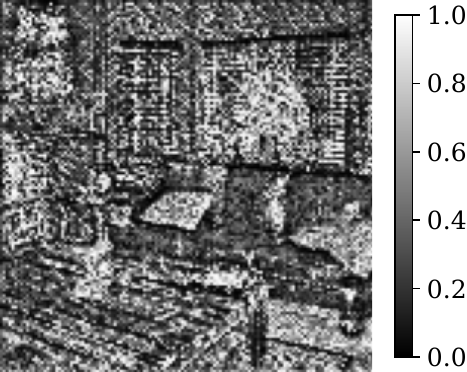} &
\includegraphics[width=0.2\linewidth]{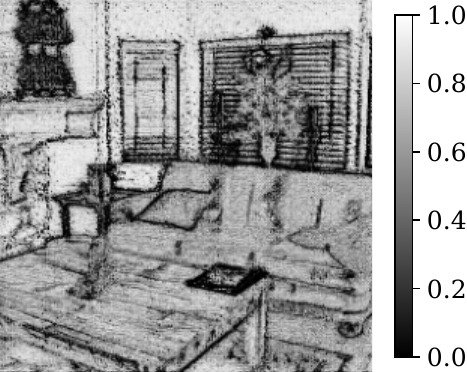} \\
Seg. Path: 27.07dB
& & & & \\
\includegraphics[width=0.16\linewidth]{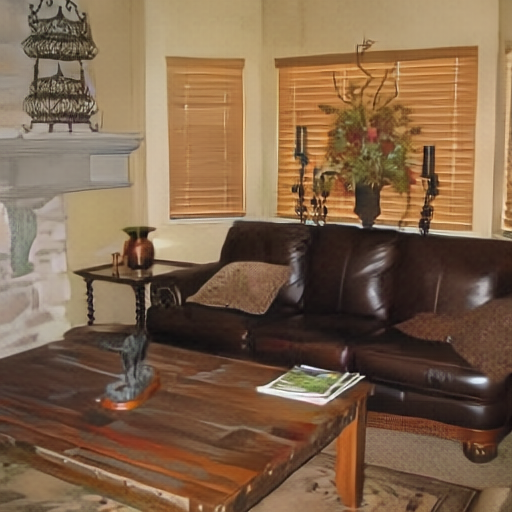} &
\includegraphics[width=0.16\linewidth]{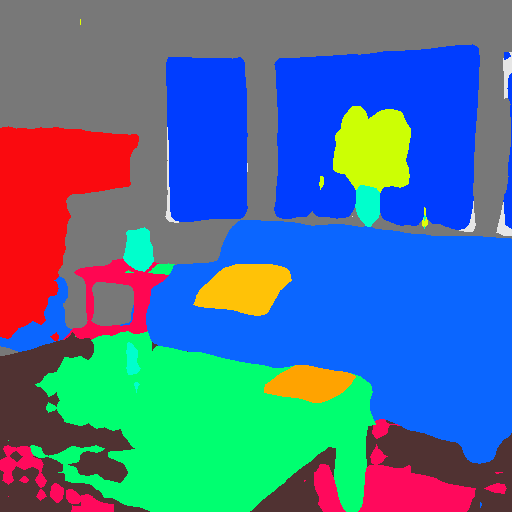} &
\includegraphics[width=0.2\linewidth]{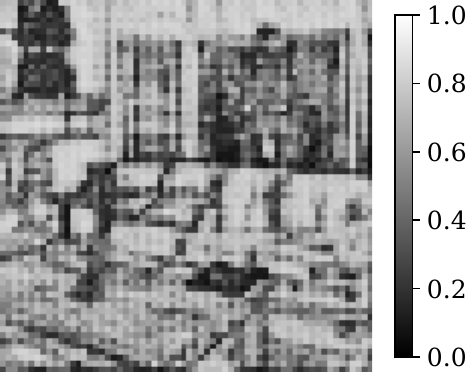} &
\includegraphics[width=0.2\linewidth]{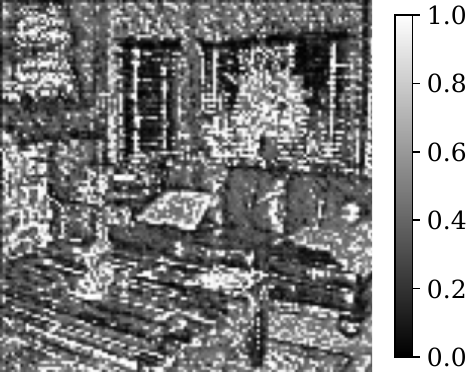} &
\includegraphics[width=0.2\linewidth]{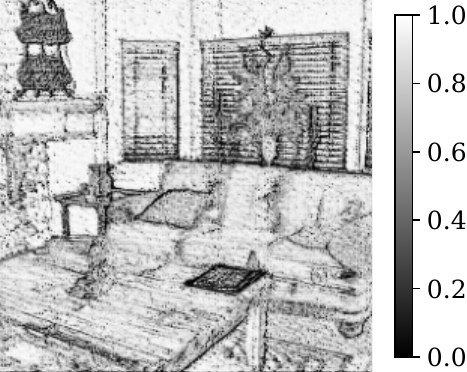} \\
MSE Path: 27.50dB
& & & & \\
\includegraphics[width=0.16\linewidth]{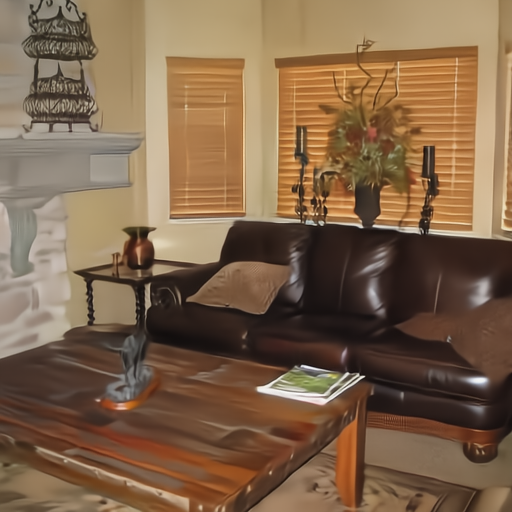} &
\includegraphics[width=0.16\linewidth]{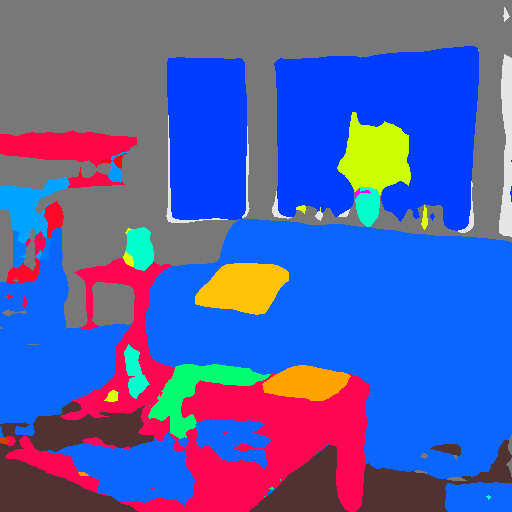} &
\includegraphics[width=0.2\linewidth]{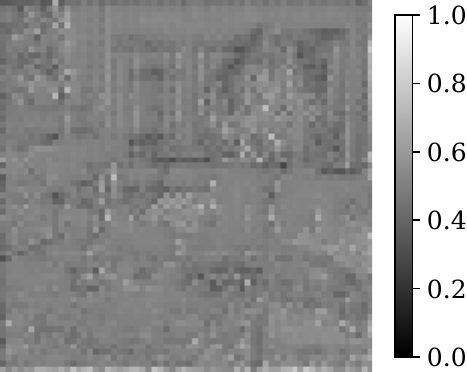} &
\includegraphics[width=0.2\linewidth]{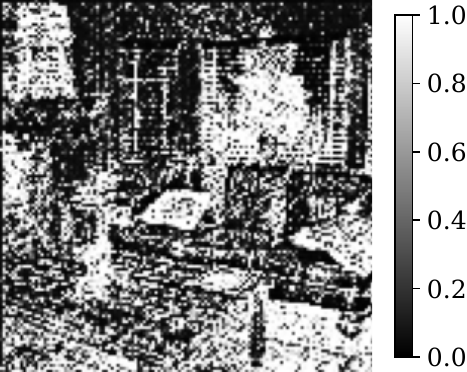} &
\includegraphics[width=0.2\linewidth]{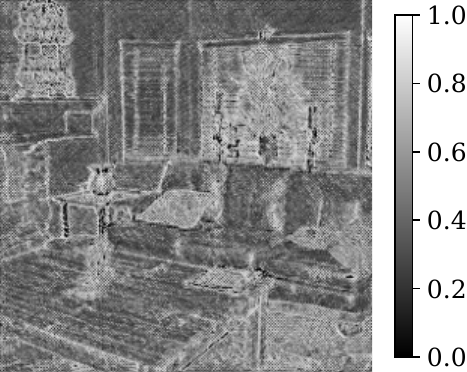} \\
\\[-2ex]\midrule

Perc. Path: 26.10dB
& &
& Original: \texttt{val00001969}
& Ground Truth \\
\includegraphics[width=0.16\linewidth]{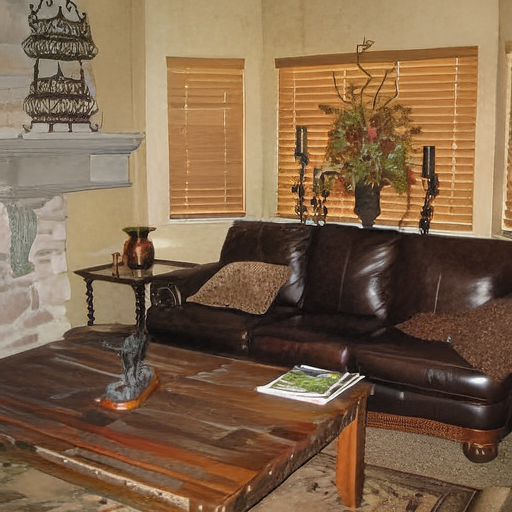} &
\includegraphics[width=0.16\linewidth]{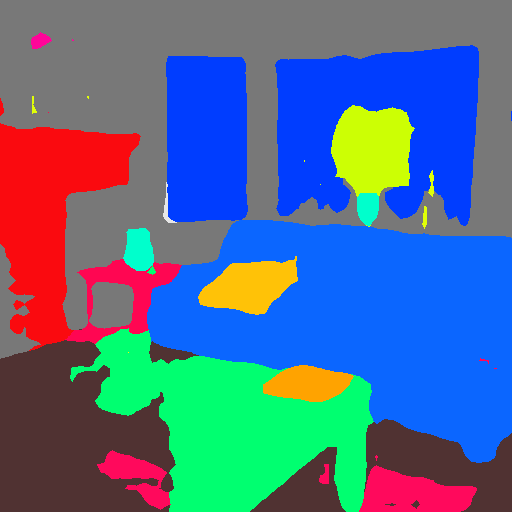} & &
\includegraphics[width=0.16\linewidth]{figures/visualization/ade20k/val00001969/val00001969_original.png} &
\includegraphics[width=0.16\linewidth]{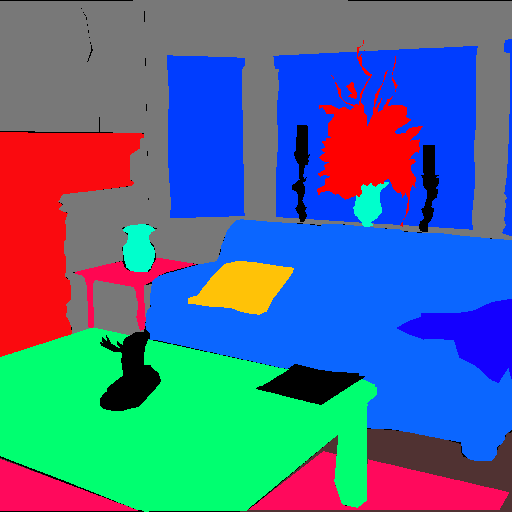} \\ \bottomrule
\end{tabular}
}
}
\caption{We visualize the reconstructed images, segmentation map and score maps predicted by importance predictors in each path. The image is from ADE20K validation set~\cite{ade20k} and resized to 512$\times$512$\times$3. The 3rd, 2nd, and 1st stages are the last three stages of the decoder, consecutively. $q$ is set to 1 for a more distinct comparison. Note that the score maps are inverted for better visualization, i.e., larger scores indicate prioritized for entering the selected side path. The bitrate is 0.1164bpp.}
\label{fig:seg_val00001969}
\end{center}
\end{figure}


\clearpage
\newpage
\section*{NeurIPS Paper Checklist}

\begin{enumerate}

\item {\bf Claims}
    \item[] Question: Do the main claims made in the abstract and introduction accurately reflect the paper's contributions and scope?
    \item[] Answer: \answerYes{} 
    \item[] Justification: We claim the contributions and scope in the abstract and Sec.~\ref{sec:intro}.

\item {\bf Limitations}
    \item[] Question: Does the paper discuss the limitations of the work performed by the authors?
    \item[] Answer: \answerYes{} 
    \item[] Justification: We discuss the limitations of the work in Sec.~\ref{sec:dive_into} and Appx.~\ref{sec:impacts}.

\item {\bf Theory Assumptions and Proofs}
    \item[] Question: For each theoretical result, does the paper provide the full set of assumptions and a complete (and correct) proof?
    \item[] Answer: \answerNA{} 
    \item[] Justification: The paper does not include theoretical results.

\item {\bf Experimental Result Reproducibility}
    \item[] Question: Does the paper fully disclose all the information needed to reproduce the main experimental results of the paper to the extent that it affects the main claims and/or conclusions of the paper (regardless of whether the code and data are provided or not)?
    \item[] Answer: \answerYes{} 
    \item[] Justification: All experimental details are provided in Secs.~\ref{sec:framework},~\ref{sec:training} and ~\ref{sec:settings}, and in Appx.~\ref{sec:impl}.

\item {\bf Open access to data and code}
    \item[] Question: Does the paper provide open access to the data and code, with sufficient instructions to faithfully reproduce the main experimental results, as described in supplemental material?
    \item[] Answer: \answerYes{} 
    \item[] Justification: We provide the instructions in Sec.~\ref{sec:training}. Code will be available upon acceptance.

\item {\bf Experimental Setting/Details}
    \item[] Question: Does the paper specify all the training and test details (e.g., data splits, hyperparameters, how they were chosen, type of optimizer, etc.) necessary to understand the results?
    \item[] Answer: \answerYes{} 
    \item[] Justification: We specify all the training and test details in Sec.~\ref{sec:settings}.

\item {\bf Experiment Statistical Significance}
    \item[] Question: Does the paper report error bars suitably and correctly defined or other appropriate information about the statistical significance of the experiments?
    \item[] Answer: \answerNo{} 
    \item[] Justification: The experiments do not involve statistical significance.

\item {\bf Experiments Compute Resources}
    \item[] Question: For each experiment, does the paper provide sufficient information on the computer resources (type of compute workers, memory, time of execution) needed to reproduce the experiments?
    \item[] Answer: \answerYes{} 
    \item[] Justification: We provide the information on the computer resources in Appx.~\ref{sec:reproducibility}.
    
\item {\bf Code Of Ethics}
    \item[] Question: Does the research conducted in the paper conform, in every respect, with the NeurIPS Code of Ethics \url{https://neurips.cc/public/EthicsGuidelines}?
    \item[] Answer: \answerYes{} 
    \item[] Justification: The research conducted in the paper conform, in every respect, with the NeurIPS Code of Ethics.

\item {\bf Broader Impacts}
    \item[] Question: Does the paper discuss both potential positive societal impacts and negative societal impacts of the work performed?
    \item[] Answer: \answerYes{} 
    \item[] Justification: We discuss the societal impacts in Appx.~\ref{sec:impacts}.
    
\item {\bf Safeguards}
    \item[] Question: Does the paper describe safeguards that have been put in place for responsible release of data or models that have a high risk for misuse (e.g., pretrained language models, image generators, or scraped datasets)?
    \item[] Answer: \answerNA{} 
    \item[] Justification: The paper poses no such risks.

\item {\bf Licenses for existing assets}
    \item[] Question: Are the creators or original owners of assets (e.g., code, data, models), used in the paper, properly credited and are the license and terms of use explicitly mentioned and properly respected?
    \item[] Answer: \answerYes{} 
    \item[] Justification: The original papers are cited. The licenses are listed in Appx.~\ref{sec:licenses}.

\item {\bf New Assets}
    \item[] Question: Are new assets introduced in the paper well documented and is the documentation provided alongside the assets?
    \item[] Answer: \answerYes{} 
    \item[] Justification: Code is available at \url{https://github.com/NJUVISION/MPA}.

\item {\bf Crowdsourcing and Research with Human Subjects}
    \item[] Question: For crowdsourcing experiments and research with human subjects, does the paper include the full text of instructions given to participants and screenshots, if applicable, as well as details about compensation (if any)? 
    \item[] Answer: \answerNA{} 
    \item[] Justification: The paper does not involve crowdsourcing nor research with human subjects.

\item {\bf Institutional Review Board (IRB) Approvals or Equivalent for Research with Human Subjects}
    \item[] Question: Does the paper describe potential risks incurred by study participants, whether such risks were disclosed to the subjects, and whether Institutional Review Board (IRB) approvals (or an equivalent approval/review based on the requirements of your country or institution) were obtained?
    \item[] Answer: \answerNA{} 
    \item[] Justification: The paper does not involve crowdsourcing nor research with human subjects.

\end{enumerate}

\end{document}